\documentclass[english]{article}

\usepackage[english]{babel}
\usepackage[utf8]{inputenc}

\usepackage{color,xcolor,ucs}

\usepackage{subfig}
\usepackage{floatrow}
\usepackage{tabularx}
\usepackage{float}
\usepackage{amsfonts}
\usepackage{helvet}         
\usepackage{courier}        
\usepackage{type1cm}        
\usepackage{amsmath}
\usepackage{amssymb}
\usepackage{makeidx}         
\usepackage{comment}         
\usepackage{graphicx}        
\usepackage{multicol}        
\usepackage[bottom]{footmisc}
\usepackage{bm}

\usepackage{url}

\usepackage{xr-hyper}
\usepackage[unicode=true, bookmarks=true,colorlinks=true]{hyperref}

\usepackage{algpseudocode,algorithm,algorithmicx}

\usepackage{geometry}
\usepackage{xspace}
\usepackage{rotating}

\usepackage{sidecap}
\usepackage{graphicx}
\usepackage{threeparttable}
\usepackage{multirow}

\usepackage[toc,page]{appendix}

\usepackage{natbib}
\setcitestyle{authoryear,open={(},close={)}}

\DeclareMathOperator*{\argmin}{arg\,min}
\DeclareMathOperator*{\argmax}{arg\,max}
\DeclareMathOperator*{\Expec}{\mathbb{E}}

\newcommand{\JIP}{\ensuremath{\texttt{JIP}}\xspace}
\newcommand{\RUB}{\ensuremath{\texttt{RUB Inference}}\xspace}

\newcommand{\prob}[1]{\ensuremath{\mathbb{P}({#1})}}

\algrenewcommand\algorithmicrequire{\textbf{Input:}}
\algrenewcommand\algorithmicensure{\textbf{Input:}}
\algnewcommand{\LineComment}[1]{\State \(\triangleright\) #1}

\title{Bayesian Incremental Inference Update by Re-using Calculations from Belief Space Planning: A New Paradigm}

\author{\renewcommand\footnotemark{}Elad I. Farhi and Vadim Indelman
	\thanks{Elad I. Farhi is with the Technion Autonomous Systems Program (TASP), Technion - Israel Institute of Technology, Haifa 32000, Israel, {\tt{eladf@campus.technion.ac.il}}. Vadim Indelman is with the Department of Aerospace Engineering, Technion - Israel Institute of Technology, Haifa 32000, Israel, {\tt vadim.indelman@technion.ac.il}.
	}	
	\vspace{-15pt}
}

\date{}

\begin{document}

\maketitle

\begin{abstract}
Inference and decision making under uncertainty are key processes in every autonomous system and numerous robotic problems.
In recent years, the similarities between inference and decision making triggered much work, from developing unified computational frameworks to pondering about the duality between the two.   
In spite of these efforts, inference and control, as well as inference and belief space planning (BSP) are still treated as two separate processes.
In this paper we propose a paradigm shift, a novel approach which deviates from conventional Bayesian inference and utilizes the similarities between inference and BSP. We make the \emph{key observation} that inference can be efficiently updated using predictions made during the decision making stage, even in light of inconsistent data association between the two.
We developed a two staged process that implements our novel approach and updates inference using calculations from the precursory planning phase. Using autonomous navigation in an unknown environment along with \texttt{iSAM2} efficient methodologies as a test case, we benchmarked our novel approach against  standard Bayesian inference, both with synthetic and real-world data (KITTI dataset).
Results indicate that not only our approach improves running time by at least a factor of two while providing  the same estimation accuracy, but it also alleviates the computational burden of state dimensionality and loop closures.
\end{abstract}

\vspace{-5pt}

\section{Introduction}
\label{sec:introduction}
Real life scenarios in autonomous systems and artificial intelligence involve agent(s) that are expected to reliably and efficiently operate under different sources of uncertainty, often with limited knowledge regarding the environment; e.g. autonomous navigation and simultaneous localization and mapping (SLAM), search and rescue scenarios, object manipulation and robot-assisted surgery. These settings necessitate probabilistic reasoning regarding high dimensional problem-specific states. For instance, in  SLAM, the state typically represents robot poses and mapped static or dynamic landmarks, while in environmental monitoring and other sensor deployment related problems the state corresponds to an environmental field to be monitored (e.g.~temperature as a function of position and perhaps time).
	
Attaining these levels of autonomy involves two key processes, inference and decision making under uncertainty. The former maintains a belief regarding the high-dimensional state given available information thus far, while the latter, also often referred to as belief space planning (BSP), is entrusted with determining the next best action(s).  

The inference problem, has been addressed by the research community extensively over the past decades. In particular, focus was given to inference over high-dimensional state spaces with SLAM being a representative problem, and to computational efficiency to facilitate online operation, as required in numerous robotics systems.  
Over the years, the solution paradigm for the inference problem has evolved. From EKF based methods \citep{Davison07pami,Haykin01}, through information form recursive \citep{Thrun04ijrr} and smoothing methods \citep{Dellaert06ijrr,Eustice06tro}, and in recent years up to incremental smoothing approaches, such as \texttt{iSAM} \citep{Kaess08tro} and \texttt{iSAM2} \citep{Kaess12ijrr}. 

Given the posterior belief from the inference stage, decision making under uncertainty and belief space planning  approaches are entrusted with providing the next optimal action sequence given a certain objective. The aforementioned is accomplished by reasoning about belief evolution for different candidate actions while taking into account different sources of uncertainty. The corresponding problem is an instantiation of a partially observable Markov decision process (POMDP) problem, known as PSAPCE-complete \citep{Papadimitriou87math}, hence computationally intractable for all but the smallest problems, i.e.~no more than a few dozen states \citep{Kaelbling98ai}. 

Over the years, numerous approaches have been developed to trade-off suboptimal performance with reduced computational complexity of POMDP, see e.g.~\cite{Pineau06jair,Kurniawati08rss, Hollinger14ijrr,Toussaint09icml}. While the majority of these approaches, including \cite{Prentice09ijrr, Platt10rss, Bry11icra, VanDenBerg12ijrr}, assumed some sources of absolute information (GPS, known landmarks) are available or considered the environment to be known, recent research relaxed these assumptions, accounting for the uncertainties in the mapped environment thus far as part of the decision making process \citep{Kim14ijrr, Indelman15ijrr} at the price of increased state dimensionality. 

A crucial component in both inference and BSP is data association (DA), i.e.~associating between sensor observations and the corresponding landmarks. Incorrect DA in inference or BSP can lead to catastrophic failures, due to wrong estimation in inference or incorrect belief propagation within BSP that would result in incorrect, and potentially unsafe, actions. Recent research thus focused on developing approaches that are robust to incorrect DA, considering both passive  \citep{Carlone14iros, Olson13ijrr, Sunderhauf12icra, Indelman16csm} and active perception \citep{Pathak18ijrr}. 
\begin{figure}
	\centering
        \subfloat[]{\includegraphics[bb={0 0 0 0},trim={0 100 275 0},clip, width=0.3\columnwidth]{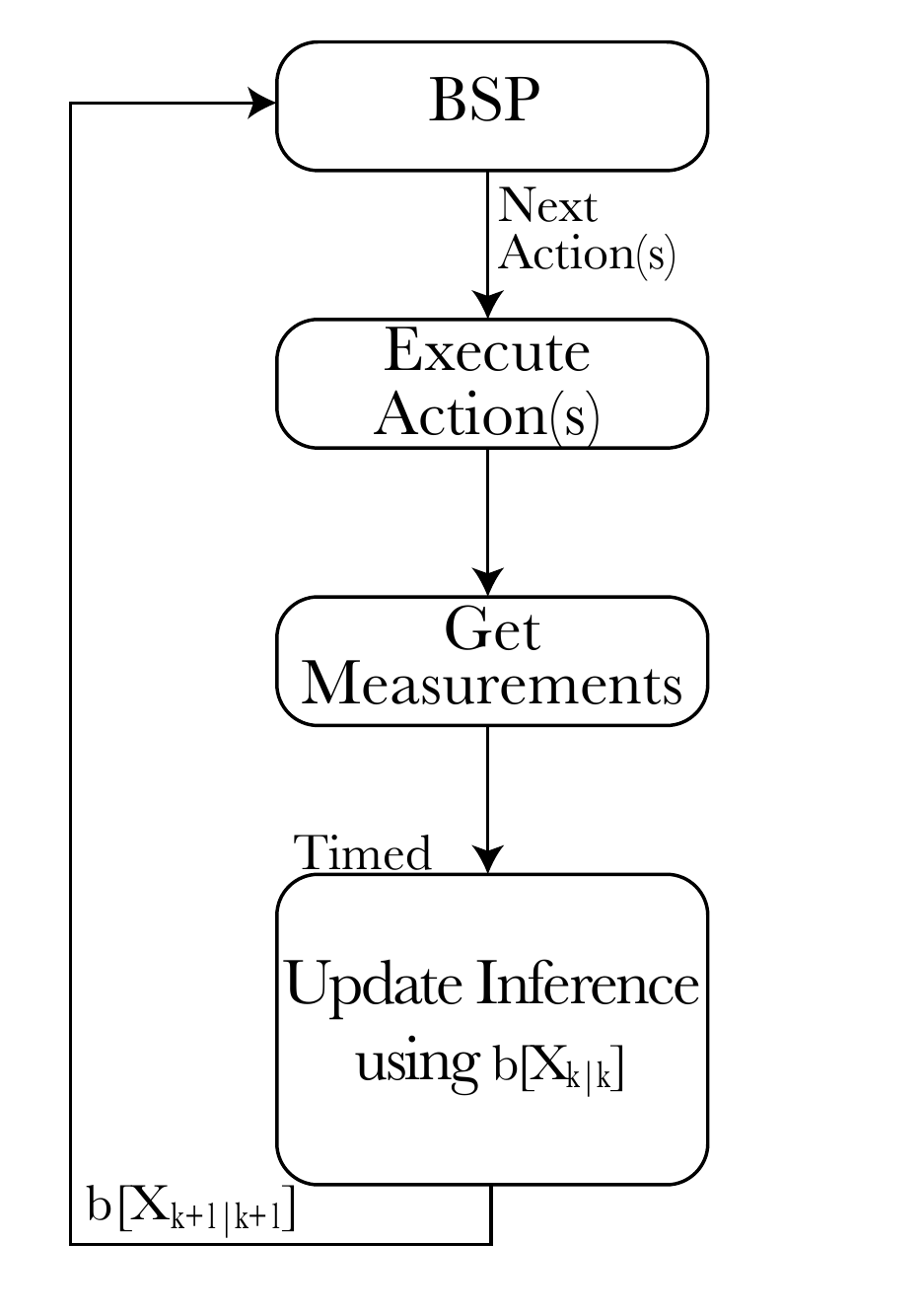}\label{fig:HighLevel_Algo:std}}
        \subfloat[]{\includegraphics[bb={0 0 0 0},trim={275 100 0 0},clip, width=0.3\columnwidth]{Figures/psudoCode_v5.pdf}\label{fig:HighLevel_Algo:ours}}
        \caption{High level algorithm for joint inference and BSP presented in a block diagram: (a)  presents a standard plan-act-infer framework with Bayesian inference and BSP treated as separate processes; (b) presents our novel approach for inference update using precursory planning. Instead of updating the belief from precursory inference with new information we propose to update the belief from a precursory planning phase. Since the only difference between (a) and (b) manifests in computation time within the inference block, it is timed for comparison. }
        \label{fig:HighLevel_Algo}
\end{figure}

Regardless of the decision making approach being used, in order to determine the next (sub)optimal actions the current belief is propagated using various action sequences. The propagated beliefs are then solved in order to provide an objective function value, thus enabling to determine the (sub)optimal actions. Solving a propagated belief is equivalent to performing inference over the belief, hence solving multiple inference problems is inevitable when trying to determine the (sub)optimal actions.
 
 However, despite the similarities between inference and decision making, the two problems have been typically treated separately. Only in recent years, the research community has started investigating and exploiting these similarities between the two processes.
For example, \cite{Kobilarov15icra} and \cite{Ta14icuas} developed Differential Dynamic Programming (DDP) and Factor Graph (FG) based unified computational frameworks, respectively, for inference and decision making. \cite{Toussaint06icml} provided an approximate solution to Markov Decision Process (MDP) problem using inference optimization methods, and \cite{Todorov08cdc} investigated the duality between optimal control and inference for MDP case.
Despite these research efforts, inference and BSP are still being handled as two separate processes.

Our \emph{key observation} is that similarities between inference and decision making paradigms could be utilized in order to save valuable computation time. 
Our approach is rooted in the joint inference and belief space planning concept, presented in \cite{Farhi17icra} and \cite{Farhi19icra_ws}, which strives to handle both inference and decision making in a partially observable setting within a unified framework, to enable sharing and updating similar calculations across inference and planning (see discussion in Section~\ref{sec:motivation}). 
In contrast to the notion of joint inference and control, which considers an MDP setting, we consider a partially observable setting (POMDP). Through the symbiotic relation enabled by considering the joint inference and BSP problems we make the following key research hypothesis: Inference can be efficiently updated using a precursory planning stage. 
This paper investigates this novel concept for inference update, considering operation in uncertain or unknown environments and compares it against the current state of the art in both simulated and real-life environments. 

\begin{figure}
	\centering
        \subfloat[]{\includegraphics[bb={0 0 0 0},trim={0 0 0 0},clip, width=0.45\textwidth]{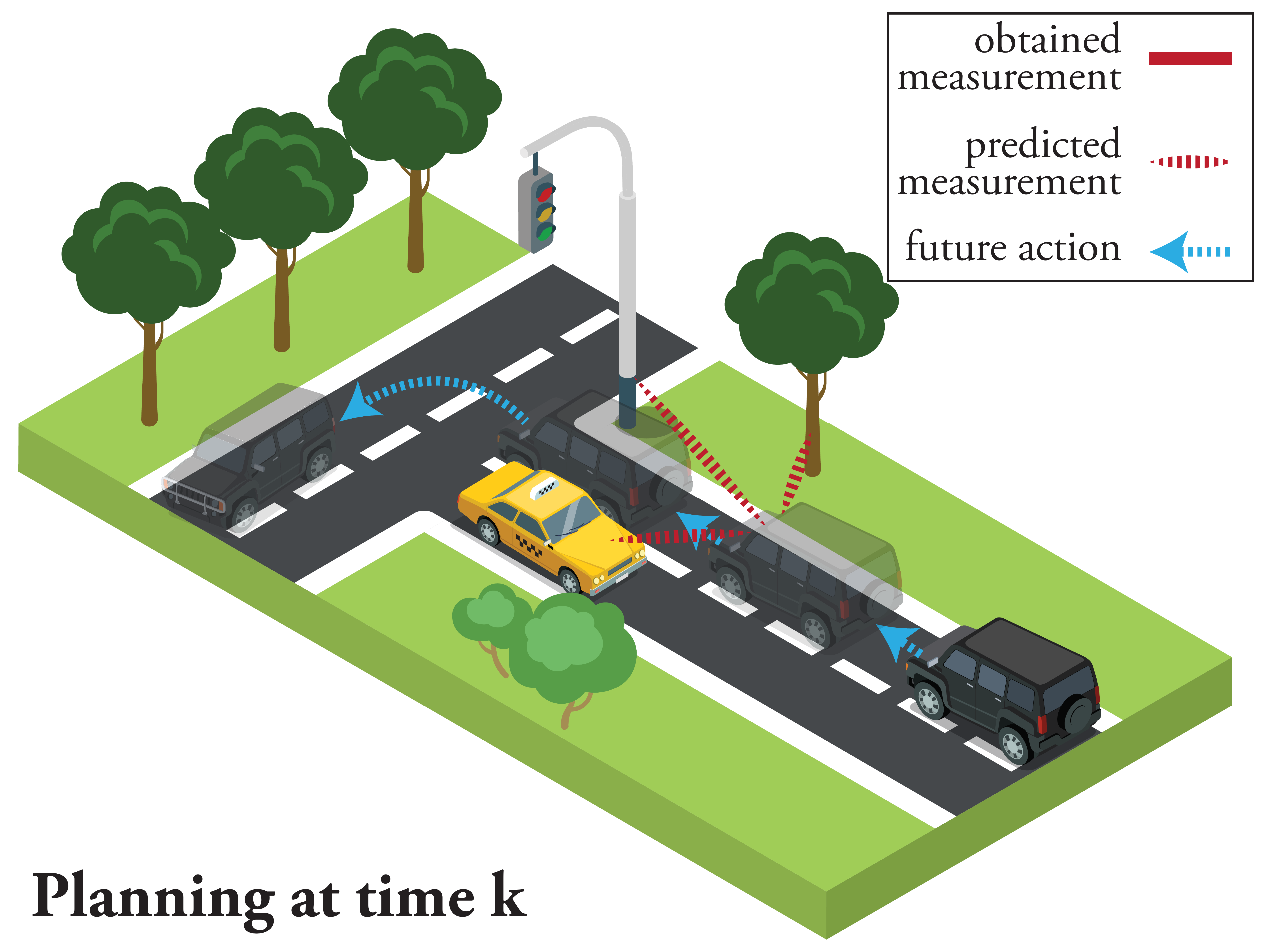}\label{fig:incDA:plan}} \vspace{-5pt}
        \subfloat[]{\includegraphics[bb={0 0 0 0},trim={0 0 0 0},clip, width=0.45\textwidth]{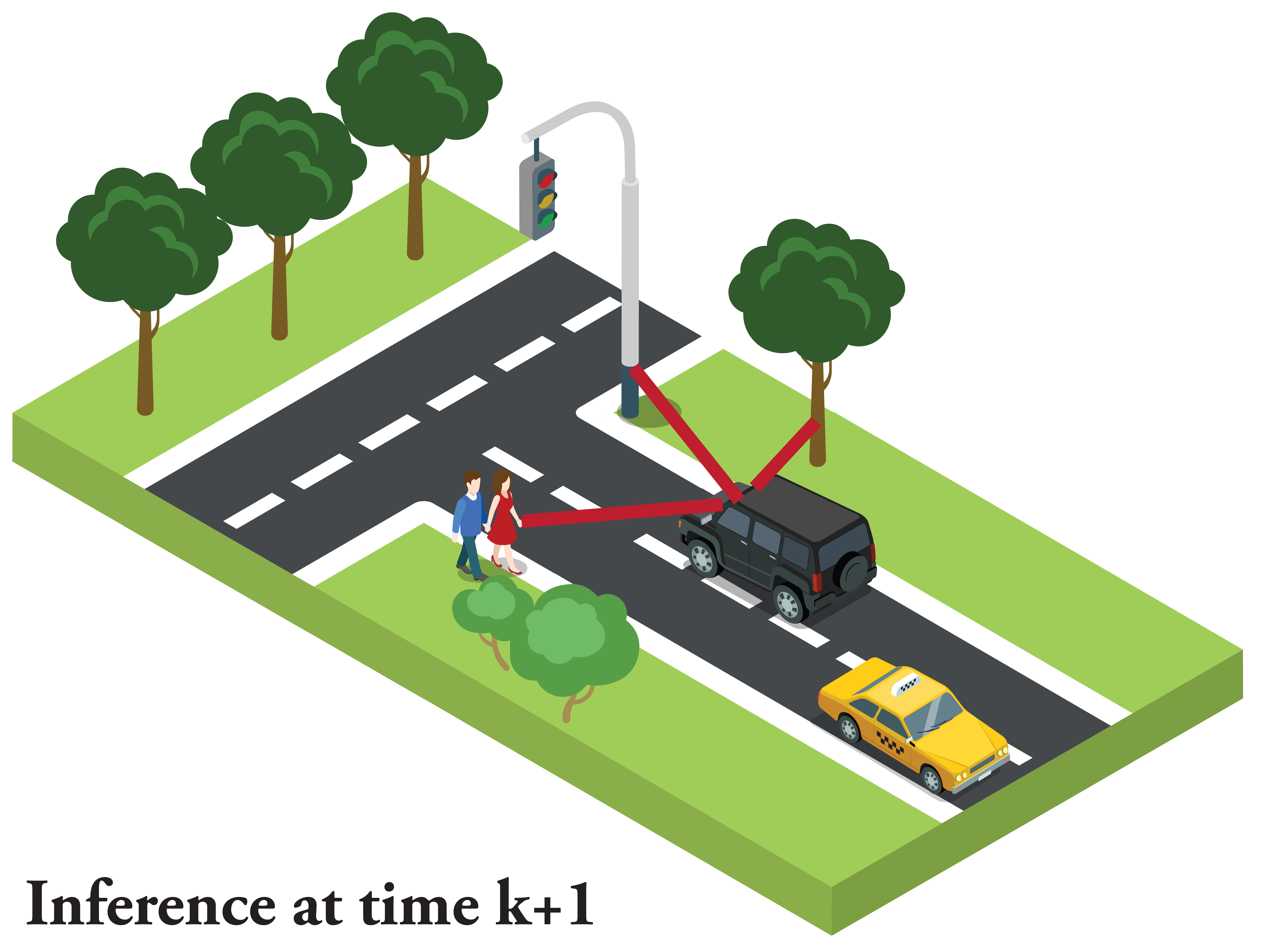}\label{fig:incDA:inf}}
        \caption{Illustration for inconsistent DA between planning and succeeding inference: (a)  at time $k$, our robot (i.e. the black jeep) plans three steps into the future. For the future step $k+1$ it predicts measurements from three landmarks (tree, traffic-light and taxi). (b) After executing the first action our robot obtained three measurements from the environment. Two of them (i.e. tree and traffic-light) match the predicted DA from precursory planning session, while the third is associated to a new landmark (i.e. the couple that came out of the taxi).}
        \label{fig:incDA}
\end{figure}

Updating inference with a precursory planning stage can be considered as a deviation from conventional Bayesian inference. Rather than updating the belief from the previous time instant with new incoming information (e.g. measurements), we propose to exploit the fact that similar calculations have already been performed within planning, in order to appropriately update the belief in inference more efficiently. We denote this novel approach by Re-Use BSP inference, or \texttt{RUB inference} in short.

The standard plan-act-infer framework of a typical autonomous system with conventional
Bayesian approach for inference update is presented in Figure~\ref{fig:HighLevel_Algo:std}. First, BSP determines the next best action(s) given the posterior belief at current time; the robot performs this action(s); information is gathered and the former belief from the precursory inference is  updated with new information (sensor measurements); the new posterior belief is then transferred back to the planning block in order to propagate it into future beliefs and provide again with the next action(s).

Our proposed concept, \texttt{RUB inference}, is presented in Figure~\ref{fig:HighLevel_Algo:ours}. 
\texttt{RUB inference} differs from the conventional Bayesian inference in two aspects: The output of the BSP process and the procedure of inference update. 
As opposed to standard Bayesian inference, in \texttt{RUB inference}, BSP output includes the next action(s) as well as the corresponding propagated future beliefs, no other changes are required in BSP in order to facilitate \texttt{RUB inference}. These beliefs are used to update inference while potentially taking care of data association aspects, rather than using the belief from precursory inference as conventionally done under Bayesian inference.
As can be seen in Figure~\ref{fig:HighLevel_Algo:ours}, the inference block contains data association (DA) update 
before the actual inference update. There are a lot of elements that can cause the DA in planning to be partially different than the DA established in the successive inference, e.g. estimation errors, disturbances, and dynamic or un-modeled unseen environments.

We start investigating this novel concept under a simplifying assumption that the DA considered in planning is consistent to that acquired during the succeeding inference, e.g. we predicted an association to a specific previously mapped landmark and later indeed observed that landmark. 
Since data association only relates to connections between variables and not to the measurement value, we are left with replacing the (potentially) incorrect measurement values, used within planning, with the actual values.
Under this assumption, we provide four exact methods to efficiently update inference using the belief calculated by the precursory planning phase. As will be seen, these methods provide the same estimation accuracy as the conventional Bayesian inference approach, with a significantly shorter computation time.

 We later relax the simplifying assumption mentioned above, and show inference can be efficiently updated using the precursory planning stage even when the DA considered in the two processes is partially different. Figure~\ref{fig:incDA} illustrates such a case of inconsistent DA using a simple navigation problem.
 At time $k$ our automated car (denoted by a black jeep), performs planning with a horizon of three steps. Figure~\ref{fig:incDA:plan} presents the chosen candidate action sequence along with the predicted measurements for future time $k+1$. Our automated car predicts that at future time $k+1$ it would obtain measurements from the tree, the traffic-light and the taxi from the opposite lane. In addition to association, these predicted measurements also have values (e.g.~pixels, distance) which depend on the state estimation (of both robot position and landmarks). Under an MPC framework, Figure~\ref{fig:incDA:inf} presents the succeeding inference for current time $k+1$, in which our automated car advanced a bit more than planned, and indeed obtained three measurements. Two of these measurements are to the tree and the traffic-light (i.e.~with consistent DA), while the third is to the couple that left the taxi (i.e.~inconsistent DA). 
In such a case, merely updating the measurement values will not resolve the difference between the aforementioned DAs; instead the DA should be updated to match the acquired data, before updating the measurement values. 
We provide a novel paradigm to update inconsistent DA, leveraging \texttt{iSAM2} graphical model based methodologies, thus setting the conditions for complete inference update via BSP regardless of DA consistency. 

To summarize, our contributions in this paper\footnote{A preliminary version of this paper appeared in \cite{Farhi17icra}.} are as follows: 
$(a)$ We introduce \texttt{RUB inference}, a novel approach for saving computation time during the inference stage by reusing calculations made during the precursory planning stage;
$(b)$ We provide four exact methods, that utilize our concept under the assumption of consistent DA. We evaluate these four methods and compare them to the state of the art in simulation. 
$(c)$ We provide a paradigm for incrementally updating inconsistent DA, thereby relaxing the afore-mentioned assumption; 
$(d)$ We evaluate our complete paradigm and compare it to the state of the art both in simulation and on real-world data, considering the problem of autonomous navigation in unknown environments. 

This paper is organized as follows. Section~\ref{sec:problem-formulation} formulates the discussed problem. Section~\ref{sec:approach} presents the suggested approach and its mathematical formulation. Section~\ref{sec:results} presents a thorough analysis of the suggested approach and a comparison to related work. Section~\ref{sec:motivation} discusses the broader perspective of \RUB. Section~\ref{sec:conclusions} captivates the conclusions of our work along with possible extensions and usage. 
To improve coherence, several aspects are covered in appendices.

\section{Background and Problem Formulation}
\label{sec:problem-formulation}
In this work, we consider the joint inference and belief space planning problem in a model predictive control (MPC) setting, i.e.~BSP is performed after each inference phase. This problem can be roughly divided into two successive and recursive stages, namely inference and planning. The former performs inference given all information up to current time, updating the belief over the state with incoming information (e.g. sensor measurements). The latter produces the next control action(s), given the belief from the former inference stage and a user defined objective function. 
\ref{eq:Q:Qnonzeros}

Let $x_{t}$ denote the robot's state at time instant $t$ and $\mathcal{L}$ represent the world state if the latter is uncertain or unknown. For example, for SLAM problem, it could represent objects or 3D landmarks. The joint state, up to time $k$, is defined as 
\begin{equation}
	X_{k}=\{x_{0},...,x_{k},\mathcal{L}\} \in \mathbb{R}^{n}.
\end{equation} 
We shall be using the notation ${t|k}$ to refer to some time instant $t$ while considering information up to time $k$; as will be shown in the sequel, this notation will allow to refer to \emph{sequential} inference and planning phases in a unified manner. 

Let $z_{t|k}$ and $u_{t|k}$ denote, respectively, the measurements and the applied control action at time $t$, while the current time is $k$. For example, $z_{k+1|k}$ represents measurements from a future time instant $k+1$ while $z_{k-1|k}$ represents measurements from a past time instant $k-1$, with the present time being $k$ in both cases. 
Representing the measurements and controls up to time $t$, given current time $k$, as
\begin{equation}
z_{1:t|k}\doteq \{z_{1|k},...,z_{t|k}\} \ , \ u_{0:t-1|k} \doteq \{u_{0|k},...,u_{t-1|k}\},
\end{equation}
the posterior probability density function (pdf) over the joint state, denoted as the \emph{belief}, is given by
\begin{equation}\label{eq:p_k} 
b[X_{t|k}] \doteq \prob{X_{t}|z_{1:t|k},u_{0:t-1|k}}.
\end{equation}
For $t\!=\!k$, Eq.~(\ref{eq:p_k}) represents the posterior at current time $k$, while for $t\!>\!k$ it represents planning stage posterior for a specific sequence of future actions and observations. 
Using Bayes rule, Eq.~(\ref{eq:p_k}) can be rewritten as
\begin{equation} \label{eq:factors}
\prob{X_{t}|z_{1:t|k},u_{0:t-1|k}} \propto \prob{x_{0}} \cdot
\prod_{i=1}^{t} \left[ \prob{x_{i}|x_{i-1},u_{i-1|k}} 
\prod_{j \in \mathcal{M}_{i|k}} \prob{z_{i|k}^{j}|x_{i},l_{j}} \right],
\end{equation}
where $\prob{X_{0}}$ is the prior on the initial joint state,  $\prob{x_{i}|x_{i-1},u_{i-1|k}}$ and $\prob{z_{i|k}^{j}|x_{i},l_{j}}$  denote, respectively,  the motion and measurement likelihood models. The set $\mathcal{M}_{i|k}$ contains all landmark indices observed at time $i$, i.e. it denotes data association (DA). The measurement of some landmark $j$ at time $i$ is denoted by $z_{i|k}^{j} \in z_{i|k}$. 
Under graphical representation of the belief, the conditional probabilities of the motion and observation models as well as the prior, can be denoted as factors (see~Appendix-B). 
Eq.~(\ref{eq:factors}) can also be represented by a multiplication of these factors
\begin{equation}\label{eq:fac_multiplication}
	\prob{X_{t}|z_{1:t|k},u_{0:t-1|k}} \propto  \prod^{t}_{i=0} \{f_{j}\}_{i|k}~,
\end{equation}
where $\{f_{j}\}_{i|k}$ represents all factors added at time $i$ while current time is $k$.
The motion and measurement models are conventionally modeled with additive  zero-mean Gaussian noise
\begin{eqnarray}
x_{i+1} &=& f(x_{i},u_{i})+w_{i} \quad , \quad  w_{i} \sim \mathcal{N}(0,\Sigma_w)\label{eq:motion_mod}
\\
z_{i}^{j} &=& h(x_{i},l_{j})+v_{i} \quad \ , \quad  v_{i} \sim \mathcal{N}(0,\Sigma_v),\label{eq:obs_mod}
\end{eqnarray}
where $f$ and $h$ are known possibly non-linear functions, $\Sigma_w$ and $\Sigma_v$ are the process and measurement noise covariance matrices respectively. 
 \subsection{Inference}\label{subsec:inference_def}
For the inference problem, $t \leq k$, i.e time instances that are equal or smaller than current time. The maximum a posteriori (MAP) estimate of the joint state $X_k$ for time $t = k$ is given by
\begin{equation}\label{eq:argPk} 
{X}_{k|k}^{\star} = \argmax_{X_{k}} \ b[X_{k|k}] 
= \argmax_{X_{k}} \ \prob{X_{k}|z_{1:k|k},u_{0:k-1|k}}.
 \end{equation}
For the Gaussian case, the MAP solution produces the first two moments of the belief through solving a Non-linear Least Squares (NLS) problem, as will be shown later on. The MAP estimate from Eq.~(\ref{eq:argPk}) is referred to as the \emph{inference solution} in which, all controls and observations until time instant $k$ are known.

\subsection{Planning in the Belief Space}\label{subsec:bsp_def}
As mentioned, the purpose of planning is to determine the next optimal action(s). Finite horizon belief space planning for $L$ look ahead steps involves inference over the beliefs 
\begin{equation}\label{eq:bsp_kl_def} 
b[X_{k+l|k}] =  \prob{X_{k+l}|z_{1:k+l|k},u_{0:k+l-1|k}}  \quad , \quad l\in[k+1,k+L]
\end{equation}
where we use the same notation as in Eq.~(\ref{eq:p_k}) to denote the current time is $k$.
The belief (\ref{eq:bsp_kl_def}) can be written recursively as a function of the belief $b[X_{k|k}]$ from the inference phase as 
\begin{equation}\label{eq:bsp_recursive} 
	b[X_{k+l|k}] = b[X_{k|k}] \cdot
	\prod_{i=k+1}^{k+l} \left[ \prob{x_{i}|x_{i-1},u_{i-1|k}}
	\prod_{j \in \mathcal{M}_{i|k}} \prob{z_{i|k}^{j}|x_{i},l_{j}} \right],
\end{equation}
for the considered action sequence $u_{k:k+l-1|k}$ at planning time $k$, and observations $z_{k+1:k+l|k}$ that are expected to be obtained upon execution of these actions. The set $\mathcal{M}_{i|k}$ denotes landmark indices that are expected to be observed at a future time instant $i$. 
It is worth stressing that the future belief (\ref{eq:bsp_recursive}) is determined by a specific realization of unknown future observations $z_{k+1:k+l|k}$, as stated in the belief definition in (\ref{eq:bsp_kl_def}). Since terms for future belief of the form $\prob{X_{k+l}|z_{1:k+l|k},u_{0:k+l-1|k}}$ will be used frequently in this paper in order not to burden the reader we use the more compact form  $b[X_{i|k}]$. Whenever $i > k$ the reader should consider the belief $b[X_{i|k}]$ as a function of a specific realization of future observations.

One can now define a general objective function
\begin{equation}\label{eq:cost}
	J(u_{k:k+L-1|k}) \! \doteq 
	\Expec_{z_{k+1:k+L|k}} \left[ \sum_{i=k+1}^{k+L} c_i \left( b[X_{i|k}],u_{i-1|k} \right)\right],
\end{equation}
with immediate costs (or rewards) $c_i$ and where the expectation considers all the possible realizations of the future observations $z_{k+1:k+L|k}$. Conceptually, one could also reason whether these observations will actually be obtained, e.g.~by considering also different realizations of $\mathcal{M}_{i|k}$. Note that for Gaussian distributions considered herein and information-theoretic costs (e.g.~entropy), it can be shown that the expectation operator can be omitted under maximum-likelihood observations assumption \citep{Indelman15ijrr}, while another alternative is to simulate future observations via sampling, e.g. \citep[Section~II-B]{Farhi19icra}, if such a simulator is available.
The optimal open-loop control can now be defined as 
\begin{equation}
u_{k:k+L-1|k}^{\star} = \argmin_{u_{k:k+L-1|k}} J(u_{k:k+L-1|k}).
\label{eq:OptimalControl}
\end{equation}
Evaluating the objective function (\ref{eq:cost}) for a candidate action sequence involves calculating belief evolution for the latter, i.e. solving the inference problem for each candidate action using predicted future associations and measurements. Note that since we consider an MPC framework, the optimal control is affectively not an open-loop control, since it is being recalculated at each single action step. 

\subsection{Problem Statement}\label{PS}
Our key observation is that inference and BSP share similar calculations.
Despite the similarities between them, they are treated as separate processes, thus duplicating costly calculations and increasing valuable computation time. This observation is impervious to any specific paradigms used for inference or planning and constitutes the difference between the use of \texttt{RUB inference} as opposed to conventional Bayesian inference. 

Our goal is to salvage valuable computation time in the inference update stage by exploiting the similarities between inference and precursory planning, thus without affecting solution accuracy or introducing new assumptions.

\section{Approach}
\label{sec:approach}
Calculating the next optimal action  $u^{\star}_{k|k} \in u^{\star}_{k:k+L-1|k}$  within BSP necessarily involves inference over the belief $b[X_{k+1|k}]$ conditioned on the same action $u^{\star}_{k|k}$. As we discuss in the sequel, this belief $b[X_{k+1|k}]$ can be different than $b[X_{k+1|k+1}]$ (the posterior at current time $k+1$) due to partially inconsistent data association and difference between measurement values considered in planning and those obtained in practice in inference. Our approach for \RUB, takes care of both of these aspects, thereby enabling to obtain $b[X_{k+1|k+1}]$ from $b[X_{k+1|k}]$. 
 
In the following, we first analyze the similarities between inference and BSP (Sections  \ref{subsec:inference} and \ref{subsec:bsp}), and use these insights in Section \ref{subsec:update_RHS} to  develop methods for inference update under a simplifying assumption of consistent DA. We then relax this assumption, by analyzing the possible scenarios for inconsistent DA between inference and precursory planning (Section \ref{ssubsec:DAtypes}), and deriving a method for updating inconsistent DA (Section \ref{ssubsec:updateDA}). 

It is worth stressing that the only thing needed to be changed in any BSP algorithm in order to support our paradigm for \RUB, is just adding more information to its output. More specifically, outputting not only the (sub)optimal action $u^{\star}_{k|k}$, but also the corresponding future belief $b[X_{k+1|k}]$ (e.g. the difference between Figures~\ref{fig:HighLevel_Algo:std} and \ref{fig:HighLevel_Algo:ours}).  

\subsection{Looking into Inference} \label{subsec:inference}
To better understand the similarities between inference and precursory planning, let us break down the inference solution to its components.
Introducing Eqs.~(\ref{eq:factors}-\ref{eq:obs_mod}) into Eq.~(\ref{eq:argPk}) and taking the negative logarithm yields the following non-linear least squares problem (NLS)
\begin{equation} \label{eq:mahal_factors}
X_{k|k}^{\star} = \argmin_{X_{k}} \Vert x_{0}-x_{0}^{\star}\Vert_{\Sigma_{0}}^{2} +
 \sum_{i=1}^{k}\!\! \left[ \Vert x_{i}\!-\!f(x_{i-1},u_{i-1|k})\Vert_{\Sigma_w}^{2}\!\!+\!\!\!\!\! \sum_{j \in \mathcal{M}_i|k}\!\!\!\!\! \Vert z_{i|k}^{j}\!\!-\!h(x_{i},l_{j}) \Vert_{\Sigma_v}^{2} \right]\! ,
\end{equation}
where $\Vert a \Vert_{\Sigma}^2\doteq a^T \Sigma^{-1} a$ is the squared Mahalanobis norm. 

Linearizing each of the terms in Eq.~(\ref{eq:mahal_factors}) and performing standard algebraic manipulations (see Appendix-A for derivation) yields
\begin{equation}\label{eq:Ax_b} 
\Delta X_{k|k}^{\star} =\argmin_{\Delta X_{k}}\Vert A_{k|k} \Delta X_{k}-b_{k|k}\Vert^{2} ,
\end{equation}
where $A_{k|k} \in \mathbb{R}^{m \times n}$ is the Jacobian matrix and $b_{k|k} \in \mathbb{R}^{m}$ is the right hand side (RHS) vector. In a more elaborated representation 
\begin{equation}\label{eq:matAx_vecb}
A_{k|k} =
\begin{bmatrix}
	\Sigma_{0}^{-\frac{1}{2}}\\
	\mathcal{F}_{1:k|k}\\
	\mathcal{H}_{1:k|k}
\end{bmatrix} 
\quad , \quad b_{k|k}= 
 \begin{bmatrix}
 	0\\
 	\breve{b}_{1:k|k}^{\mathcal{F}}\\
 	\breve{b}_{1:k|k}^{\mathcal{H}}
 \end{bmatrix},
\end{equation}
where $\mathcal{F}_{1:k|k}$, $\mathcal{H}_{1:k|k}$, $\breve{b}_{1:k|k}^{\mathcal{F}}$ and $\breve{b}_{1:k|k}^{\mathcal{H}}$ (see Appendix-A) denote the Jacobian matrices and RHS vectors of all motion and observation terms accordingly, for time instances $1\!:\!k$ when the current time is $k$.  These Jacobians, along with the corresponding RHS can be referred to by
\begin{equation}\label{eq:matAx_vecb_step2}
\mathcal{A}_{1:k|k} = \begin{bmatrix}
	\mathcal{F}_{1:k|k}\\
	\mathcal{H}_{1:k|k}
\end{bmatrix}
\quad,\quad  \breve{b}_{1:k|k} = 
 \begin{bmatrix}
 	\breve{b}_{1:k|k}^{\mathcal{F}}\\
 	\breve{b}_{1:k|k}^{\mathcal{H}}
 \end{bmatrix} ,
\end{equation}
 While there are a few methods to solve Eq.~(\ref{eq:Ax_b}),  we choose QR factorization as presented, e.g., in \cite{Kaess08tro}.
 The QR factorization of the Jacobian matrix $A_{k|k}$ is given by the orthonormal rotation matrix $Q_{k|k}$ and the upper triangular matrix $R_{k|k}$
\begin{equation} \label{eq:QR_A} 
A_{k|k} = Q_{k|k}R_{k|k}.
\end{equation}
Eq.~(\ref{eq:QR_A}) is introduced into Eq.~(\ref{eq:Ax_b}), thus producing
\begin{equation} \label{eq:QR_d} 
R_{k|k}\Delta X_{k} = d_{k|k},
\end{equation}
where $R_{k|k}$ is un upper triangular matrix and $d_{k|k}$ is the corresponding RHS vector, given by the original RHS vector and the orthonormal rotation matrix $Q_{k|k}$ 
\begin{equation} \label{eq:d_Qtb}
d_{k|k} \doteq Q_{k|k}^{T}b _{k|k}.
 \end{equation} 
We can now solve Eq.~(\ref{eq:QR_d}) for $\Delta X_{k}$ via back substitution, update the linearization point, and repeat the process until convergence. 
Eq.~(\ref{eq:QR_d}) can also be presented using a Bayes tree (BT) \citep{Kaess10tr}. A BT is a graphical representation of a factorized Jacobian matrix (the square root information matrix) $R$ and the corresponding RHS vector $d$, in the form of a directed tree. More on the formulation of inference using graphical models can be found in Appendix-B. One can substantially reduce running time by exploiting sparsity and updating the QR factorization from the previous step with new information instead of calculating a factorization from scratch, see e.g.~iSAM2 algorithm \citep{Kaess12ijrr}.

Given the inference solution, the belief $b[X_{k|k}]$ can be approximated by the Gaussian
\begin{equation}\label{eq:bsp_k_def}
 b[X_{k|k}] \doteq \prob{X_{k}|z_{1:k|k},u_{0:k-1|k}} = \mathcal{N}(X_{k|k}^{\star}, \Lambda_{k|k}^{-1}), 
 \end{equation}
while the information matrix is given by
\begin{equation} \label{eq:info_mat}
\Lambda_{k|k} = A_{k|k}^{T}A _{k|k}= R_{k|k}^{T}R _{k|k},
 \end{equation} 
and the factorized Jacobian matrix $R_{k|k}$ along with the corresponding RHS vector $d_{k|k}$ can be used to update the linearization point and to recover the MAP estimate. In other words, the factorized Jacobian matrix $R_{k|k}$ and the corresponding RHS vector $d_{k|k}$  are sufficient for performing a single iteration within Gaussian belief inference. 

\subsection{Looking into Planning} \label{subsec:bsp}
An interesting insight, that will be exploited in the sequel, is that the underlying equations of BSP are similar to those seen in Section \ref{subsec:inference}. 
In particular, evaluating the belief at the $L$th look ahead step, $b[X_{k+L|k}]$, involves MAP inference over a certain action sequence $u_{k:k+l-1|k}$ and future measurements $z_{k+1:k+l|k}$, which in turn, as in Section~\ref{subsec:inference}, can be described as an NLS problem
\begin{equation}\label{eq:mahal_factors_bsp}
X_{k+L|k}^{\star} = \argmin_{X_{k+L}} \Vert X_{k}-X_{k|k}^{\star}\Vert_{\Lambda_{k|k}^{-1}}^{2} +  
\sum_{i=k+1}^{k+L}\!\!\left[  \Vert x_{i}-f(x_{i-1},u_{i-1|k})\Vert_{\Sigma_w}^{2}\!\!\!+\!\!\!\!\!\sum_{j \in \mathcal{M}_i|k}\!\! \Vert z_{i|k}^{j}-h(x_{i},l_{j})\Vert_{\Sigma_v}^{2}\right] 
\end{equation}
For $i\! > \!k$,  the set $\mathcal{M}_{i|k}$ contains \emph{predicted} associations for future time instant $i$; hence, we can claim that $\forall i>k$ it is possible that $\mathcal{M}_{i|k} \neq \mathcal{M}_{i|i}$.
In other words, it is possible that associations from the planning stage, $\mathcal{M}_{k+1|k}$, would be partially different than the associations from the corresponding inference stage $\mathcal{M}_{k+1|k+1}$. Moreover, the likelihood for inconsistent DA between planning and the corresponding inference rises as we look further into the future, i.e.~with the distance $\Vert i-k \Vert$ increasing; e.g.~$\mathcal{M}_{k+j|k}$ and $\mathcal{M}_{k+j|k+j}$ are less likely to be identical for $j = 10$ than they are for $j=1$.

Predicting the unknown measurements $z_{k+1:k+L|k}$ in terms of both association and values can be done in various ways. In this paper the DA is predicted using current state estimation, and measurement values are obtained using the maximum-likelihood (ML) assumption, i.e. assuming zero innovation \citep{Dellaert06ijrr}. 
The robot pose is first propagated using the motion model~(\ref{eq:motion_mod}). 
All landmark estimations are then transformed to the robot's new camera frame. Once in the robot camera frame, all landmarks that are within the robot's field of view are considered to be seen by the robot (predicted DA).
The estimated position of each landmark, that is considered as visible by the robot, is being projected to the camera image plane \citep{Hartley04book}, thus generating measurements.
It is worth mentioning that the aforementioned methodology is not able to predict occurrences of new landmarks, since it is based solely on the map the robot built thus far, i.e.~current joint state estimation. The ability to predict occurrences of new landmarks would increase the advantage of \RUB over conventional Bayesian inference (as discussed in the sequel), hence is left for future work. 

Once the predicted measurements are acquired, by following a similar procedure to the one presented in Section \ref{subsec:inference}, for each action sequence we get
\begin{equation}
\label{eq:Ax_b_bsp} 
\Delta X_{k+L|k}^{\star} = \argmin_{\Delta X_{k+L}}\Vert A_{k+L|k} \Delta X_{k+L}-b_{k+L|k} \Vert^{2}.
\end{equation}
The Jacobian matrix $A_{k+L|k}$ and RHS vector $b_{k+L|k}$ are defined as
\begin{equation}\label{eq:matAx_vecb_bsp}
A_{k+L|k} \doteq \begin{bmatrix}
A_{k|k}\\
\mathcal{A}_{k+1:k+L|k}\\
\end{bmatrix} 
~ ,~
b_{k+L|k} \doteq \begin{bmatrix}
b_{k|k}\\
\breve{b}_{k+1:k+L|k}\\
\end{bmatrix}, 
\end{equation}
where $A_{k|k}$ and $b_{k|k}$ are taken from inference, see Eq.~(\ref{eq:Ax_b}), and $\mathcal{A}_{k+1:k+L|k}$ and $\breve{b}_{k+1:k+L|k}$ correspond to the new terms obtained at the first $L$ look ahead steps (e.g. see Eq.~(\ref{eq:matAx_vecb_step2})). 
Note that although $\mathcal{A}_{k+1:k+L|k}$ is not a function of the (unknown) measurements $z_{k+1:k+L|k}$, it is a function of the predicted DA, $\mathcal{M}_{k+1:k+L|k}$ \citep{Indelman15ijrr}. 
Performing QR factorization, yields
\begin{equation}\label{eq:QR_A_bsp}
	A_{k+L|k} = Q^A_{k+L|k}R_{k+L|k}, 	
\end{equation}
from which the information matrix, required in the information-theoretic cost, can be calculated. Using Eq.~(\ref{eq:matAx_vecb_bsp}) the belief that correlates to the specific action sequence can be estimated, enabling evaluating the objective function (\ref{eq:cost}). Determining the best action via Eq.~(\ref{eq:OptimalControl}) involves repeating this process for different candidate actions.

\subsection{Similarities between Inference and BSP}\label{subsec:similarities}
In an MPC setting, only the first action from the sequence $u^{\star}_{k:k+L-1|k}$ is executed, i.e.
\begin{equation}\label{eq:performed_action}
	u_{k|k+1} = u^{\star}_{k|k} \in u^{\star}_{k:k+L-1|k}.	
\end{equation}
In such case the difference between the belief obtained from BSP (for action $u^{\star}_{k|k}$)
\begin{equation}\label{eq:belief_planning}
	b[X_{k+1|k}] \equiv \prob{X_{k+1}|z_{1:k|k},u_{0:k-1|k},z_{k+1|k},u^{\star}_{k|k}} ,
\end{equation}
 and the belief from the succeeding inference 
 \begin{equation}\label{eq:belief_inf}
	b[X_{k+1|k+1}] \equiv \prob{X_{k+1}|z_{1:k|k},u_{0:k-1|k},z_{k+1|k+1},u_{k|k+1}} ,
\end{equation}
is rooted in the set of measurements (i.e.~$z_{k+1|k+1}$ vs.~$z_{k+1|k}$), and the corresponding factors added at time instant $k+1$. 
These factor sets, denoted by $\{ f_i\}_{k+1|k}$ and $\{ f_j\}_{k+1|k+1}$ accordingly, can differ from one another in data association and measurement values. 
Since solving the belief requires linearization (\ref{eq:Ax_b}), it is important to note that both beliefs, $b[X_{k+1|k}]$ and $b[X_{k+1|k+1}]$, make use of the \emph{same} initial linearization point $\bar{X}_{k+1}$ for the common variables. 
In particular, as in this work we do not reason within planning about new, unmapped thus far, landmarks, it follows that 
\begin{equation}
X_{k+1|k} = 
\begin{bmatrix}
X_{k|k}\\
x_{k+1}\\
\end{bmatrix} 
\ \ , \ \ 
X_{k+1|k+1} = 
\begin{bmatrix}
X_{k|k}\\
x_{k+1}\\
L_{k+1}^{new}
\end{bmatrix} 
\end{equation}
where $L_{k+1}^{new}$ represents the new landmarks that were added to the belief for the first time at time instant $k+1$. The linearization point for the common variables is $[X^{\star}_{k|k} \ , \ f(x_k,u_{k|k}^{\star})]$ for planning, and $[X^{\star}_{k|k} \ , \ f(x_k,u_{k|k+1})]$ for succeeding inference, where $f(.)$ is the motion model (\ref{eq:motion_mod}). Since the (sub)optimal action provided by BSP is the one executed in the succeeding inference i.e. Eq.~(\ref{eq:performed_action}), the motion models are identical hence the same linearization point is used in both inference and precursory planning.

When considering the belief from planning (\ref{eq:belief_planning}), which is propagated with the next action (\ref{eq:performed_action}) and predicted measurements, with the previously factorized form of $A_{k|k}$ and $b_{k|k}$, we get 
\begin{equation}\label{eq:A_k1_k_stripdown}
A^{R}_{k+1|k} \doteq \begin{bmatrix}
R_{k|k}\\
\mathcal{A}_{k+1|k}\\
\end{bmatrix} 
~ ,~
b^{d}_{k+1|k} \doteq \begin{bmatrix}
d_{k|k}\\
\breve{b}_{k+1|k}\\
\end{bmatrix}. 
\end{equation}
Similarly, when considering the a posteriori belief from inference (\ref{eq:belief_inf}), propagated with the next action (\ref{eq:performed_action}) and acquired measurements, with the previously factorized form of $A_{k|k}$ and $b_{k|k}$, we get
\begin{equation}\label{eq:A_k1_k1_stripdown}
A^{R}_{k+1|k1} \doteq \begin{bmatrix}
R_{k|k}\\ 
\mathcal{A}_{k+1|k+1}\\
\end{bmatrix} 
~ ,~
b^{d}_{k+1|k+1} \doteq \begin{bmatrix}
d_{k|k}\\
\breve{b}_{k+1|k+1}\\
\end{bmatrix}. 
\end{equation}
For the same action (\ref{eq:performed_action}), the difference between Eq.~(\ref{eq:A_k1_k_stripdown}) to the equivalent representation of standard Bayesian inference (\ref{eq:A_k1_k1_stripdown}) originates from the factors added at time $k+1$ 
\begin{equation}\label{eq:match_Jac}
	\mathcal{A}_{k+1|k} \stackrel{?}{=} \mathcal{A}_{k+1|k+1}~,
\end{equation}
\begin{equation}\label{eq:match_RHS}
	\breve{b}_{k+1|k} \stackrel{?}{=} \breve{b}_{k+1|k+1}~.
\end{equation}
Since the aforementioned share the same action sequence, the same linearization point and the same models, the differences remain limited to the DA and measurement values at time $k+1$. 

In planning, DA is based on predicting which landmarks would be observed. This DA could very possibly be different than the actual landmarks the robot observes, as presented in Sec.~\ref{subsec:bsp}. This inconsistency in DA manifests in both the Jacobian matrices and the RHS vectors. Even in case of consistent DA, the predicted measurements (if exist) would still be different than the actual measurements due to various reasons, e.g.~the predicted position is different than the ground truth of the robot, measurement noise, inaccurate models. 

While for consistent DA and the same linearization point Eq.~(\ref{eq:match_Jac}) will always be true, the RHS vectors, specifically Eq.~(\ref{eq:match_RHS}), would still be different due to the difference in measurement values considered in planning and actually obtained in inference. 

It is worth stressing that consistent data association between inference and precursory planning suggests that all predictions for state variable (new or existing) associations were in fact true. In addition to the new robot state added each time instant, new variables could also manifest in the form of landmarks. 
Consistent DA implies that the future appearance of all new landmarks has been perfectly predicted during planning. Since for the purpose of this work, we use a simple prediction mechanism unable to predict new landmarks (see Section~\ref{subsec:bsp}), consistent DA would inevitably mean no new landmarks in inference, i.e~$L_{k+1}^{new}$ is an empty set.

We start developing our method by assuming consistent DA between inference and precursory planning. In such a case the difference is limited to the RHS vectors. Later we relax this assumption by dealing with possible DA inconsistency prior to the update of the RHS vector, thus addressing the general and complete problem of inference update using \RUB paradigm.

\subsection{Inference Update from BSP assuming Consistent Data Association} \label{subsec:update_RHS}

Let us assume that the DA between inference and precursory planning is consistent, whether the cause is a "lucky guess" during planning or whether the DA inconsistency has been resolved beforehand.
Recalling the definition of $\mathcal{M}_{i|k}$ (see e.g.~Eq.~(\ref{eq:bsp_recursive})), this assumption is equivalent to writing
\begin{equation}\label{eq:DA_assumption}
	\mathcal{M}_{k+1|k}\equiv\mathcal{M}_{k+1|k+1}.
\end{equation}
In other words, landmarks considered to be observed at a future time $k+1$, will indeed be observed at that time. Note this does \emph{not} necessarily imply that actual measurements and robot poses will be as considered within the planning stage, but it does necessarily  state that both are considering the same variables and the same associations.

We now observe that the motion models in both $b[X_{k+1|k+1}]$ and $b[X_{k+1|k}]$ are evaluated considering the \emph{same} control (i.e.~the optimal control $u^{\star}_{k}$). Moreover, the robot pose $x_{k+1}$ is initialized to the \emph{same} value in both cases as $f(x_k,u^{\star}_k)$, see e.g.~Eq.(27) in \cite{Indelman15ijrr}, and thus the linearization point of all probabilistic terms in inference and planning is \emph{identical}. This, together with the aforementioned assumption (i.e.~Eq.~(\ref{eq:DA_assumption}) holds) allows us to write $A_{k+1|k} = A_{k+1|k+1}$, and hence
\begin{equation}\label{eq:DA_assumption_R}
	R_{k+1|k+1} \equiv R_{k+1|k},
\end{equation}
for the \emph{first iteration} in the inference stage at time $k+1$. 
Hence, in order to solve $b[X_{k+1|k+1}]$ we are left to find the RHS vector $d_{k+1|k+1}$, while $R_{k+1|k+1}$ can be \emph{entirely re-used}.

\begin{table}
	\caption{Notations for Section~\ref{subsec:update_RHS}} 
	\centering 
	\begin{tabular}{c c} 
		\hline\hline 
		\textbf{Variable} & \textbf{Description}  \\ [0.5ex] 
		\hline 
		$ \Box_{t|k}$ & Of time $t$ while current time is $k$ \\[1ex]
		$\Delta X_{k}$ & State perturbation around linearization point \\[1ex]
		$\mathcal{M}_{t|k}$ & Data Association at time $t$ \\[1ex]
		$A_{t|k}$ & Jacobian matrix at time $t$ \\[1ex]
		$b_{t|k}$ & RHS vector at time $t$ \\[1ex]
		$\mathcal{A}_{t|k}$ & Jacobian part related to all factors added at time $t$ \\[1ex]
		$\mathcal{F}_{t|k}$ & Jacobian part related to motion factor added at time $t$ \\ [1ex]
		$\mathcal{H}_{t|k}$ & Jacobian part related to all factors added at time $t$ without the motion factor \\[1ex]
		$\breve{b}_{t|k}$ & RHS vector related to all factors added at time $t$ \\ [1ex]
		$\breve{b}^{\mathcal{F}}_{t|k}$ & RHS vector related to motion factor at time $t$ \\ [1ex]
		$\breve{b}^{\mathcal{H}}_{t|k}$ & RHS vector related to all factors added at time $t$ without the motion factor\\ [1ex]
		$R_{t|k}$ & Factorized Jacobian, i.e. square root information matrix \\ [1ex]
		$d_{t|k}$ & Factorized RHS vector \\[1ex]
		$A^R_{t|k}$ & Factorized $ \left[R^T_{t-1|k} ,\mathcal{A}^T_{t|k} \right]^T$ \\[1ex]
		$R^{\mathcal{F}}_{t|k}$ & Factorized $ \left[R^T_{t-1|k} ,{\mathcal{F}}^T_{t|k} \right]^T$   \\ [1ex]
		$d^{\mathcal{F}}_{t|k}$ & Factorized $ \left[d^T_{t-1|k} , {\breve{b}^{\mathcal{F}T}}_{t|k} \right]^T$ \\ [1ex]
		$R^{aug}_{t|k}$ & Factorized Jacobian at time $t-1$ zero padded to match factorized Jacobian at time $t$\\ [1ex]
		$d^{aug}_{t|k}$ & Factorized RHS vector at time $t-1$ zero padded to natch factorize RHS vector at time $t$ \\ [1ex]
		$Q^A_{t|k}$ & Rotation matrix for factorizing $A_{t|k}$ into $R_{t|k}$ \\[1ex]
		$Q_{t|k}$ & Rotation matrix for factorizing $A^R_{t|k}$ into $R_{t|k}$ \\[1ex]
		$Q^{\mathcal{F}}_{t|k}$ & Rotation matrix for factorizing $ \left[R^T_{t-1|k} ,{\mathcal{F}}^T_{t|k} \right]^T$ into $R^{\mathcal{F}}_{t|k}$ \\ [1ex] 
		$Q^{\mathcal{H}}_{t|k}$ & Rotation matrix for factorizing $ \left[{R^\mathcal{F}}^T_{t|k} ,\mathcal{H}^T_{t|k} \right]^T$ into $R_{t|k}$ \\ [1ex] 
		\hline\hline
	\end{tabular}
	\label{table:RHSupdate} 
\end{table}

In the sequel we present four methods that can be used for updating the RHS vector, and examine computational aspects of each. The four methods use two different approaches to update the RHS vector: while the first two (\texttt{OTM} and \texttt{OTM-OO}), utilize the rotation matrix available from factorization, the last two (\texttt{DU} and \texttt{DU-OO}) utilize information downdate / update principles. After we review the methods we  shortly discuss the advantages and disadvantages of each (Sec.~\ref{ssubsec:discussionMethods}).
It is worth stressing that each of these methods results in the same RHS vector which is also identical to the RHS vector that would have been obtained by the standard inference update. With both the factorized Jacobian matrix (i.e. R) and the RHS vector identical to the standard inference update approach, RUB inference provides  the same estimation accuracy for the inference solution.  

\subsubsection{The Orthogonal Transformation Matrix Method - \texttt{OTM}} \label{subsubsec:otm}

In the \texttt{OTM} method, we obtain $d_{k+1|k+1}$  following the definition as written in Eq.~(\ref{eq:d_Qtb}). Recall that at time $k+1$ in the inference stage, the posterior should be updated with new terms that correspond, for example, to motion model and obtained measurements. The RHS vector's augmentation, that corresponds to these new terms is denoted by $\breve{b}_{k+1|k+1}$, see Eq.~(\ref{eq:matAx_vecb_step2}). Given $R_{k|k}$ and $d_{k|k}$ from the inference stage at time $k$, the augmented system at time $k+1$ is
\begin{equation}
A^{R}_{k+1|k+1}\Delta X_{k+1}\doteq \begin{bmatrix}
R_{k|k}\\
\mathcal{A}_{k+1|k+1}
\end{bmatrix}
\Delta X_{k+1} = 
\begin{bmatrix}
d_{k|k}\\
\breve{b}_{k+1|k+1} 
\end{bmatrix} 
\end{equation}
which after factorization of $A^{R}_{k+1|k+1}$ (see Eqs.~(\ref{eq:QR_A})-(\ref{eq:d_Qtb})) becomes
\begin{equation}
R_{k+1|k+1} \Delta X_{k+1} = d_{k+1|k+1},
\end{equation}
where 
\begin{equation}\label{eq:dk1_Qk1dkbk1}
d_{k+1|k+1}= Q_{k+1|k+1}^{T}
\begin{bmatrix}
d_{k|k} \\
\breve{b}_{k+1|k+1}
\end{bmatrix}.
\end{equation} 
As deduced from Eq.~(\ref{eq:dk1_Qk1dkbk1}), the calculation of $d_{k+1|k+1}$ requires $Q_{k+1|k+1}$ . Since $A^{R}_{k+1|k}\equiv A^{R}_{k+1|k+1}$ (see Section \ref{subsec:update_RHS}), we get $Q_{k+1|k+1}=Q_{k+1|k}$. However, $Q_{k+1|k}$, is already available from the precursory planning stage, see Eq.~(\ref{eq:QR_A_bsp}), and thus calculating $d_{k+1|k+1}$ via Eq.~(\ref{eq:dk1_Qk1dkbk1}) does \emph{not} involve QR factorization in practice.
To summarize, under the \texttt{OTM} method we obtain the RHS vector $d_{k+1|k+1}$ in the following manner:
\begin{equation}\label{eq:get_dk1k1_otm}
	d_{k+1|k+1}= Q_{k+1|k}^{T}
	\begin{bmatrix}
		d_{k|k} \\
		\breve{b}_{k+1|k+1}
	\end{bmatrix}.
\end{equation}
where $Q_{k+1|k}^{T}$ is available from the factorization of precursory planning, $d_{k|k}$ is the RHS from inference at time $k$, and $\breve{b}_{k+1|k+1}$ are the new un-factorized RHS values obtained at time $k+1$.

\subsubsection{The OTM - Only Observations Method - \texttt{OTM-OO}}\label{subsubsec:otmoo}

The \texttt{OTM-OO} method is a variant of the \texttt{OTM} method. \texttt{OTM-OO} aspires to utilize even more information from the planning stage. Since the motion models from inference and the precursory planning first step are identical, i.e.~same function $f(.,.)$, see Eqs.~(\ref{eq:mahal_factors}) and (\ref{eq:mahal_factors_bsp}), and as in both cases the \emph{same} control is considered - Eq.~(\ref{eq:performed_action}), there is no reason to change the motion model data from the RHS vector $d_{k+1|k}$. In order to enable the aforementioned, we require the matching rotation matrix. One way would be to break down the planning stage as described in Section~\ref{subsec:bsp} into two stages, in which the motion and observation models are updated separately. Usually this breakdown is performed either way since a propagated future pose is required for predicting future measurements.
 
So following Section \ref{subsubsec:otm}, instead of using $d_{k|k}$, we attain from planning the RHS vector already with the motion model ($d_{k+1|k}^{\mathcal{F}}$), augment it with the new measurements and rotate it with the corresponding rotation matrix obtained from the planning stage
 \begin{equation}\label{eq:dk1_Qk1dkbk1_meth3}
	d_{k+1|k+1}= Q_{k+1|k}^{\mathcal{H}^{T}}
	\begin{bmatrix}
    	d_{k+1|k}^{\mathcal{F}} \\
		\breve{b}_{k+1|k+1}^{\mathcal{H}}
	\end{bmatrix}. 
\end{equation} 
The rotation matrix $Q_{k+1|k}^{\mathcal{H}}$ is given from the precursory planning stage where
\begin{equation}\label{eq:QR_RkAk1mot}
	Q_{k+1|k}^{\mathcal{H}} R_{k+1|k}= 
	\begin{bmatrix}
    	R_{k+1|k}^{\mathcal{F}} \\
		\mathcal{H}_{k+1|k}
	\end{bmatrix},   
\end{equation}
and where $R^{\mathcal{F}}_{k+1|k}$ is the factorized Jacobian propagated with the motion model given by
\begin{equation} 
	Q_{k+1|k}^{\mathcal{F}}R_{k+1|k}^{\mathcal{F}}= 
	\begin{bmatrix}
    	R_{k|k}\\
		\mathcal{F}_{k+1|k}
	\end{bmatrix}.   
\end{equation}
As will be seen later on, the OTM-OO method would prove to be the most computationally efficient between the four suggested methods.
 
\subsubsection{The Downdate Update Method - \texttt{DU}}\label{subsubsec:du}

In the \texttt{DU} method we propose to re-use the $d_{k+1|k}$ vector from the planning stage to calculate $d_{k+1|k+1}$. 

While not necessarily required within the planning stage, $d_{k+1|k}$ could be calculated at that stage from $b_{k+1|k}$ and $Q_{k+1|k}$, see Eqs.~(\ref{eq:matAx_vecb_bsp})-(\ref{eq:QR_A_bsp}). However, $b_{k+1|k}$ (unlike $A_{k+1|k}$) is a function of the unknown future observations $z_{k+1|k}$, which would seem to complicate things. Our solution to this issue is as follows: We assume \emph{some} value for the observations $z_{k+1|k}$ and then calculate $d_{k+1|k}$ within the planning stage. As in inference at time $k+1$, the actual measurements $z_{k+1|k+1}$ will be different, we remove the contribution of $z_{k+1|k}$ to $d_{k+1|k}$ via information downdating \citep[Sec.~V-A]{Cunningham13icra}, and then appropriately incorporate $z_{k+1|k+1}$ to get $d_{k+1|k+1}$ using the same mechanism.

More specifically, downdating the measurements $z_{k+1|k}$ from $d_{k+1|k}$ is done via \citep[Sec.~V-A]{Cunningham13icra}
\begin{equation}\label{eq:downdate_dk1k2}
d_{k+1|k}^{aug} = R_{k+1|k}^{aug^{-T}}(R_{k+1|k}^{T}d_{k+1|k}-\mathcal{A}_{k+1|k}^{T}\breve{b}_{k+1|k}),
\end{equation}
where $\breve{b}_{k+1|k}$ is a function of  $z_{k+1|k}$, see Eqs.~(\ref{eq:mahal_factors_bsp})-(\ref{eq:matAx_vecb_bsp}), and where $R_{k+1|k}^{aug}$ is the downdated $R_{k+1|k}$ matrix which is given by
\begin{equation}\label{eq:downdate_Rk1k2}
	R_{k+1|k}^{aug^{T}}R_{k+1|k}^{aug} = A^{R^{T}}_{k+1|k}A^{R}_{k+1|k}-\mathcal{A}_{k+1|k}^{T}\mathcal{A}_{k+1|k}.
\end{equation}
Interestingly, the above calculations are not really required: Since we already have $d_{k|k}$ from the previous inference stage, we can attain the downdated $d_{k+1|k}^{aug}$ vector more efficiently by augmenting $d_{k|k}$ with zero padding. 
\begin{equation}\label{eq:downdate_dk1k2_short}
d_{k+1|k}^{aug} = 
\begin{bmatrix}
	d_{k|k} \\
	0 
\end{bmatrix}	
\end{equation}
where $d_{k+1|k}^{aug}$ is the downdated RHS vector and $0$ is a zero padding to match dimensions. 
Similarly, $R_{k+1|k}^{aug}$ can be calculated as 

\begin{equation}\label{eq:downdate_Rk1k2_short}
	R_{k+1|k}^{aug} = 
	\begin{bmatrix}
	R_{k|k} & 0 \\
	0 & 0 
\end{bmatrix},
\end{equation}
where $R_{k|k}$ is zero padded to match dimensions of $R_{k+1|k}$ . 

Now, all which is left to get $d_{k+1|k+1}$, is to incorporate the new measurements $z_{k+1|k+1}$ (encoded in $\breve{b}_{k+1|k+1}$). We utilize the information downdating mechanism in \citep[Sec.~V-A]{Cunningham13icra}, in order to update information. Intuitively, instead of downdating information from $d_{k+1|k}$, we would like to add information to $d_{k+1|k}^{aug}$. So by appropriately adjusting Eq.(\ref{eq:downdate_dk1k2}) this can be done via 
\begin{equation}\label{eq:update_dk1k2}
	d_{k+1|k+1} = R_{k+1|k+1}^{-T}(R_{k+1|k}^{aug^{T}}d_{k+1|k}^{aug}+\mathcal{A}_{k+1|k+1}^{T}\breve{b}_{k+1|k+1}),
\end{equation}
where according to Eq.~(\ref{eq:DA_assumption}) $R_{k+1|k+1} \equiv R_{k+1|k}$ and $\mathcal{A}_{k+1|k+1}\equiv \mathcal{A}_{k+1|k}$, $R_{k+1|k}^{aug}$ is given by Eq.(\ref{eq:downdate_Rk1k2_short}), $d_{k+1|k}^{aug}$ is given by Eq.(\ref{eq:downdate_dk1k2_short}), and $\breve{b}_{k+1|k+1}$ are the new un-factorized RHS values obtained at time $k+1$.

To summarize, under the \texttt{DU} method we obtain the RHS vector $d_{k+1|k+1}$ in the following manner:
\begin{equation}\label{eq:get_dk1k1_du}
	d_{k+1|k+1} = R_{k+1|k}^{-T} \left( \begin{bmatrix}
	R_{k|k} & 0 \\
	0 & 0 
\end{bmatrix} 
	^{T} \begin{bmatrix}
	d_{k|k} \\
	0 
\end{bmatrix} + \mathcal{A}_{k+1|k}^{T}\breve{b}_{k+1|k+1} \right).
\end{equation}

\subsubsection{The DU - Only Observations Method - \texttt{DU-OO}}\label{subsubsec:duoo}

The \texttt{DU-OO} method is a variant of the DU method, where, similarly to Section \ref{subsubsec:otmoo}, we utilize the fact that there is no reason to change the motion model data from the RHS vector $d_{k+1|k}$. Hence we would downdate all data with the exception of the motion model, and then update accordingly. As opposed to Section \ref{subsubsec:du}, now we do need to downdate using \cite[Sec.~V-A]{Cunningham13icra} 
\begin{equation}\label{eq:downdate_dk1k4}
	d_{k+1|k}^{\mathcal{F}} = R_{k+1|k}^{\mathcal{F}^{-T}}(R_{k+1|k}^{T}d_{k+1|k}-\mathcal{H}_{k+1|k}^{T}\breve{b}_{k+1|k}^{\mathcal{H}}),
	\end{equation}
where $d_{k+1|k}^{\mathcal{F}}$ is the RHS vector, downdated from all predicted measurements with the exception of the motion model, and $R_{k+1|k}^{\mathcal{F}}$ is the equivalent downdated $R_{k+1|k}$ matrix which is given by
\begin{equation}\label{eq:downdate_Rk1k4}
	R_{k+1|k}^{\mathcal{F}^{T}}R_{k+1|k}^{\mathcal{F}} = A^{R^{T}}_{k+1|k}A^{R}_{k+1|k}-\mathcal{H}_{k+1|k}^{T}\mathcal{H}_{k+1|k},
\end{equation}
where $\mathcal{H}_{k+1|k}$ denotes the portion of the planning stage Jacobian,  of the predicted factors with the exception of the motion model. Now, all which is left, is to update $d_{k+1|k}^{\mathcal{F}}$ with the new measurements from the inference stage
\begin{equation}\label{eq:update_dk1k4}
	d_{k+1|k+1} = R_{k+1|k+1}^{-T}(R_{k+1|k}^{\mathcal{F}^{T}}d_{k+1|k}^{\mathcal{F}}+\mathcal{H}_{k+1|k+1}^{T}\breve{b}_{k+1|k+1}^{\mathcal{H}}),
	\end{equation}
where according to Eq.~(\ref{eq:DA_assumption}) $R_{k+1|k+1} \equiv R_{k+1|k}$ and $\mathcal{H}_{k+1|k+1}\equiv \mathcal{H}_{k+1|k}$, $R_{k+1|k}^{\mathcal{F}}$ is given by Eq.(\ref{eq:downdate_Rk1k4}), $d_{k+1|k}^{\mathcal{F}}$ is given by Eq.(\ref{eq:downdate_dk1k4}), and $\breve{b}_{k+1|k+1}$ are the new un-factorized RHS values obtained at time $k+1$.

By introducing Eq.~(\ref{eq:downdate_dk1k4}) into Eq.(\ref{eq:update_dk1k4}) we can also avert from calculating $R_{k+1|k}^{\mathcal{F}}$ so under the \texttt{DU-OO} assumption we obtain the RHS vector $d_{k+1|k+1}$ in the following manner:

\begin{equation}
	d_{k+1|k+1} = R_{k+1|k}^{-T}\left( R_{k+1|k}^{T}d_{k+1|k} + \mathcal{H}_{k+1|k}^{T} \left(\breve{b}_{k+1|k+1}^{\mathcal{H}} - \breve{b}_{k+1|k}^{\mathcal{H}} \right) \right),
	\end{equation}
which can be rewritten as
\begin{equation}\label{eq:get_dk1k1_duoo}
	d_{k+1|k+1} = d_{k+1|k} + R_{k+1|k}^{-T}\mathcal{H}_{k+1|k}^{T} \left(\breve{b}_{k+1|k+1}^{\mathcal{H}} - \breve{b}_{k+1|k}^{\mathcal{H}} \right).
	\end{equation}

\subsubsection{Discussion - RHS update Methods}\label{ssubsec:discussionMethods}
\begin{figure}
	\centering
        \includegraphics[bb={0 0 0 0},trim={0 0 0 0},clip, width=0.5\textwidth]{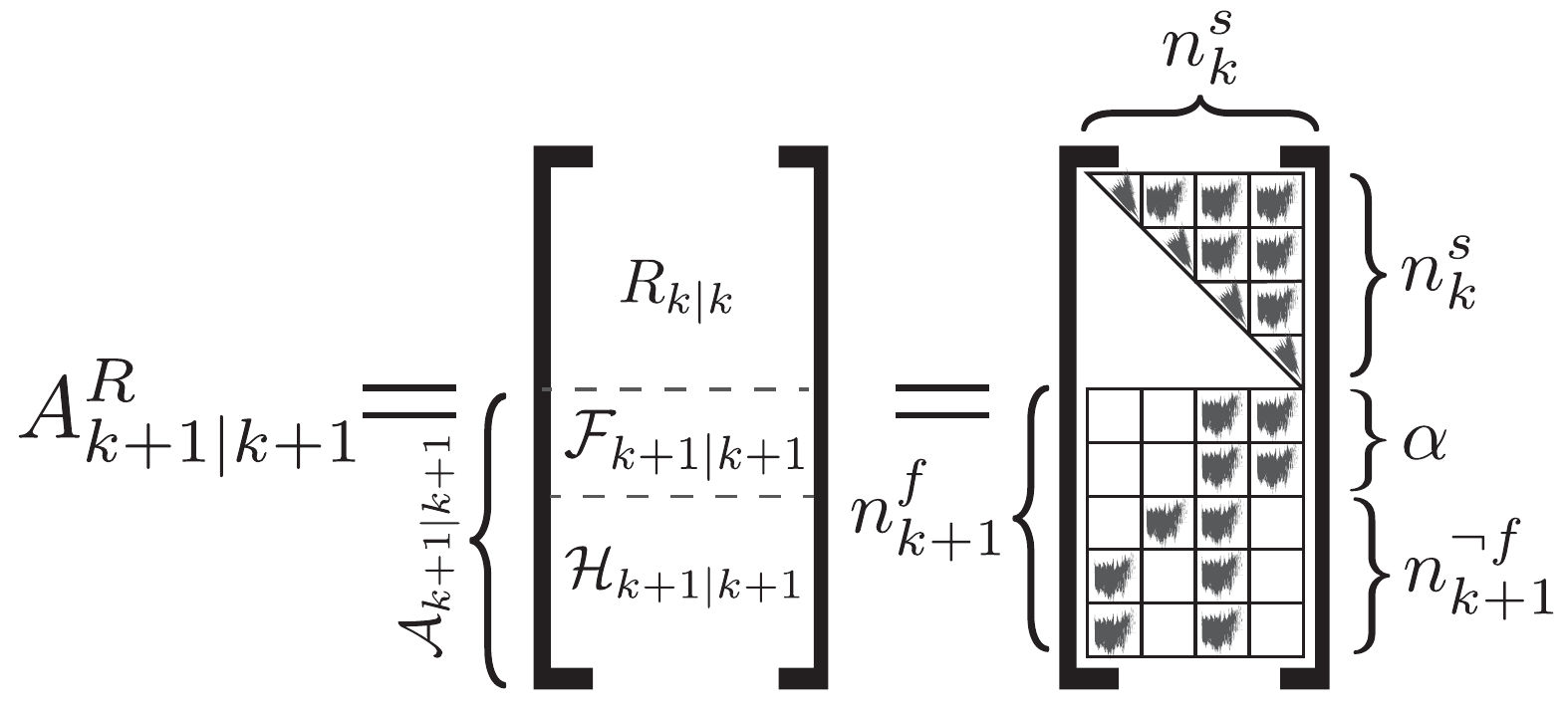}
                \caption{Illustration of the Jacobian matrix $A^R_{k+1|k+1}$ introduced in Eq.~(\ref{eq:A_k1_k1_stripdown}), on its components and dimensions. These notations are used along Section~\ref{ssubsec:discussionMethods}, and brought here for the reader's convenience.}
        \label{fig:Q:param}
\end{figure}
In this section we would like to give the reader some intuition regarding the advantages and disadvantages of the \texttt{OTM} approach when compared to the \texttt{DU} approach. Since both provide the same desired solution, the difference between them would manifest in computation time and ease of use. In the sequel we cover both starting with the complexity of each.

Let us compare the complexity required for updating the RHS by \texttt{OTM}, see Eq.~(\ref{eq:get_dk1k1_otm}), against the complexity required for updating the RHS by \texttt{DU}, see Eq.~(\ref{eq:get_dk1k1_du}). For \texttt{OTM} we have a single multiplication between a sparse rotation matrix $Q_{k+1|k}$ and a vector, both in the dimension of the joint state at time $k$ plus the number of rows of the linearized new factors (i.e. depending on number of factors and their types). 
The complexity of \texttt{OTM} would be given by the number of non zeros in the rotation matrix $Q_{k+1|k}$. In Appendix-C we provide some understanding on the creation of the rotation matrix $Q_{k+1|k}$, and also develop an expression for the number of non zeros in $Q_{k+1|k}$. We direct the reader to Figure~\ref{fig:Q:param} for illustration of the new notations used in this discussion.
Following the development in Appendix-C, the number of non zeros in $Q_{k+1|k}$ is represented by two potentially dominant terms separated by a simple condition
\begin{equation}\label{eq:nonZeroQ}
	O(\texttt{OTM}) = \begin{cases}
		O\! \left( \left(n^s_k + n^f_{k+1} - j\right) ^{2} \right)  & n^f_{k+1} \geq n^s_{k} \geq 6\\
 	O\! \left( n^s_k \cdot n^f_{k+1} \right)  & n^f_{k+1} < n^s_{k} 	
 \end{cases}
 \end{equation}
where $j$ denotes the column index of the left-most entry in $\mathcal{A}_{k+1|k+1}$, $n^s_{k}$ denotes the size of the joint state vector at the precursory time $k$, $n^f_{k+1}$ denotes the number of rows in the linearized new factors $\mathcal{A}_{k+1|k+1}$. 
The condition in Eq.~(\ref{eq:nonZeroQ}) is a simple upper bound to the real expression (see Eq.~(\ref{eq:Q:realCond})), resulting with a cleaner condition without affecting the solution.

It is worth stressing that depending on its type, each state occupies more than a single row / column in the Jacobian, e.g. 6DOF robot pose occupies six rows and six columns. Similarly, depending on its type, each factor occupies more than a single row in the Jacobian, e.g.~a monocular factor occupies two rows in the Jacobian.

 For \texttt{DU} in addition to multiplications between upper triangular matrices and vectors, we have a matrix inverse. Differently from \texttt{OTM} here the matrix dimensions are of the joint state vector at time $k+1$, hence the worst case scenario for \texttt{DU} is a fully dense upper triangular matrix inverse 
\begin{equation}\label{eq:complexity_DU}
	O(\texttt{DU}) = O\! \left( {n^s_{k+1}}^{2} \right),	
\end{equation}
where $n^s_{k+1}$ represents the size of the joint state vector at time $k+1$. 

For the case of $n^f_{k+1} < n^s_{k}$ we should compare
\begin{equation}
	n^s_k \cdot n^f_{k+1} \overset{?}{\lessgtr}  n^s_{k+1} \cdot n^s_{k+1}.
\end{equation}
Assuming states are not removed from the state vector, we can say 
\begin{equation}
	n^s_k \leq n^s_{k+1},
\end{equation}
then evidently 
\begin{equation}\label{eq:Q:case2}
	n^s_k \cdot n^f_{k+1} <  n^s_{k+1} \cdot n^s_{k+1}.
\end{equation}
For the case of $n^f_{k+1} \geq n^s_{k} \geq 6$ we should compare
\begin{equation}
	n^s_k + n^f_{k+1} - j \overset{?}{\lessgtr}  n^s_{k+1} \qquad ,\qquad j \in [1,n^s_k],
\end{equation}
so for this case \texttt{OTM} is computationally superior to \texttt{DU} if
\begin{equation}\label{eq:Q:case1}
	 n^f_{k+1} <  n^s_{k+1} - n^s_k + j \qquad ,\qquad j \in [1,n^s_k].
\end{equation}
It is worth stressing that unlike Eq.~(\ref{eq:Q:case2}), Eq.~(\ref{eq:Q:case1}) is dependent on state ordering in the form of the left-most non zero entry in $\mathcal{A}_{k+1|k+1}$.

Concluding the complexity analysis of \texttt{OTM} and \texttt{DU}, \texttt{OTM} will be computationally superior to \texttt{DU} if the following holds
\begin{equation}\label{eq:Q:otmBetter}
	 \left( n^f_{k+1} < n^s_{k}\right) \quad \cup \quad \left( n^f_{k+1} <  n^s_{k+1} - n^s_k + j \quad \cap \quad n^f_{k+1} \geq n^s_{k} \right). 
\end{equation}
In other words, if the number of rows in $\mathcal{A}_{k+1|k+1}$ is smaller than the size of the state vector at time $k$ \texttt{OTM} is computationally superior to \texttt{DU}. If the number of rows in $\mathcal{A}_{k+1|k+1}$ is larger or equal to the size of the state vector at time $k$, than \texttt{OTM} is computationally superior to \texttt{DU} only if the number of rows in $\mathcal{A}_{k+1|k+1}$ is smaller than the size of the added states at time $k+1$ plus the column index of the left-most state in $\mathcal{A}_{k+1|k+1}$.

Although most of the time \texttt{DU} is computationally inferior, unlike \texttt{OTM} that requires access to the rotation matrix which might not be easily available in every planning paradigm, \texttt{DU} makes use in a more readily available information: the inference solution of precursory time, the predicted factors, the new RHS vector at time $k+1$, and the factorized Jacobian from precursory planning. Therefore the advantage in using \texttt{DU} lies in the information availability with minimal adjustments to the planning stage.

Since \texttt{OTM-OO} would prove to perform the best empirically, let us get some intuition on why it is more efficient than \texttt{OTM}.
The \texttt{OO} addition to \texttt{OTM}, refers to the use of the motion propagated belief $R^{\mathcal{F}}_{k+1|k}$ $d^{\mathcal{F}}_{k+1|k}$ rather than the use of precursory inference solution $R_{k|k}$ $d_{k|k}$. The dimension of $R^{\mathcal{F}}_{k+1|k}$ is larger from that of $R_{k|k}$ by a single robot pose, while the number of rows of $\mathcal{H}_{k+1|k+1}$ is smaller by a single robot pose from that of $\mathcal{A}_{k+1|k+1}$. Let us assume without affecting generality that our robot pose dimension is $\alpha$. Under this assumption we can calculate Eq.~(\ref{eq:nonZeroQ}) for both \texttt{OTM} and \texttt{OTM-OO}. Let $n^{\neg f}_{k+1}$ denote the number of rows of the newly added factors at time $k+1$ without the motion factor, i.e. $\mathcal{H}_{k+1|k+1}$ number of rows, so the complexity of \texttt{OTM} would be
\begin{equation}\label{eq:otmCompare}
	O(\texttt{OTM}) = \begin{cases}
		O\! \left( \left(n^s_k + \left(n^{\neg f}_{k+1} + \alpha\right) - j\right) ^{2} \right)  & \left(n^{\neg f}_{k+1} + \alpha\right) \geq n^s_{k}\\
 	O\! \left( n^s_k \cdot \left(n^{\neg f}_{k+1} + \alpha\right) \right) = O\! \left( n^s_k \cdot n^{\neg f}_{k+1} + \alpha \cdot n^s_k \right)  & \left(n^{\neg f}_{k+1} + \alpha\right) < n^s_{k} 	
 \end{cases},
 \end{equation}
while the complexity of \texttt{OTM-OO} would be
\begin{equation}\label{eq:otmooCompare}
	O(\texttt{OTM-OO}) = \begin{cases}
		O\! \left( \left(\left(n^s_k+\alpha\right) + n^{\neg f}_{k+1} - j' \right) ^{2} \right)  & n^{\neg f}_{k+1}  \geq \left(n^s_{k}+ \alpha\right)\\
 	O\! \left( \left(n^s_{k}+ \alpha\right) \cdot n^{\neg f}_{k+1} \right) = O\! \left( n^s_{k} \cdot n^{\neg f}_{k+1} + \alpha \cdot n^{\neg f}_{k+1} \right)  & n^{\neg f}_{k+1} < \left(n^s_{k}+ \alpha\right) 	
 \end{cases},
 \end{equation}
where $j' \in [1,\left(n^s_{k}+ \alpha\right)]$, opposed to $j \in [1,n^s_{k}]$.
From comparing Eqs.~(\ref{eq:otmCompare} - \ref{eq:otmooCompare}), for the case where the size of added factors is larger than the state, we can deduce that other than the difference between $j$ and $j'$, they are the same. Judging the second case, we can see they differ by the difference between the size of the state at time $k$ and the number of $\mathcal{A}_{k+1|k+1}$ rows. As we will see later on, \texttt{OTM-OO} empirically proves to be more efficient than \texttt{OTM}, which means that the state at time $k$ is in fact larger than the number of size of $\mathcal{A}_{k+1|k+1}$ rows. 
 
Revisiting Eq.~(\ref{eq:Q:otmBetter}) in-light of the understanding that the state at time $k$ is in fact larger than the number of $\mathcal{A}_{k+1|k+1}$ rows we can say that \texttt{OTM} is computationally superior to \texttt{DU} without any restricting conditions.
 
\subsection{Inconsistent Data Association}\label{subsec:UpdateDA_chapter}

In order to address the more general and realistic scenario, the DA might require correction before proceeding to update the new acquired measurements. In the sequel we cover the possible scenarios of inconsistent data association and its graphical materialization, followed by a paradigm to update inconsistent DA from planning stage according to the actual DA attained in the consecutive inference stage.
We later examine both the computational aspects and the sensitivity of the paradigm to various parameters both on simulated and real-life data.

\subsubsection{Types of inconsistent DA}\label{ssubsec:DAtypes}
We would now discuss, without losing generality, the actual difference between the two aforementioned beliefs $b[X_{k+1|k}]$ and $b[X_{k+1|k+1}]$. As already presented in Section~\ref{subsec:update_RHS}, in case of a consistent DA i.e. $\mathcal{M}_{k+1|k} = \mathcal{M}_{k+1|k+1}$, the difference between the two beliefs is narrowed down to the RHS vectors $d_{k+1|k}$ and $d_{k+1|k+1}$ which encapsulates the measurements $z_{k+1|k}$ and $z_{k+1|k+1}$ respectively.
However, in the real world it is possible that the DA predicted in precursory planning would prove to be inconsistent to the DA attained in inference.

There are six possible scenarios representing the relations between DA in inference and precursory planning: 
\begin{itemize}
	\item In planning, association is assumed to either a new or existing variable, while in inference no measurement is received. 
	\item In planning it is assumed there will be no measurement to associate to, while in inference a measurement is received and associated to either a new or existing variable.
	\item In planning, association is assumed to an existing variable, while in inference it is to a new variable.  
	\item In planning, association is assumed to a new variable, while in inference it is to an existing variable.
	\item In planning, association is assumed to an existing variable, while in inference it is also to an existing variable (whether the same or not). 
	\item In planning, association is assumed to a new variable, while in inference it is also to a new variable (whether the same or not). 
\end{itemize}
While the first four bullets always describe inconsistent DA situations (e.g.~in planning we assumed a known tree would be visible but instead we saw a new bench, or vice versa), the last two bullets may provide consistent DA situations. In case associations in planning and in inference are to the same (un)known variables we would have a consistent DA. 

While different planning paradigms might diminish occurrences of inconsistent DA, e.g.~by better predicting future associations, none can avoid it completely. Methods to better predict future observations/associations will be investigated in future work, potentially leveraging Reinforcement Learning (RL) techniques.
As mentioned in Section~\ref{subsec:bsp}, in this paper we do not predict occurrences of new landmarks, hence every new landmark in inference would result in inconsistent DA.

In the following section we provide a method to update inconsistent DA, regardless of a specific inconsistency scenario or a solution paradigm. This method utilizes the incremental methodologies of iSAM2 \citep{Kaess12ijrr} in order to efficiently update the belief from the planning stage to have consistent DA with the succeeding inference.
\begin{table}
	\caption{Notations for Section~\ref{ssubsec:updateDA}} 
	\centering 
	\begin{tabular}{c c} 
		\hline\hline 
		\textbf{Variable} & \textbf{Description}  \\ [0.5ex] 
		\hline 
		$ \Box_{t|k}$ & Of time $t$ while current time is $k$ \\[1ex]
		$\mathcal{FG}_{t|k}$ & Factor graph (FG) at time $t$ \\[1ex]
		$\mathcal{T}_{t|k}$ & Bayes Tree (BT) at time $t$ \\[1ex]
		$\mathcal{M}_{t|k}$ & Data Association (DA) at time $t$ \\[1ex]
		$\mathcal{M}^{\bigcap}_{t}$ & Consistent DA at time $t$ \\[1ex]
		$\mathcal{M}^{rmv}_{t}$ & DA at time $t$ from planning inconsistent with inference, indicating factors to be removed\\[1ex]
		$\mathcal{M}^{add}_{t}$ & DA at time $t$ from inference inconsistent with planning, indicating factors to be added \\[1ex]
		$ \{ f_r \}^{rmv}_{t}$ & Factors at time $t$ from planning inconsistent with inference, to be removed\\[1ex]
		$ \{ f_s \}^{add}_{t}$ & Factors at time $t$ from inference inconsistent with planning, to be added\\[1ex]
		$\{ X\}^{inv}_{t}$ & All states at time $t$, involved in $ \{ f_r \}^{rmv}_{t}$ and $ \{ f_s \}^{add}_{t}$ \\ [1ex]
		$\mathcal{T}^{inv}_{t}$ & Sub-BT of $\mathcal{T}_{t|k}$ composed of all cliques containing $\{ X\}^{inv}_{t}$ \\[1ex]
		$\{ X\}^{inv\star}_{t}$ & All states at time $t$, related to the sub-BT $\mathcal{T}^{inv}_{t}$ \\ [1ex]
		$\mathcal{FG}^{inv}_{t}$ & The detached part of $\mathcal{FG}_{t|k}$ containing $\{ X\}^{inv\star}_{t}$ \\[1ex]
		$\mathcal{FG}^{upd}_{t}$ &  The FG $\mathcal{FG}^{inv}_{t}$ after DA update \\ [1ex]
		$\mathcal{T}^{upd}_{t}$ & The sub-BT eliminated from $\mathcal{FG}^{upd}_{t}$ \\ [1ex]
		$\mathcal{FG}^{upd}_{t|k}$ & The Factor Graph at time $t$ with all-correct DA \\ [1ex]
		$\mathcal{T}^{upd}_{t|k}$ & The Bayes Tree at time $t$ with all-correct DA \\ [1ex]
		\hline\hline
	\end{tabular}
	\label{table:updateDA} 
\end{table}
\subsubsection{Updating Inconsistent DA}\label{ssubsec:updateDA}
\begin{figure}
        \centering
        \subfloat[]{\includegraphics[trim={0 642 190 0},clip, width=0.48\columnwidth]{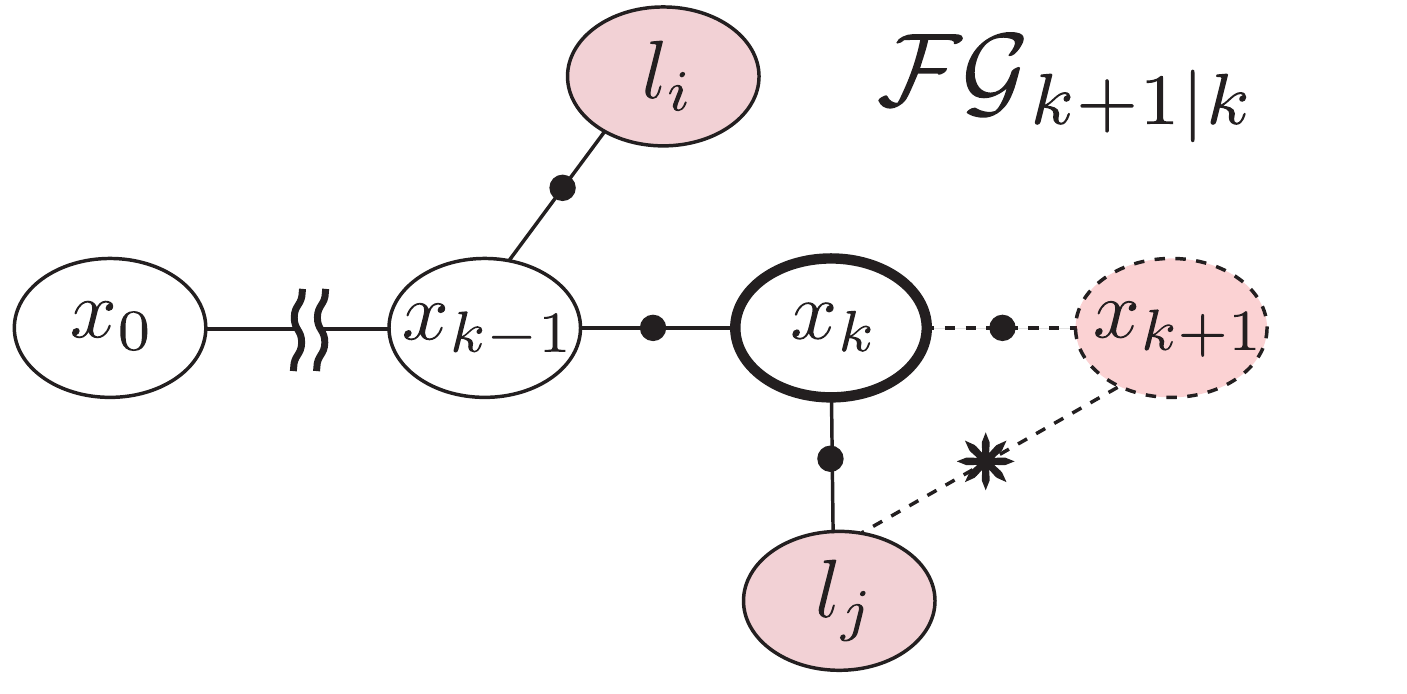}\label{fig:FG_k1_k}}
         \subfloat[]{\includegraphics[trim={0 639 190 0},clip, width=0.48\columnwidth]{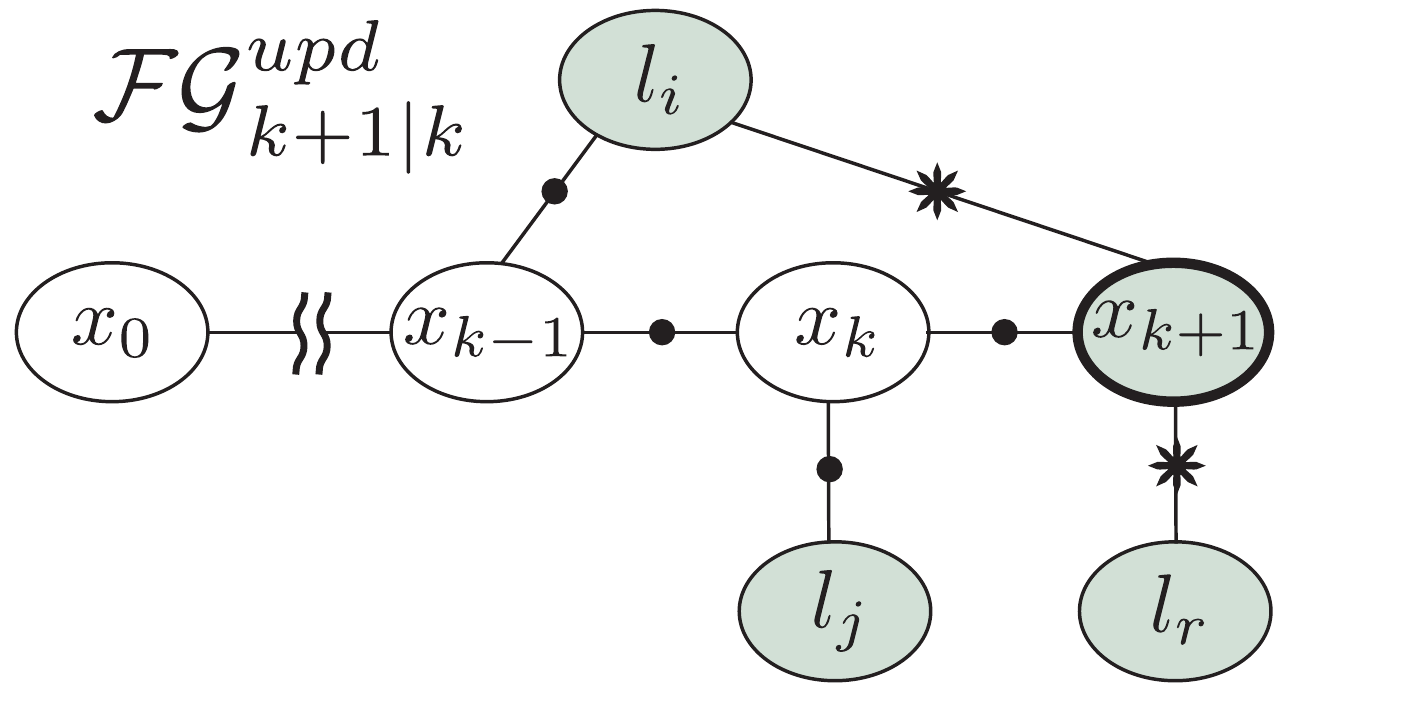}\label{fig:FG_k1_k1}}\\
          \subfloat[]{\includegraphics[trim={0 638 0 0},clip, width=0.6\columnwidth]{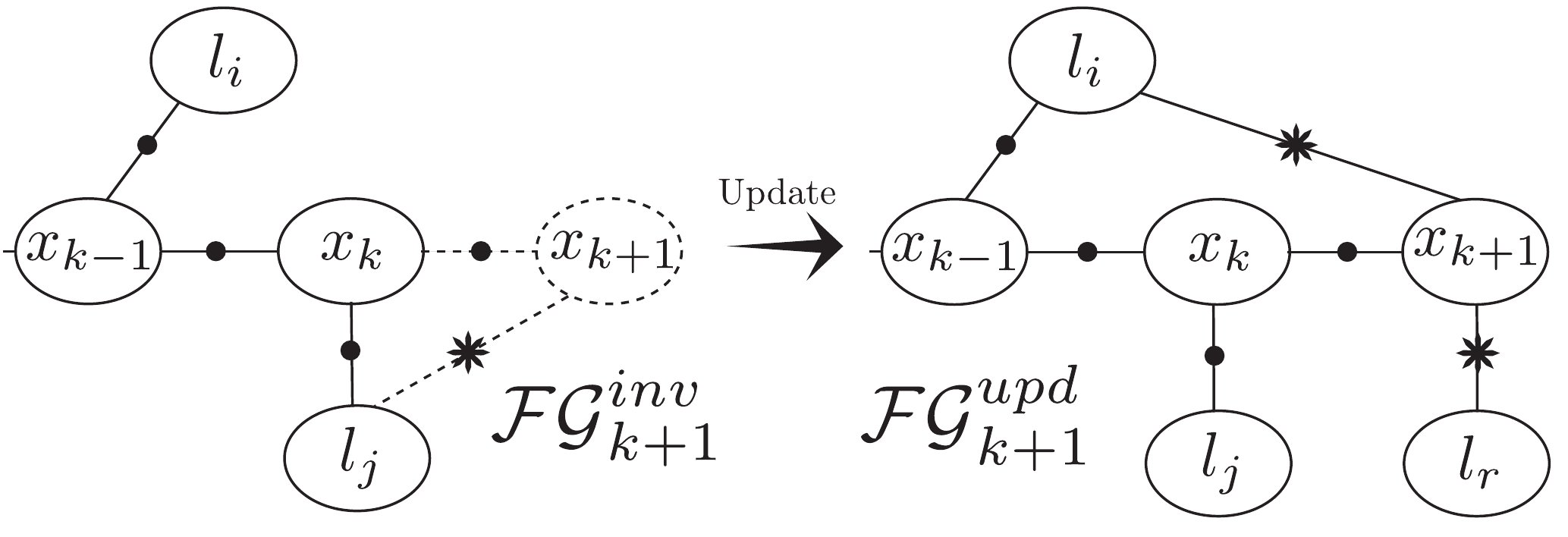}\label{fig:FG_update}}\\
           \subfloat[]{\includegraphics[trim={0 290 310 275},clip, width=0.28\columnwidth]{Figures/FG_BN_lc_6.pdf}\label{fig:BT_k1_k}}
            \subfloat[]{\includegraphics[trim={290 278 0 275},clip, width=0.28\columnwidth]{Figures/FG_BN_lc_6.pdf}\label{fig:BT_k1_k1}}
        \caption{ The process of incremental DA update, following on \texttt{iSAM2} methodologies. (a) and (b) show factor graphs for $b[X_{k+1|k}]$ and $b[X_{k+1|k+1}]$, respectively, which differ due to incorrect association considered in the planning phase - $l_j$ was predicted to be observed within planning, while in practice $l_i$ and $l_r$ were observed at time instant $k+1$. In (a), current-time robot pose is bolded, horizon factors and states are dotted. Involved variables from DA comparison are marked in red in (a) and green in (b). The belief $b[X_{k+1|k}]$, represented by a Bayes tree shown in (d), is divided in two: sub Bayes tree containing all involved variables and parent cliques up to the root (marked in blue) and the rest of the Bayes tree in white. The former sub Bayes tree is re-eliminated by (i) forming the corresponding portion of the factor graph, as shown in the left figure of (c); (ii) removing incorrect DA and adding correct DA factors, which yields the factor graph shown in the right figure of (c); (iii) re-eliminating that factor graph into a sub Bayes tree, marked blue in (e), and  re-attaching the rest of the Bayes tree. While the obtained Bayes tree now has a correct DA, it is conditioned on (potentially) incorrect measurement values for consistent-DA factors, which therefore need to be updated (as detailed in Section \ref{subsec:update_RHS}), to recover the posterior belief $b[X_{k+1|k+1}]$.}
        \label{fig:FG-BT}
\end{figure}

Inconsistent DA can be interpreted as disparate connections between variables. As discussed earlier, these connections, denoted as factors, manifest in rows of the Jacobian matrix or in factor nodes of a FG.
Two FGs with different DA would thus have different graph topology. We demonstrate the inconsistent DA impact over graph topology using the example presented in Figure~\ref{fig:FG-BT}:  Figure~\ref{fig:FG_k1_k} represents the belief $b[X_{k+1|k}]$ from planning stage, and Figure~\ref{fig:FG_k1_k1} represents the belief $b[X_{k+1|k+1}]$ from the inference stage.  Even-though the same elimination order is used, the inconsistent DA would also create a different topology between the resulting BTs, e.g.~the resulting BTs for the aforementioned FGs are Figure~\ref{fig:BT_k1_k} and Figure~\ref{fig:BT_k1_k1} accordingly. 

Performing action $u_{k|k+1}$, provides us with new measurements $z_{k+1|k+1}$, which are gathered to the factor set $\{ f_j \}_{k+1|k+1}$ (see Appendix-B for factor definition). From the precursory planning stage we have the belief $b[X_{k+1|k}]$ along with the corresponding factor set $\{ f_i \}_{k+1|k}$ for time $k+1$. Since we performed inference over this belief during the planning stage, we have already eliminated the FG, denoted as $\mathcal{FG}_{k+1|k}$, into a BT denoted as $\mathcal{T}_{k+1|k}$, e.g.~see Figure~\ref{fig:FG_k1_k} and Figure~\ref{fig:BT_k1_k}, respectively. 

We would like to update both the FG $\mathcal{FG}_{k+1|k}$ and the BT $\mathcal{T}_{k+1|k}$ from the planning stage, using the new factors $\{ f_j \}_{k+1|k+1}$ from the inference stage. 
Without losing generality we  use Figure~ \ref{fig:FG-BT} to demonstrate and explain the DA update process.
Let us consider all factors of time $k+1$ from both planning $\{ f_i \}_{k+1|k}$ and inference $\{ f_j \}_{k+1|k+1}$. We can divide these factors into three categories: 

The first category contains factors with consistent DA - Good Factors. These factors originate from only the last two DA scenarios, in which both planning and inference considered either the same existing variable or a new one. Consistent DA factors do not require our attention (other than updating the measurements in the RHS vector). 
Indices of consistent DA factors can be obtained by intersecting the DA from planning with that of inference:
\begin{equation}\label{eq:same_DA}
		\mathcal{M}^{\bigcap}_{k+1} = \mathcal{M}_{k+1|k} ~\bigcap ~\mathcal{M}_{k+1|k+1}.
\end{equation}
The second category - Wrong Factors, contains factors from planning stage with inconsistent DA to inference, which therefore should be removed from $\mathcal{FG}_{k+1|k}$. These factors can originate from all DA scenarios excluding the second.
Indices of inconsistent DA factors from planning, can be obtained by calculating the relative complement of $ \mathcal{M}_{k+1|k}$ with respect to $ \mathcal{M}_{k+1|k+1}$:
\begin{equation}\label{eq:rmv_DA}
		\mathcal{M}^{rmv}_{k+1} =  \mathcal{M}_{k+1|k} ~\backslash~ \mathcal{M}_{k+1|k+1}. 
\end{equation}
The third category - New Factors, contains factors from the inference stage with inconsistent DA to planning; hence, these factors should be added to $\mathcal{FG}_{k+1|k}$. These factors can originate from all DA scenarios excluding the first.
Indices of inconsistent DA factors from inference, can be obtained by calculating the relative complement of $ \mathcal{M}_{k+1|k+1}$ with respect to $ \mathcal{M}_{k+1|k}$:
\begin{equation}\label{eq:good_DA}
		\mathcal{M}^{add}_{k+1} = \mathcal{M}_{k+1|k+1} ~\backslash~ \mathcal{M}_{k+1|k}.
\end{equation}
We now use our example from Figure~\ref{fig:FG-BT} to illustrate these different categories:
\begin{itemize}
 \item The first category - Good Factors, contains all factors from time $k+1$ that appear both in Figure \ref{fig:FG_k1_k} and \ref{fig:FG_k1_k1}, i.e. the motion model factor between $x_k$ to $x_{k+1}$.
 \item The second category - Wrong Factors, contains all factors that appear only in Figure~\ref{fig:FG_k1_k}, i.e. the star marked factor in Figure~\ref{fig:FG_k1_k}. In this case the inconsistent DA is to an existing variable, landmark $l_j$ was considered to be observed in planning but is not seen in the succeeding inference.
 \item The third category - New Factors, contains all factors that appear only in Figure~\ref{fig:FG_k1_k1}, i.e. the star marked factors in Figure~\ref{fig:FG_k1_k1}. In this case the inconsistent DA is both to an existing and a new variable. Instead of landmark $l_j$ that was considered to be observed in planning, a different existing landmark $l_i$ has been seen, along with a new landmark $l_r$.
\end{itemize}

\begin{algorithm}
\caption{- Data Association Update}\label{alg:DaUpdate}
\begin{algorithmic}[1]
\linespread{1.6}\selectfont
\Function{UpdateDA}{$\mathcal{FG}_{k+1|k}$ , $\mathcal{M}_{k+1|k}$ , $\mathcal{FG}_{k+1|k+1}$ , $\mathcal{M}_{k+1|k+1}$}
 \State $\mathcal{M}^{rmv}_{k+1}$ $\gets$  $\mathcal{M}_{k+1|k} ~\backslash~ \mathcal{M}_{k+1|k+1}$ \label{alg:DaUpdate:rmvFac_index}
\Comment{ indices of factors required to be removed}
\State $\mathcal{M}^{add}_{k+1}$ $\gets$  $\mathcal{M}_{k+1|k+1} ~\backslash~ \mathcal{M}_{k+1|k}$ \label{alg:DaUpdate:addFac_index}
\Comment{ indices of factors required to be added}
\State  $\{ f_r\}^{rmv}_{k+1}$  $\gets$  $\prod\limits_{r \in \mathcal{M}^{rmv}_{k+1}} \{f_r\}_{k+1}$ \label{alg:DaUpdate:rmvFac}
\Comment{ factors required to be removed}
\State $\{ f_s\}^{add}_{k+1}$ $\gets$ $\prod\limits_{s \in \mathcal{M}^{add}_{k+1}} \{f_s\}_{k+1}$ \label{alg:DaUpdate:addFac}
\Comment{ factors required to be added}
\State $\{ X\}^{inv}_{k+1}$   $\gets$   $Variables(\{ f_r\}^{rmv}_{k+1})$ $\bigcup$  $Variables(\{ f_s\}^{add}_{k+1})$ \label{alg:DaUpdate:invVar}
\Comment{ get involved variables}
\State $\mathcal{T}^{inv}_{k+1}$  $\gets$  $\mathcal{T}_{k+1|k}^{\{ X\}^{inv}_{k+1}}$ \label{alg:DaUpdate:invBT}
\Comment{ get corresponding sub-BT}
\State $\{ X\}^{inv\star}_{k+1}$   $\xleftarrow[]{\text{get all variables}}$  $\mathcal{T}^{inv}_{k+1}$ \label{alg:DaUpdate:invVar_upd}
\Comment{ update involved variables}
\State $\mathcal{FG}^{inv}_{k+1}$  $\gets$  $\mathcal{FG}_{k+1|k}^{\{ X\}^{inv\star}_{k+1}}$ \label{alg:DaUpdate:invFG}
\Comment{ get corresponding sub-FG}
\State $\mathcal{FG}^{upd}_{k+1}$  $\gets$  $[ \mathcal{FG}^{inv}_{k+1} \backslash \{ f_r\}^{rmv}_{k+1}] \bigcup \{ f_s\}^{add}_{k+1}$ \label{alg:DaUpdate:invFG_upd}
\Comment{Update the sub Factor Graph}
\State $\mathcal{T}^{upd}_{k+1}$  $\xleftarrow[]{\text{eliminate}}$  $ \mathcal{FG}^{upd}_{k+1}$\label{alg:DaUpdate:invBT_upd}
\Comment{ re-eliminate the updated sub-FG into BT}
\State $\mathcal{FG}^{upd}_{k+1|k}$  $\gets$  $[\mathcal{FG}_{k+1|k} \backslash \mathcal{FG}^{inv}_{k+1}] \bigcup \mathcal{FG}^{upd}_{k+1}$\label{alg:DaUpdate:updFG}
\Comment{Update the Factor Graph}
\State $\mathcal{T}^{upd}_{k+1|k}$  $\gets$  $[\mathcal{T}_{k+1|k} \backslash \mathcal{T}^{inv}_{k+1}] \bigcup \mathcal{T}^{upd}_{k+1}$\label{alg:DaUpdate:updBT}
\Comment{Update the Bayes Tree}
\State \textbf{return} $\mathcal{FG}^{upd}_{k+1|k}$ , $\mathcal{T}^{upd}_{k+1|k}$ .
\EndFunction
\end{algorithmic}
\end{algorithm}
Once the three aforementioned categories are determined, we use iSAM2 methodologies, presented in \cite{Kaess12ijrr}, to incrementally update $\mathcal{FG}_{k+1|k}$ and $\mathcal{T}_{k+1|k}$, see Alg.~\ref{alg:DaUpdate}.
The involved factors are denoted by all factors from planning needed to be removed (Wrong Factors), and all factors from inference needed to be added (New Factors),
\begin{equation}\label{eq:involved_vars}
	\{ f_r\}^{rmv}_{k+1} = \!\!\!\!\prod_{r \in \mathcal{M}^{rmv}_{k+1}} \!\!\!\!f_r \quad , \quad 
	\{ f_s\}^{add}_{k+1} = \!\!\!\!\prod_{s \in \mathcal{M}^{add}_{k+1}} \!\!\!\!f_s .
\end{equation}
The involved variables, denoted by $\{ X\}^{inv}_{k+1}$, are all variables related to the factor set $\{ f_r\}^{rmv}_{k+1}$ and the factor set $\{ f_s\}^{add}_{k+1}$ (Alg.~\ref{alg:DaUpdate}, line~\ref{alg:DaUpdate:invVar}), e.g. the colored variables in Figures~\ref{fig:FG_k1_k} and \ref{fig:FG_k1_k1} accordingly.
In $\mathcal{T}_{k+1|k}$, all cliques between the ones containing $\{ X\}^{inv}_{k+1}$  up to the root are marked and denoted as the involved cliques, e.g. colored cliques in Figure~\ref{fig:BT_k1_k}.
The involved cliques are detached and denoted by $\mathcal{T}^{inv}_{k+1} \subset \mathcal{T}_{k+1|k}$ (line~\ref{alg:DaUpdate:invBT}). 
This sub-BT $\mathcal{T}^{inv}_{k+1}$, contains more variables than just $\{ X\}^{inv}_{k+1}$. The involved variable set $\{ X\}^{inv}_{k+1}$, is then updated to contain all variables from $\mathcal{T}^{inv}_{k+1}$ and denoted by $\{ X\}^{inv \star}_{k+1}$ (line~\ref{alg:DaUpdate:invVar_upd}).
The part of $\mathcal{FG}_{k+1|k}$, that contains all involved variables $\{ X\}^{inv \star}_{k+1}$ is detached and denoted by $\mathcal{FG}^{inv}_{k+1}$ (line~\ref{alg:DaUpdate:invFG}). While  $\mathcal{T}^{inv}_{k+1}$ is the corresponding sub-BT to the acquired sub-FG $\mathcal{FG}^{inv}_{k+1}$.

In order to finish updating the DA, all that remains is updating the sub-FG  $\mathcal{FG}^{inv}_{k+1}$ with the correct DA and re-eliminate it to get an updated BT. 
All factors $\{ f_r\}^{rmv}_{k+1}$ are removed from $\mathcal{FG}^{inv}_{k+1}$, then all factors $\{ f_r\}^{add}_{k+1}$ are added (line~\ref{alg:DaUpdate:invFG_upd}). The updated sub-FG is denoted by $\mathcal{FG}^{upd}_{k+1}$, e.g. update illustration in Figure~\ref{fig:FG_update}. 

By re-eliminating $\mathcal{FG}^{upd}_{k+1}$, a new updated BT, denoted by $\mathcal{T}^{upd}_{k+1}$, is obtained (line~\ref{alg:DaUpdate:invBT_upd}), e.g. the colored sub-BT in Figure~\ref{fig:BT_k1_k1}. This BT is then re-attached back to $\mathcal{T}_{k+1|k}$ instead of $\mathcal{T}^{inv}_{k+1}$, subsequently the new BT is now with consistent DA and is denoted as $\mathcal{T}^{upd}_{k+1|k}$ (line~\ref{alg:DaUpdate:updBT}). In a similar manner $\mathcal{FG}^{upd}_{k+1|k}$ is obtained by re-attaching $\mathcal{FG}^{upd}_{k+1}$ instead of $\mathcal{FG}^{inv}_{k+1}$ to $\mathcal{FG}_{k+1|k}$(line~\ref{alg:DaUpdate:updFG}). 
 At this point the DA in both the FG and the BT is fixed. For example, by completing the aforementioned steps, Figures~\ref{fig:FG_k1_k} and \ref{fig:BT_k1_k} will have the same topology as Figures~\ref{fig:FG_k1_k1} and \ref{fig:BT_k1_k1}.

After the DA update, the BT  $\mathcal{T}^{upd}_{k+1|k}$ has consistent DA to that of $\mathcal{M}_{k+1|k+1}$. However, it is still not identical to $\mathcal{T}_{k+1|k+1}$ due to difference between measurement values predicted in planning to the values obtained in inference. The DA update dealt with inconsistent DA factors and their counterparts. For these factors the new measurements from inference were updated in the corresponding RHS vector values within the BT. The consistent DA factors, on the other hand, were left untouched; therefore, these factors do not contain the new measurement values from inference but measurement values from the planning stage instead. These inconsistent measurements are thus baked into the RHS vector $d_{k+1|k}$ and in the appropriate cliques of the BT $\mathcal{T}^{upd}_{k+1|k}$. In order to update the RHS vector $d_{k+1|k}$, or equivalently update the corresponding values within relevant cliques of the BT, one can use any of the methods presented in Section~\ref{subsec:update_RHS}.

\section{Results}
\label{sec:results}
In this section we present an extensive analysis of the proposed paradigm for \texttt{RUB inference} and benchmark it against the standard Bayesian inference approach using iSAM2 efficient methodologies as a proving-ground.

We consider the problem of autonomous navigation and mapping in an unknown environment as a testbed for the proposed paradigm, first in a simulated environment and later-on in a real-world environment (as discussed in the sequel). The robot performs inference to maintain a belief over its current and past poses and the observed landmarks thus far (i.e.full-SLAM), and uses this belief to decide its next actions within the framework of belief space planning. As mentioned earlier, our proposed paradigm is indifferent to a specific method of inference or decision making.

In order to test the computational effort, we compared inference update using iSAM2 efficient methodology, once based on the standard Bayesian inference paradigm \citep{Kaess12ijrr} (here on denoted as \texttt{iSAM}), and second based on our proposed \texttt{RUB inference} paradigm.

All of our complementary methods (see Section~\ref{subsec:update_RHS}), required to enable inference update based on the \texttt{RUB inference} paradigm, were implemented in MATLAB and are \emph{encased within the inference block}. The \texttt{iSAM} approach uses the \texttt{GTSAM C++} implementation with the supplied MATLAB wrapper \citep{Dellaert12tr}. Considering the general rule of thumb, that MATLAB implementation is at least one order of magnitude slower, the comparison to \texttt{iSAM} as a reference is conservative. All runs were executed on the same Linux machine, with Xeon E3-1241v3 $3.5$ GHz processor with $32$ GB of memory. 

In order to get better understanding of the difference between our proposed paradigm and the standard Bayesian inference, we refer to the high-level algorithm diagram given in Figure \ref{fig:HighLevel_Algo}, which  depicts a plan-act-infer framework.  Figure~\ref{fig:HighLevel_Algo:std} represents a standard Bayesian inference, where only the first inference update iteration is timed for comparison reasons.  Figure~\ref{fig:HighLevel_Algo:ours} shows our novel paradigm \texttt{RUB inference}, while the DA update, along with the first inference update iteration, are being timed for comparison.
The computation time comparison is  made only over the inference stage, since the rest of the plan-act-infer framework is \emph{identical} in both cases.

As mentioned, our proposed paradigm does not affect estimation accuracy. We verify that in the following experiments, by comparing the estimation results obtained using our approach and \texttt{iSAM}. Both provide essentially the same results in all cases; we provide an explicit accuracy comparison with real-world data experiment (Section \ref{subsec:kitti}).
 
\subsection{Simulated Environment}
\subsubsection{Basic Analysis - Sanity Check}\label{ssubsec:BasicAnalysis}
The purpose of this experiment is to provide with a basic comparison between the suggested paradigm for \texttt{RUB inference} and the existing standard Bayesian inference. This simulation performs a single horizon BSP calculation, followed by an inference step with a single inference update. The simulation provides a basic analysis of running time for each method, denoted by the \emph{vertical} axis, for a \emph{fully dense information matrix} and with no loop closures. The presented running time is a result of an average between $10^3$ repetitions per step per method. Although a fully dense matrix does not represent a real-world scenario, it provides a sufficient initial comparison. The simulation analyzes the sensitivity of each method to the initial state vector size, denoted by the \emph{horizontal} axis, and to the number of new factors, denoted by the different graphs.
Since we perform a single horizon step with a single inference update, no re-linearization is necessary; hence, iSAM comparison is valid. The purpose of this check is to provide a simple sensitivity analysis of our methods to state dimension and number of new factors per step, while compared against standard batch update (denoted as \texttt{STD}) and \texttt{iSAM} paradigm. While both \texttt{STD} and \texttt{iSAM} are based on the standard Bayesian inference paradigm, the rest of the methods are based on the novel \texttt{RUB inference} paradigm.

\begin{figure}
        \centering
        \subfloat[]{\includegraphics[bb={0 0 0 0},trim={0 0 0 0},clip, width=0.3\columnwidth]{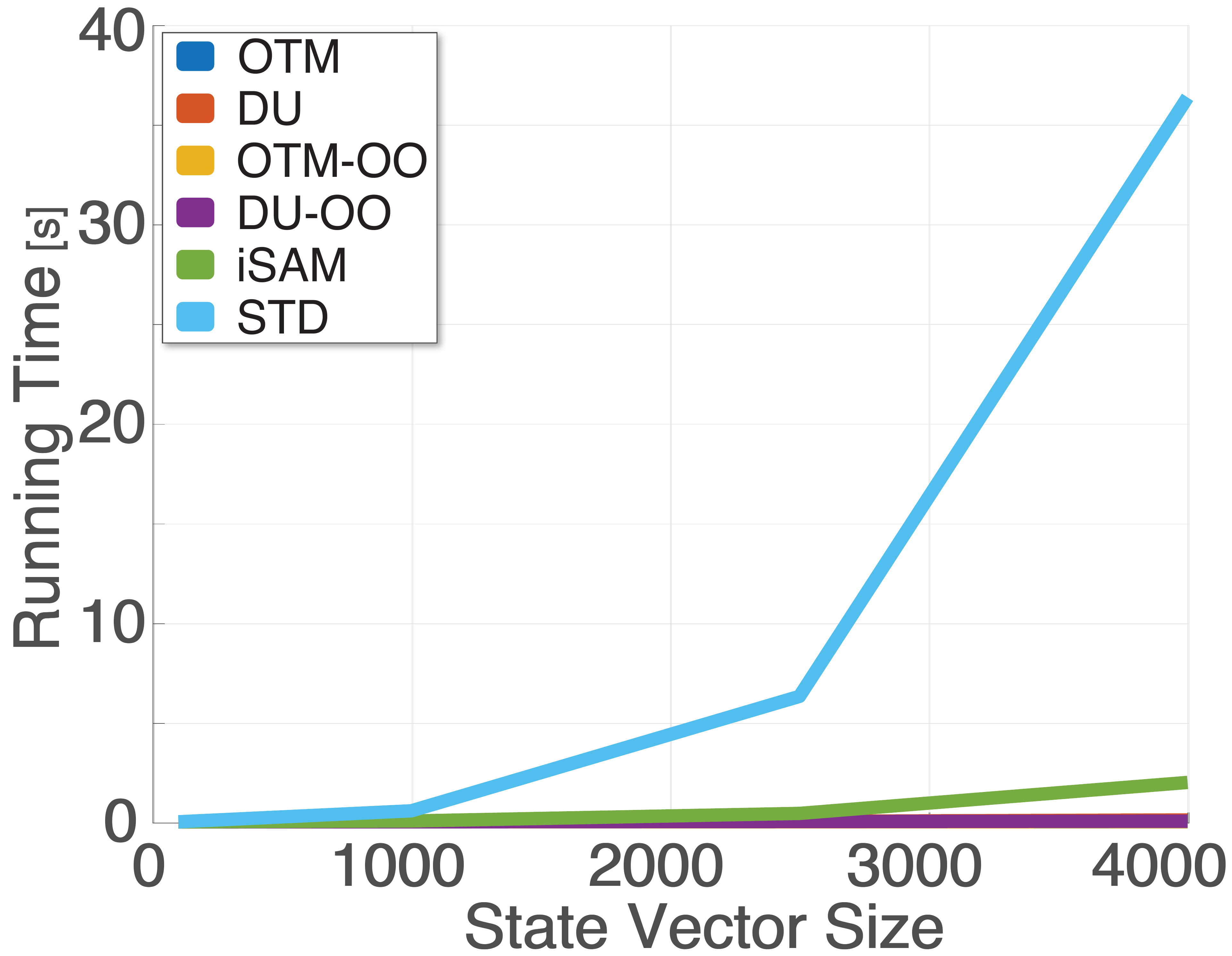}\label{fig:basic_Comp_all:2}}
        \subfloat[]{\includegraphics[bb={0 0 0 0},trim={0 0 0 0},clip, width=0.3\columnwidth]{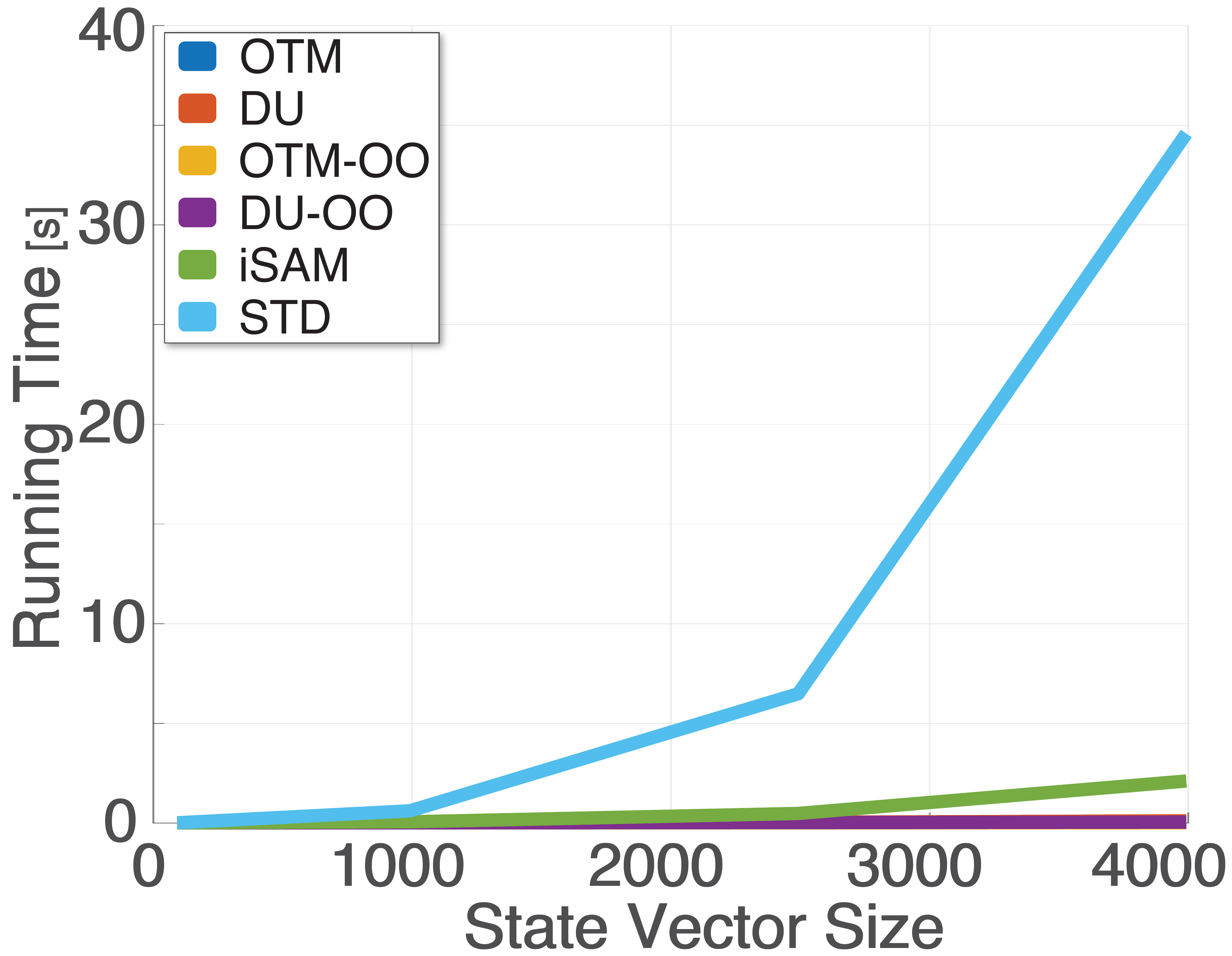}\label{fig:basic_Comp_all:100}}
        \subfloat[]{\includegraphics[bb={0 0 0 0},trim={0 0 0 0},clip, width=0.3\columnwidth]{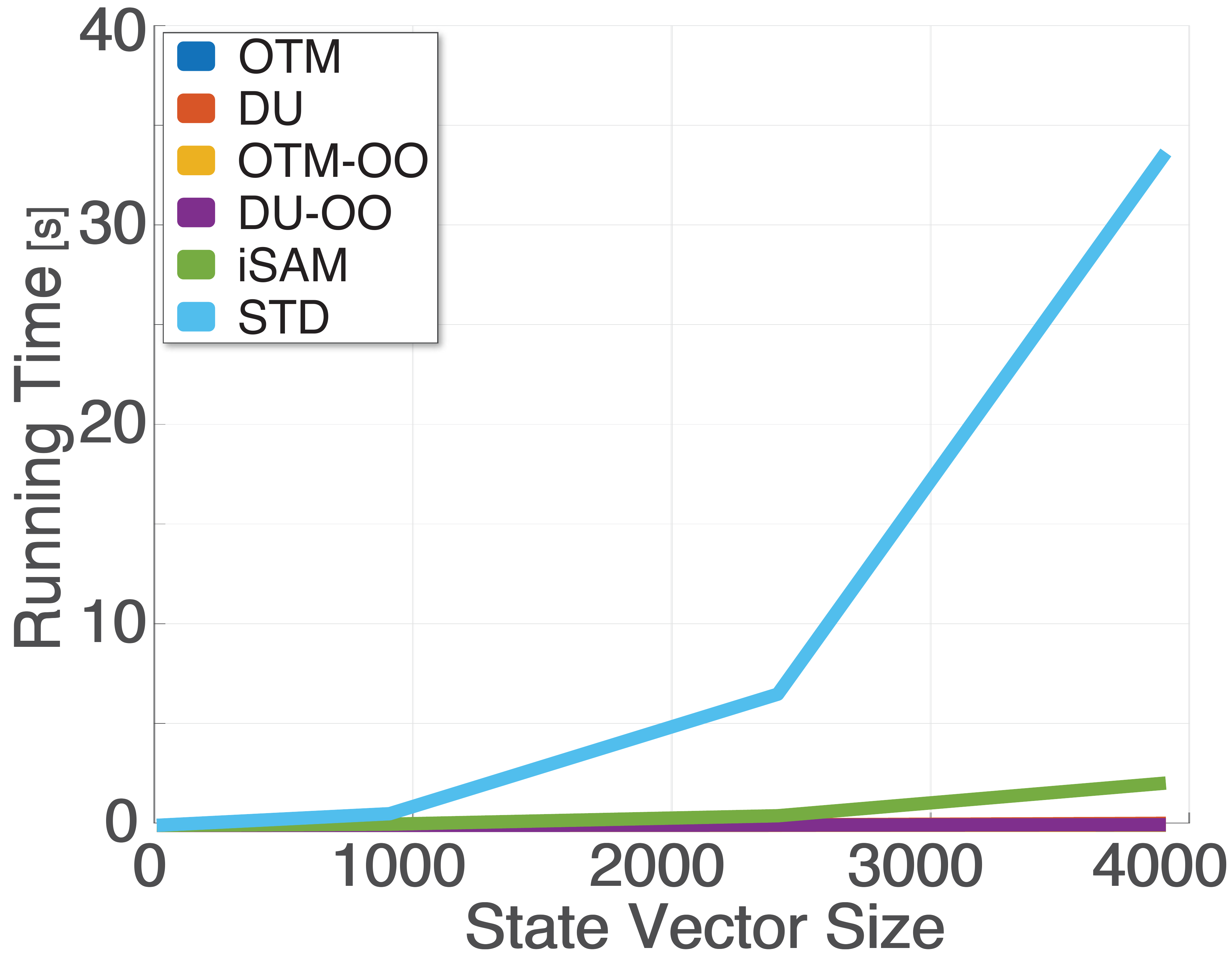}\label{fig:basic_Comp_all:200}} \\
        \subfloat[]{\includegraphics[bb={0 0 0 0},trim={0 0 0 0},clip, width=0.3\columnwidth]{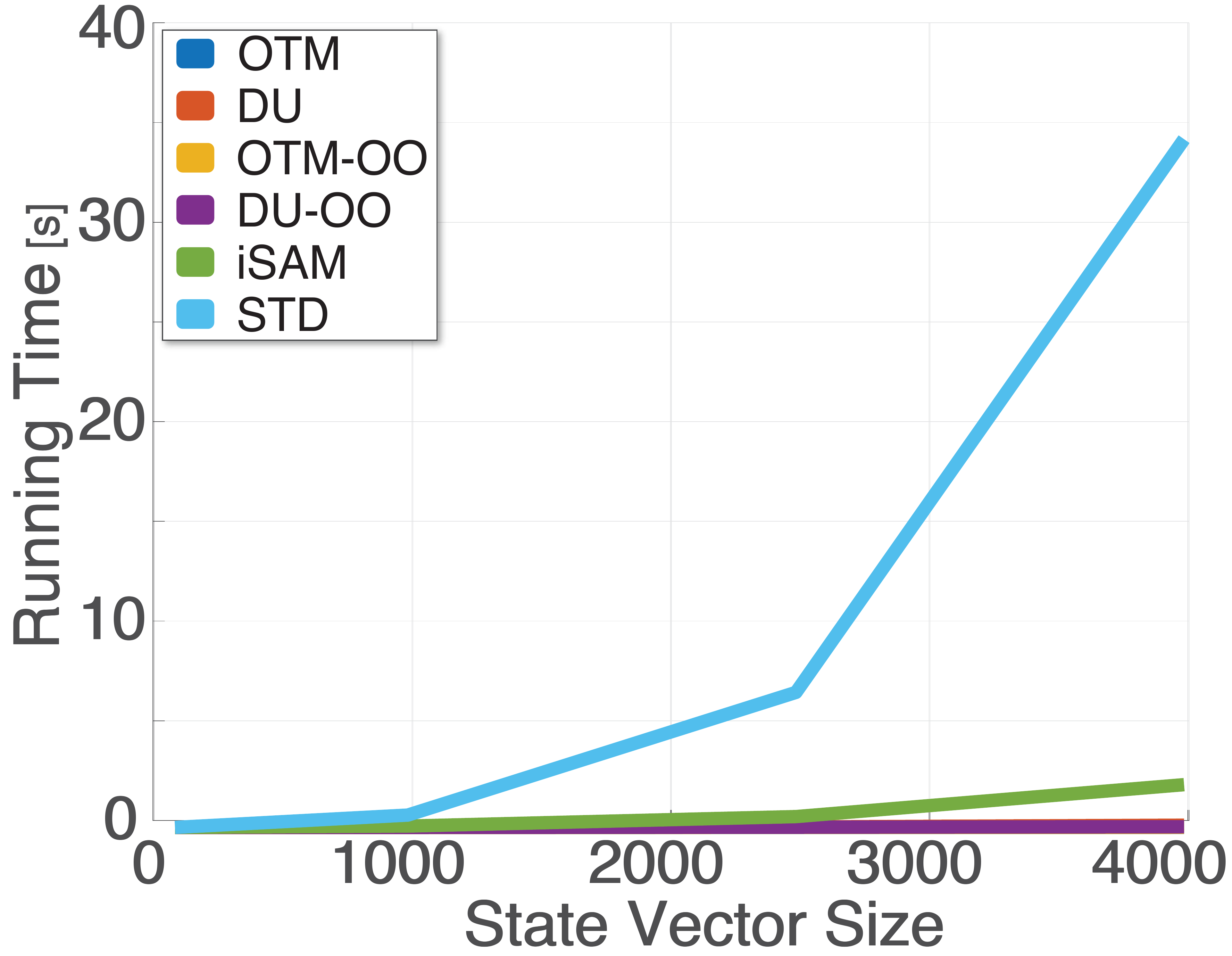}\label{fig:basic_Comp_all:300}}
        \subfloat[]{\includegraphics[bb={0 0 0 0},trim={0 0 0 0},clip, width=0.3\columnwidth]{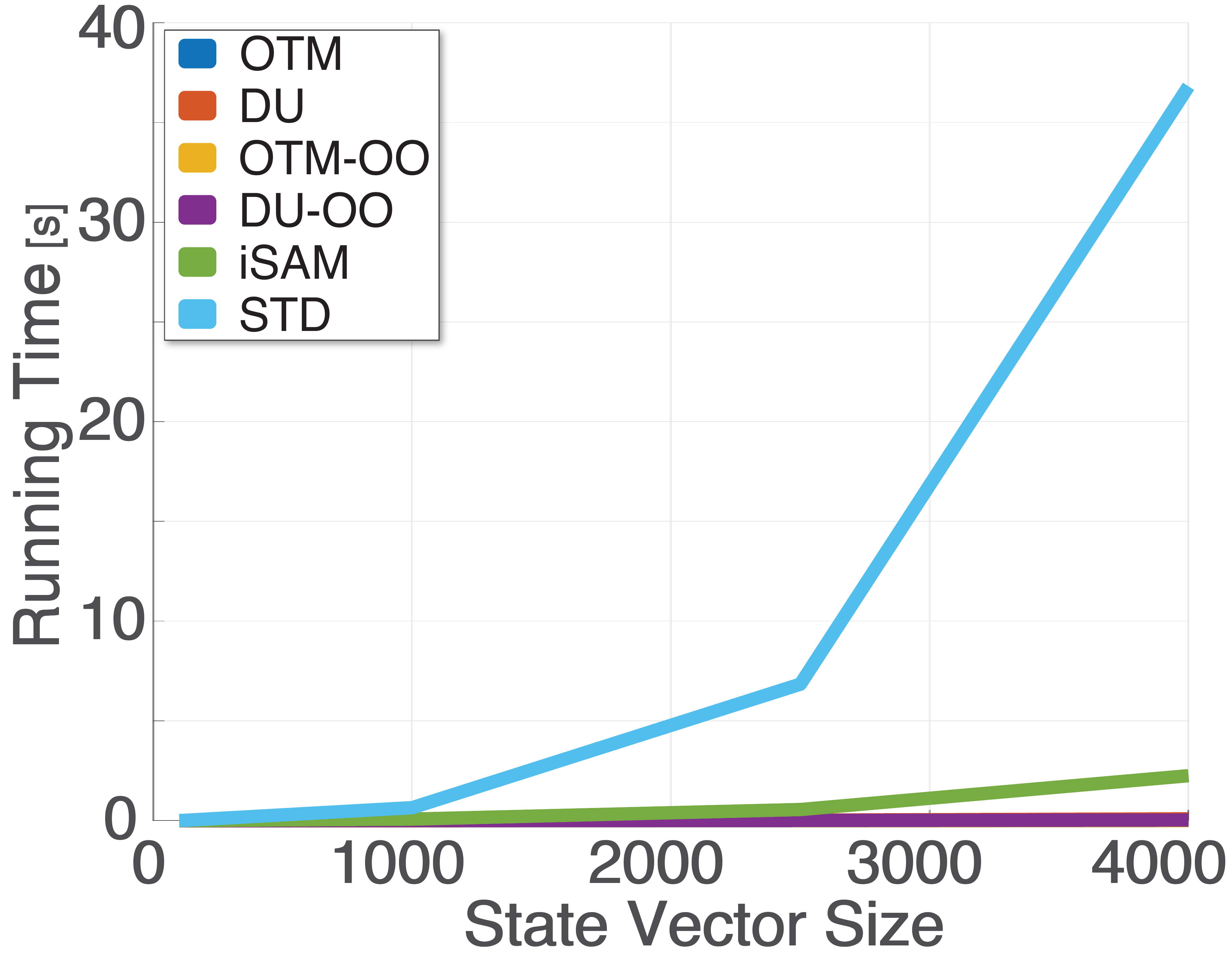}\label{fig:basic_Comp_all:400}}
        \subfloat[]{\includegraphics[bb={0 0 0 0},trim={0 0 0 0},clip, width=0.3\columnwidth]{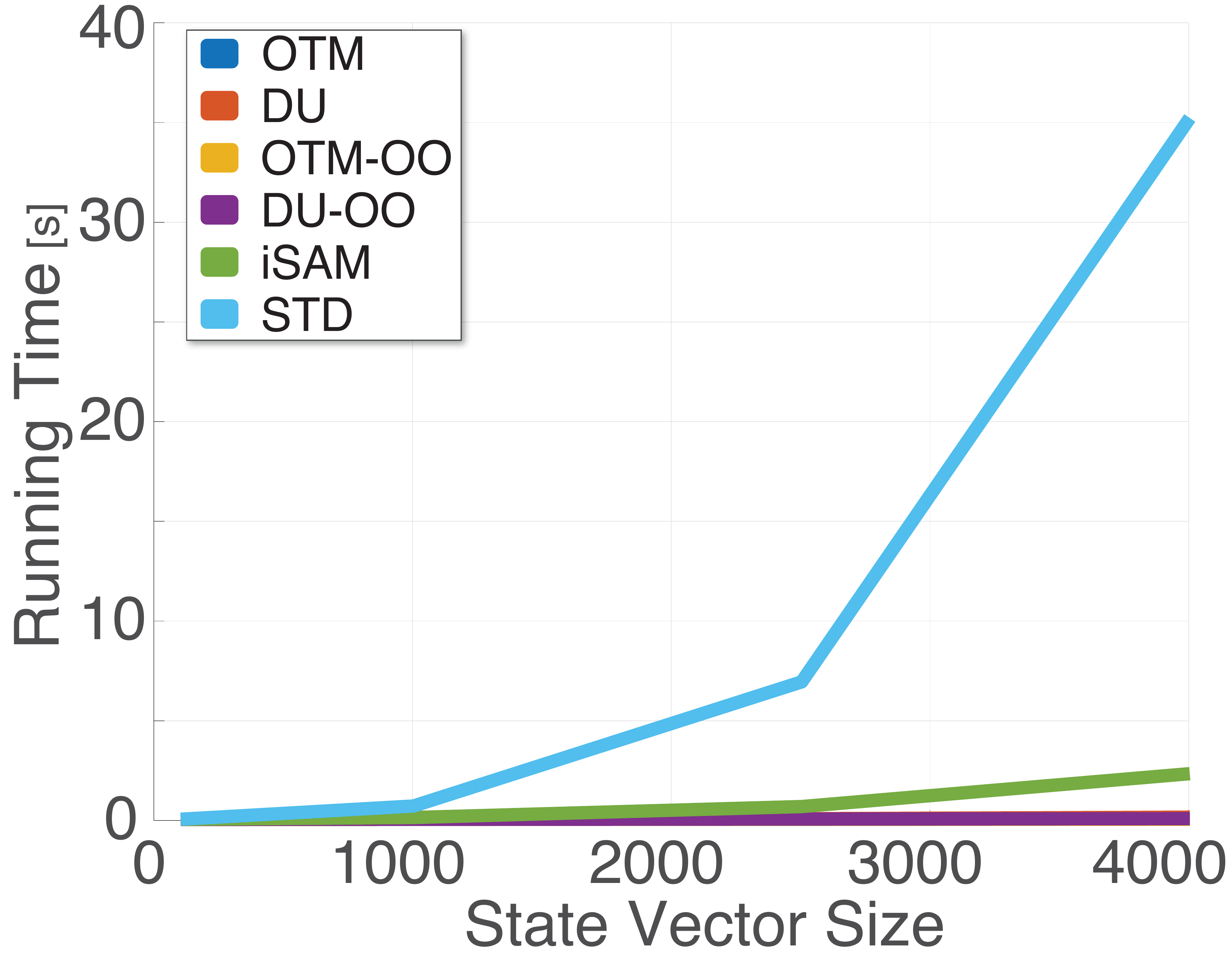}\label{fig:basic_Comp_all:500}}
        \caption{Method comparison through basic analysis simulation, checking sensitivity to new added measurements and the size of the inference state vector over all the tested methods i.e. \texttt{STD}, \texttt{iSAM} and our four methods, i.e. \texttt{OTM}, \texttt{UD}, \texttt{OTM-OO} and \texttt{UD-OO}. Each graph represents a different number for new rows added to the Jacobian matrix (a) 2 rows (b) 100 rows (c) 200 rows (d) 300 rows (e) 400 rows (f) 500 rows. Due to orders of magnitude issues we also provide zoom-in to our four methods in Figure~\ref{fig:basic_Comp_1t4}}
        \label{fig:basic_Comp_all}
\end{figure}
\begin{figure}
        \centering
        \subfloat[]{\includegraphics[bb={0 0 0 0},trim={0 0 0 0},clip, width=0.3\columnwidth]{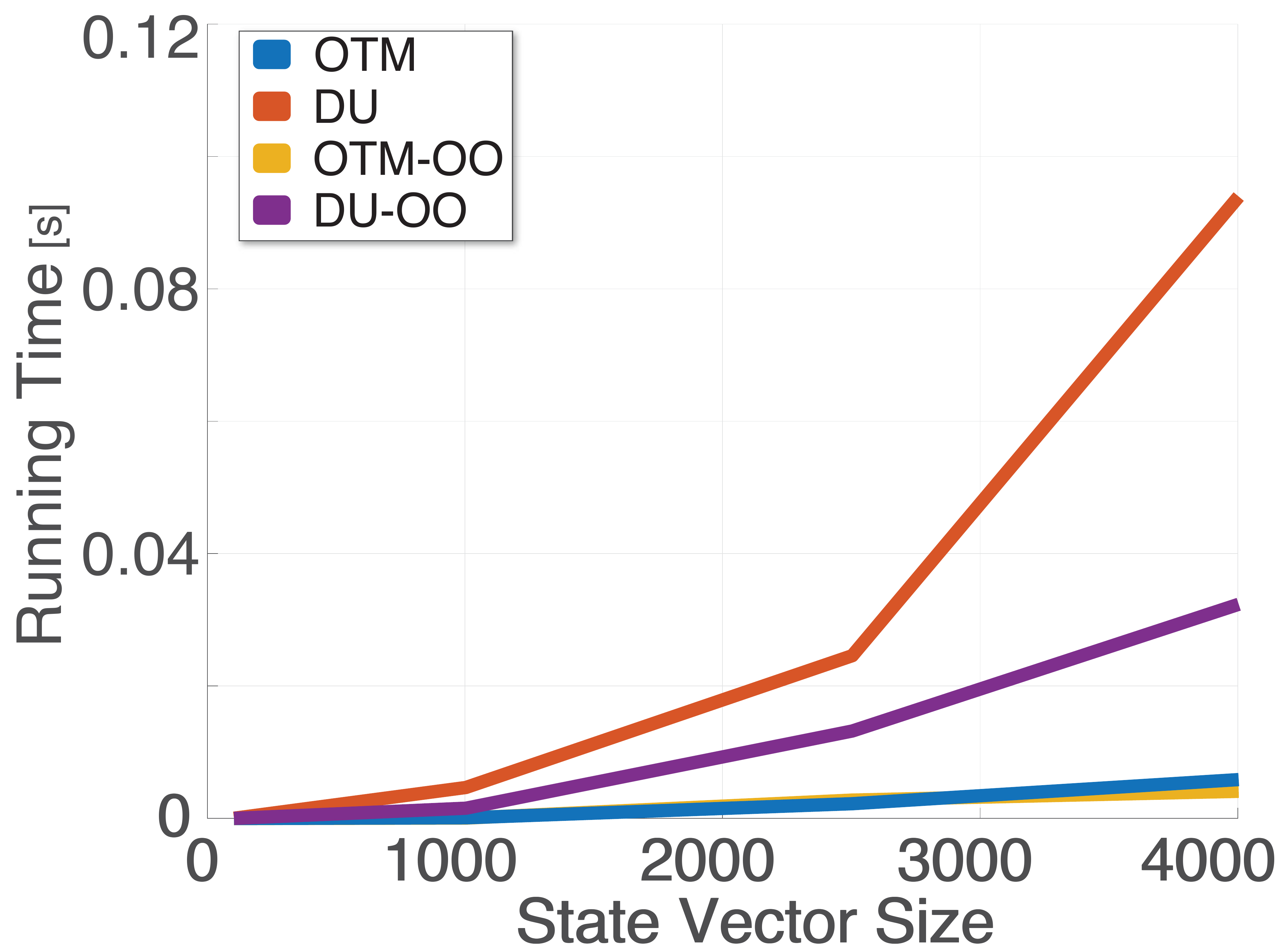}\label{fig:basic_Comp_1t4:2}}
        \subfloat[]{\includegraphics[bb={0 0 0 0},trim={0 0 0 0},clip, width=0.3\columnwidth]{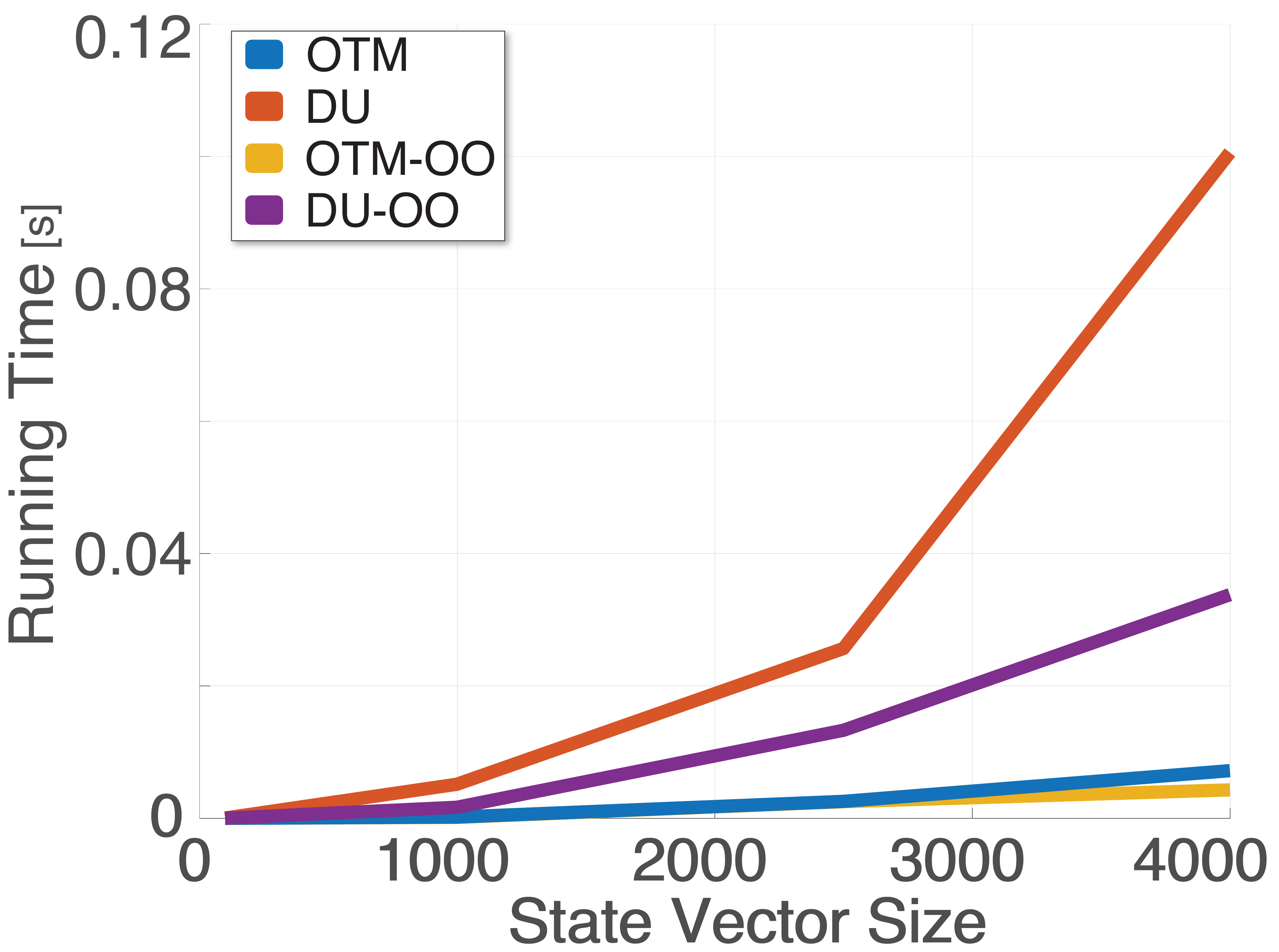}\label{fig:basic_Comp_1t4:100}}
        \subfloat[]{\includegraphics[bb={0 0 0 0},trim={0 0 0 0},clip, width=0.3\columnwidth]{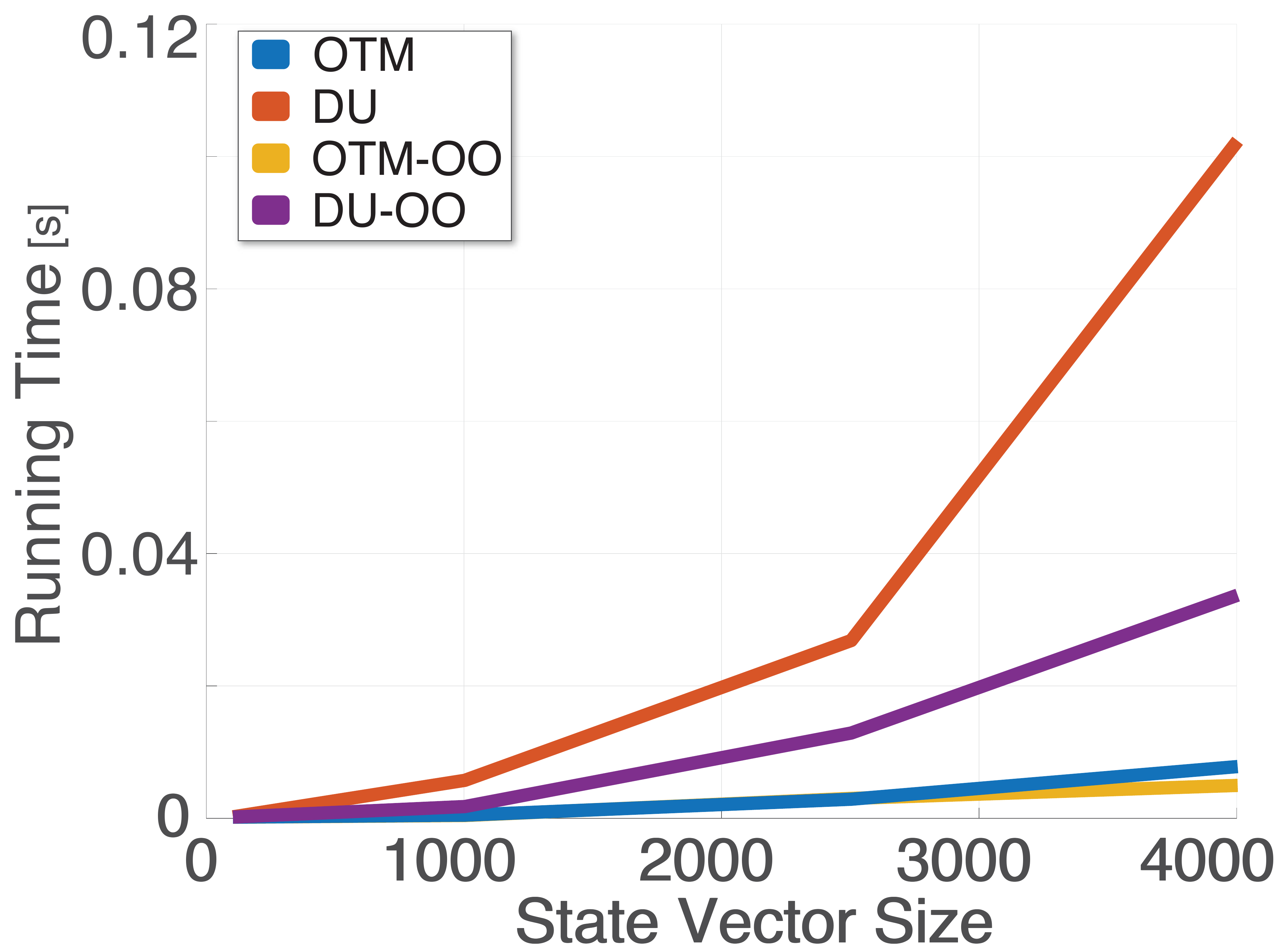}\label{fig:basic_Comp_1t4:200}} \\
        \subfloat[]{\includegraphics[bb={0 0 0 0},trim={0 0 0 0},clip, width=0.3\columnwidth]{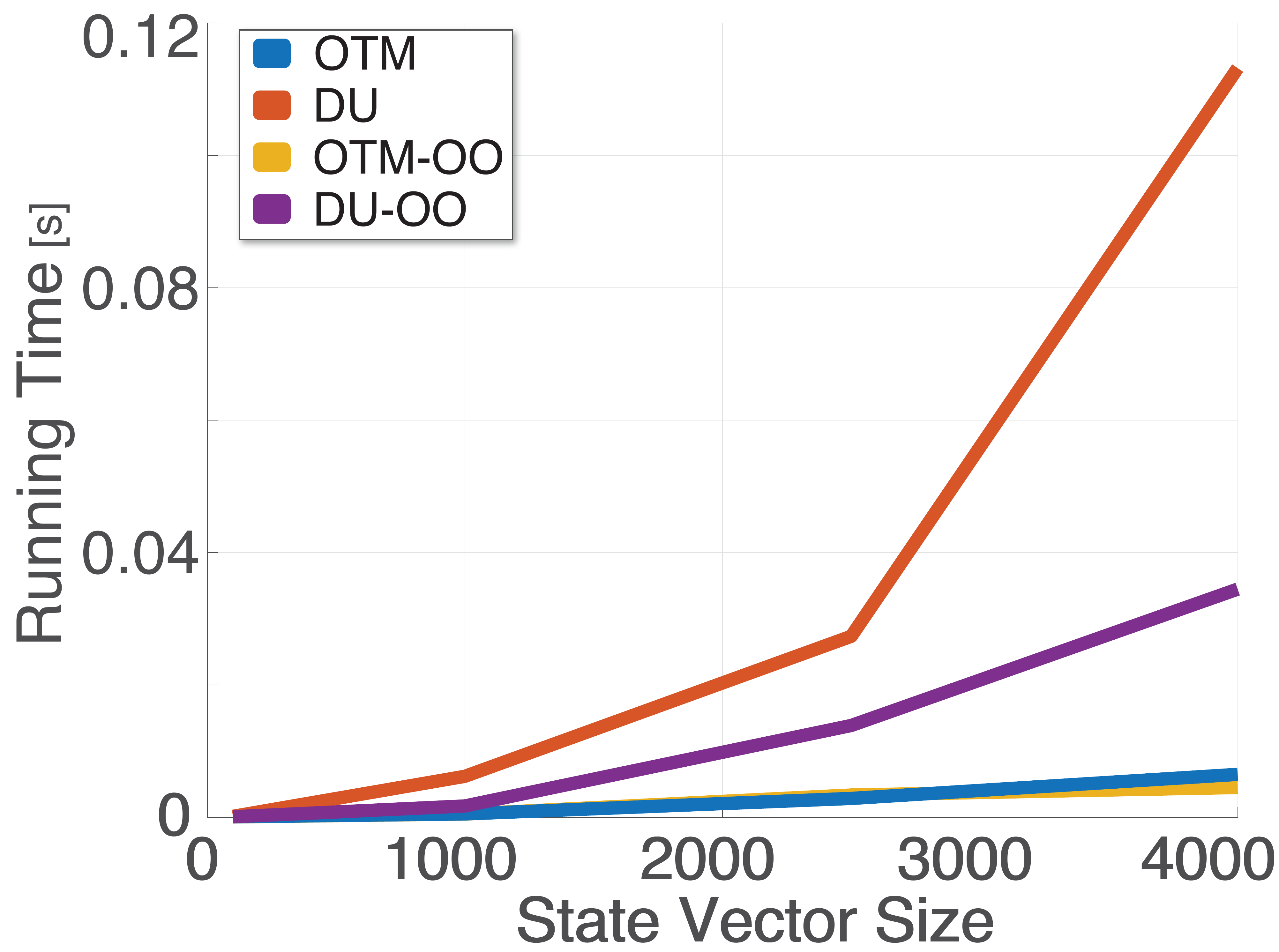}\label{fig:basic_Comp_1t4:300}}
        \subfloat[]{\includegraphics[bb={0 0 0 0},trim={0 0 0 0},clip, width=0.3\columnwidth]{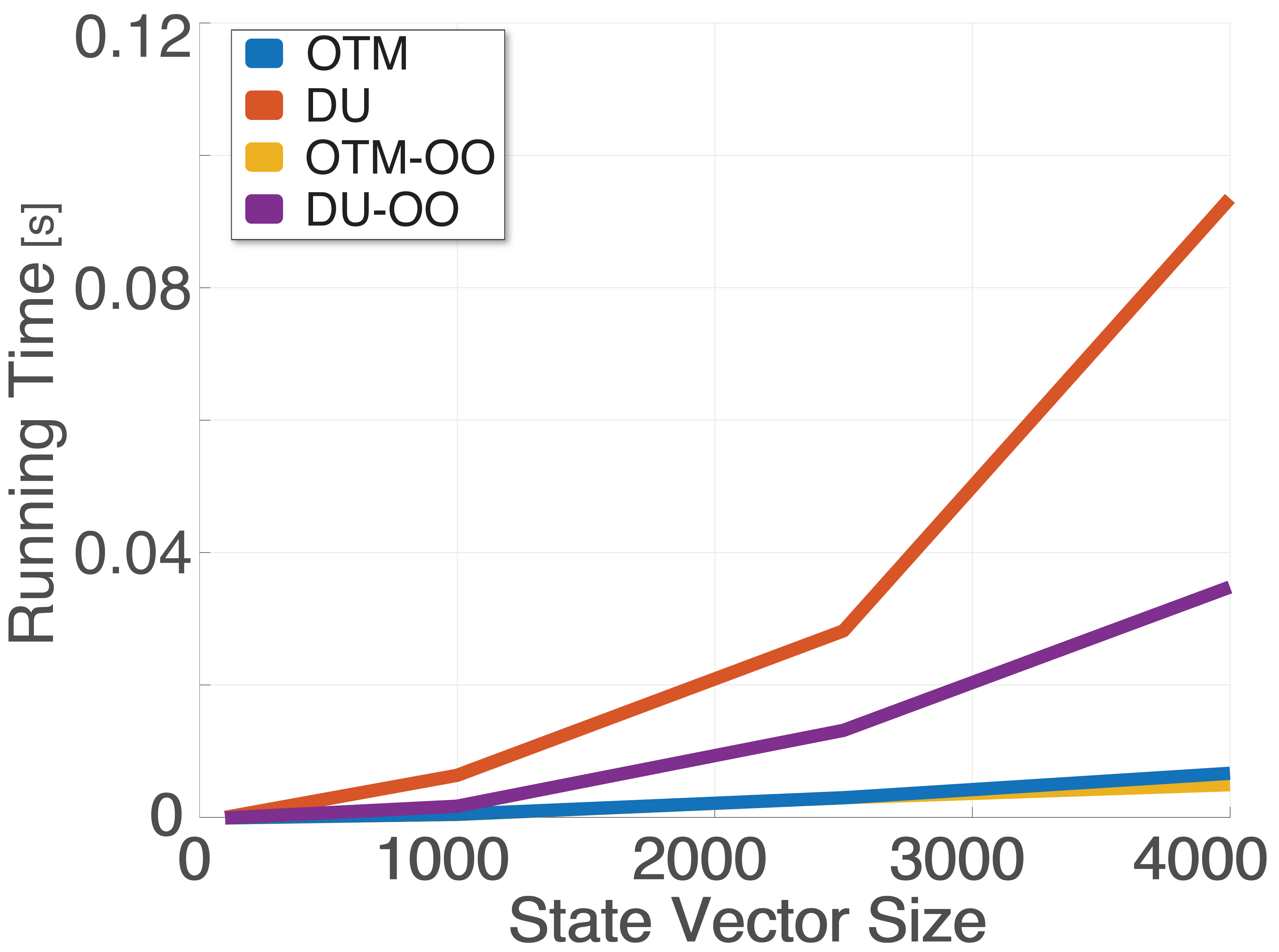}\label{fig:basic_Comp_1t4:400}}
        \subfloat[]{\includegraphics[bb={0 0 0 0},trim={0 0 0 0},clip, width=0.3\columnwidth]{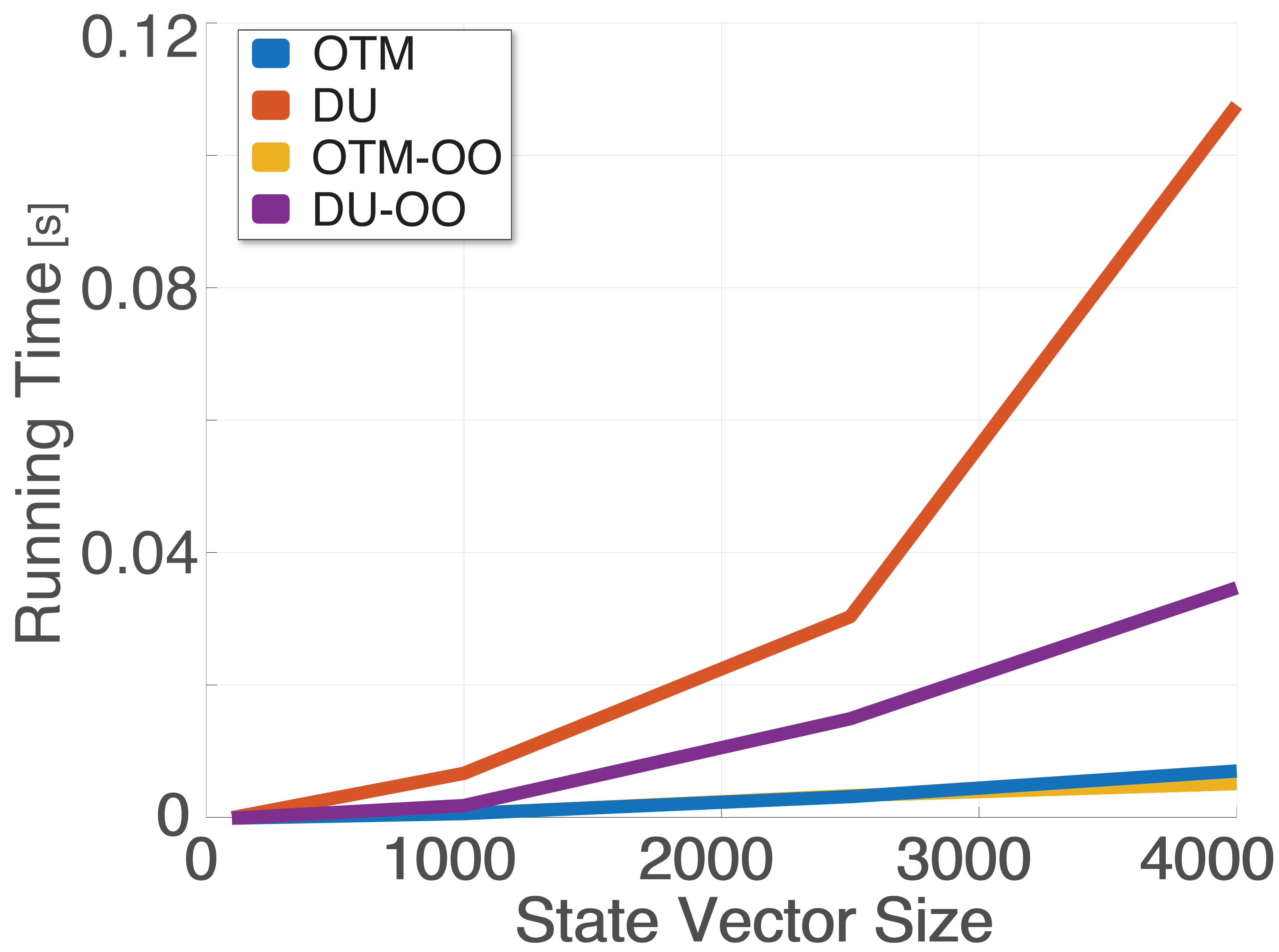}\label{fig:basic_Comp_1t4:500}}
        \caption{Zoom-in on Figure~\ref{fig:basic_Comp_all}, checking sensitivity to new added measurements and the size of the inference state vector over our four methods  i.e. \texttt{OTM}, \texttt{UD}, \texttt{OTM-OO} and \texttt{UD-OO}. Each graph represents a different number for new rows added to the Jacobian matrix (a) 2 rows (b) 100 rows (c) 200 rows (d) 300 rows (e) 400 rows (f) 500 rows. }
        \label{fig:basic_Comp_1t4}
\end{figure}
Figure~\ref{fig:basic_Comp_all} presents average timing results for all methods, while Figures~~\ref{fig:basic_Comp_all:2}~-~\ref{fig:basic_Comp_all:500} represent different number of new rows added to the Jacobian matrix (equivalent to adding new measurements), [2 100 200 300 400 500] respectively. After inspecting the results, we found that for all methods, running time is a non-linear, positive-gradient function of the inference state vector size and a linear function of the number of new measurements. Moreover, the running time dependency over the number of new measurements diminish as the inference state vector size grows. 
For all inspected parameters our methods score the lowest running time with a difference of up to \emph{three orders of magnitude} comparing to \texttt{iSAM}.

Figure~\ref{fig:basic_Comp_1t4} provides a zoom-in of Figure~\ref{fig:basic_Comp_all}, focusing on our suggested methods. Interestingly while we can clearly see that the \texttt{OTM} methodology is more efficient than the DU method, and the \texttt{DU-OO} is more efficient than \texttt{DU}, no such think can be said on \texttt{OTM} and \texttt{OTM-OO}. From inspecting Figures~~\ref{fig:basic_Comp_1t4:2}~-~\ref{fig:basic_Comp_1t4:500} we can see that up to a state vector size of about 2500 there is no visible difference between \texttt{OTM} and \texttt{OTM-OO} performance, while for larger sizes the latter slightly outperforms the former.

Thus scoring all methods from the fastest to the slowest with a time difference of \emph{four orders of magnitude} between the opposites:

\begin{center}
\texttt{OTM-OO} $\Rightarrow$ \texttt{OTM} $\Rightarrow$ \texttt{DU-OO} $\Rightarrow$ \texttt{DU}$\Rightarrow$ \texttt{iSAM}$\Rightarrow$ \texttt{STD}
\end{center}

\subsubsection{BSP in Unknown Environment - Consistent DA}\label{ssubsec:ConstDA}
\begin{figure}
        \centering
        \subfloat[]{\includegraphics[bb={0 0 0 0},trim={10 375 10 0},clip, width=0.45\columnwidth]{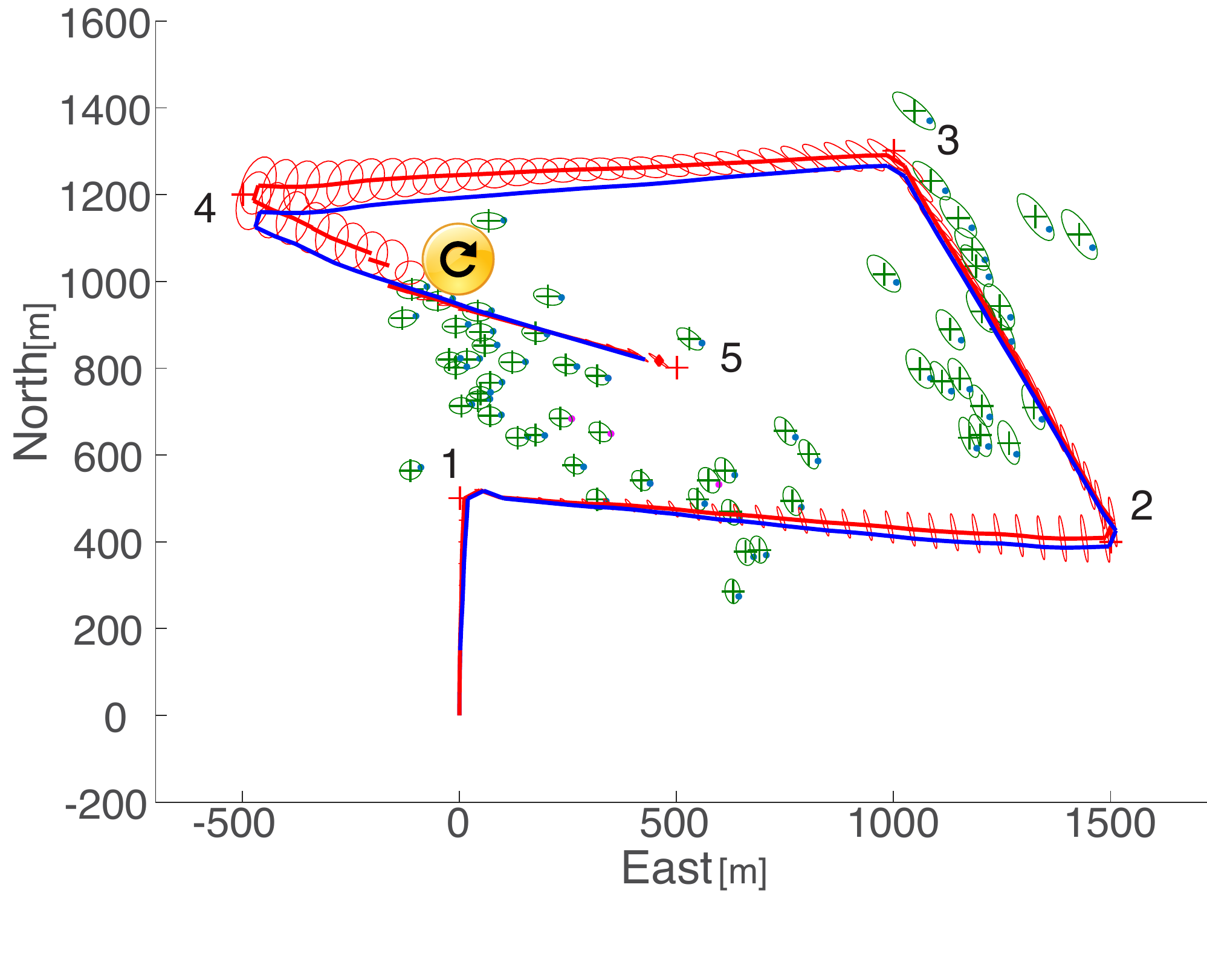}\label{fig:map_nolc}}
        \subfloat[]{\includegraphics[bb={0 0 0 0},trim={10 375 15 0},clip, width=0.45\columnwidth]{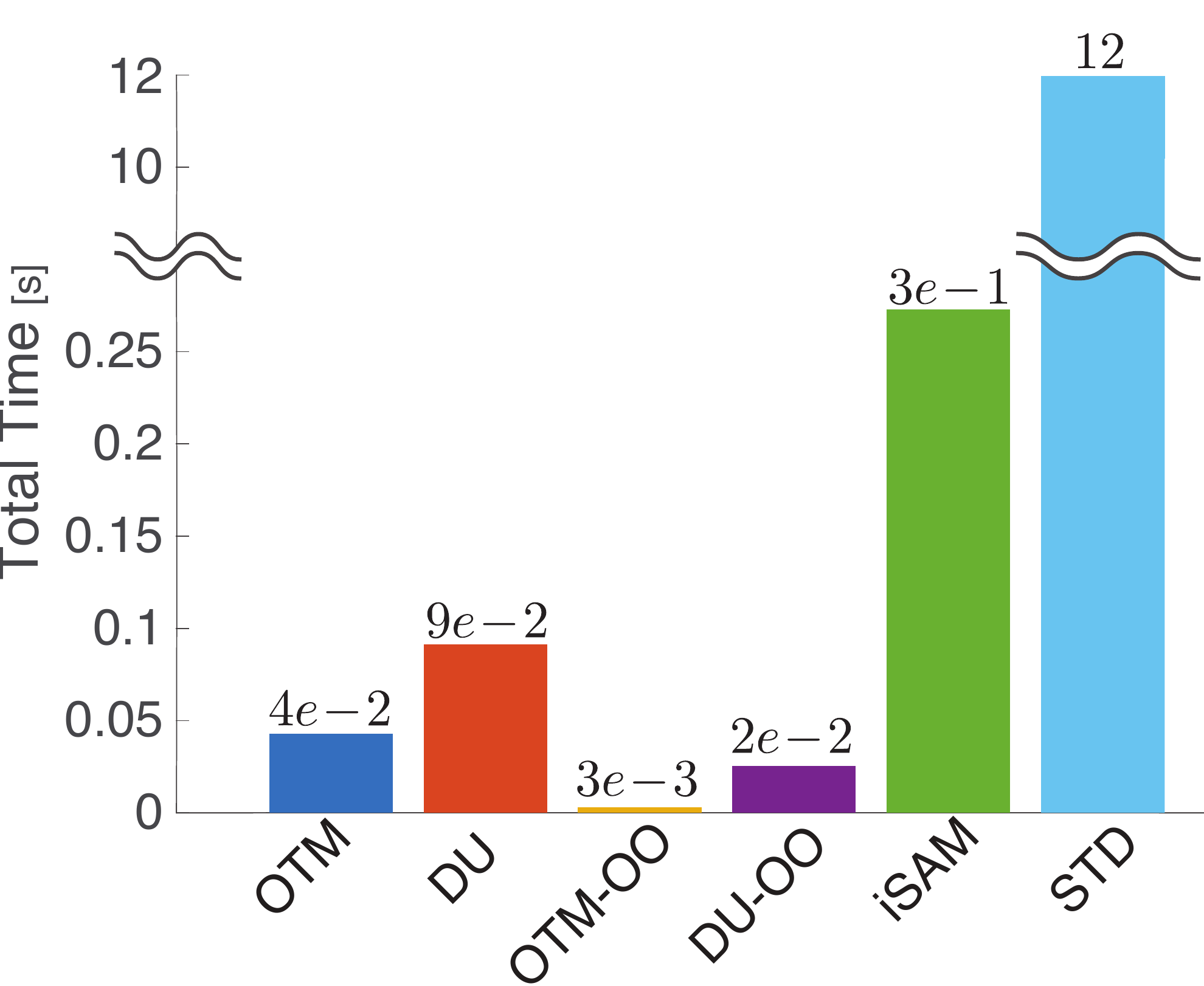}\label{fig:bar_nolc}}
        \caption{Second simulation layout and results: (a) The Synthetic Environment, where landmarks are marked in green, targets are numbered and marked with red crosses, the ground truth is denoted by a blue line, the estimated trajectory is denoted by a red line while the covariance is visualized by red ellipse (b) Total average running time of inference update for each method.}
        \label{fig:map1_bar}
\end{figure}

The purpose of this experiment is to further examine the suggested paradigm of \texttt{RUB inference}, in a real world scenario, under the simplifying assumption of consistent DA.  
The second simulation performs BSP over continuous action space, in an unknown synthetic environment. In contrast to Section~\ref{ssubsec:BasicAnalysis}, since now the synthetic environment replicates a real world scenario, the obtained information matrix is now sparse (e.g. Fig.~\ref{fig:FG_BT_Ab_Rd}). A robot was given five targets (see Figure~\ref{fig:map_nolc}) while all landmarks were a-priori unknown, and was required to visit all targets whilst not crossing a covariance value threshold. The largest loop closure in the trajectory of the robot, and the first in a series of large loop closures, is denoted by a yellow $\circlearrowright$ sign across all relevant graphs. 
The robot performs BSP over continuous action space, with a finite horizon of five look ahead steps \citep{Indelman15ijrr}. During the inference update stage each of the aforementioned methods were timed performing the first inference update step. It is worth mentioning that our paradigm is agnostic to the specific planning method or whether the action space is discrete or continuous. 

The presented running time is a result of an average between $10^3$ repetitions per step per method. Similarly to Section~\ref{ssubsec:BasicAnalysis}, as can be seen in Figure~\ref{fig:bar_nolc}, the suggested MATLAB implemented methods are up to \emph{two orders of magnitude} faster than \texttt{iSAM} used in a MATLAB C++ wrapper.
Interestingly, the use of sparse information matrices changed the methods' timing hierarchy. While \texttt{OTM-OO} still has the best timing results ($3 \!\times\! 10^{-3}$ sec), \emph{two orders of magnitude faster than} \texttt{iSAM}, \texttt{OTM} and \texttt{DU-OO} switched places. So the timing hierarchy from fastest to slowest is: 

\begin{center}
\texttt{OTM-OO}$\Rightarrow$ \texttt{DU-OO} $\Rightarrow$ \texttt{OTM} $\Rightarrow$ \texttt{DU}$\Rightarrow$ \texttt{iSAM}$\Rightarrow$ \texttt{STD}
\end{center}
\begin{figure}
        \centering
        \subfloat[]{\includegraphics[bb={0 0 0 0},trim={5 300 5 0},clip, width=0.33\columnwidth]{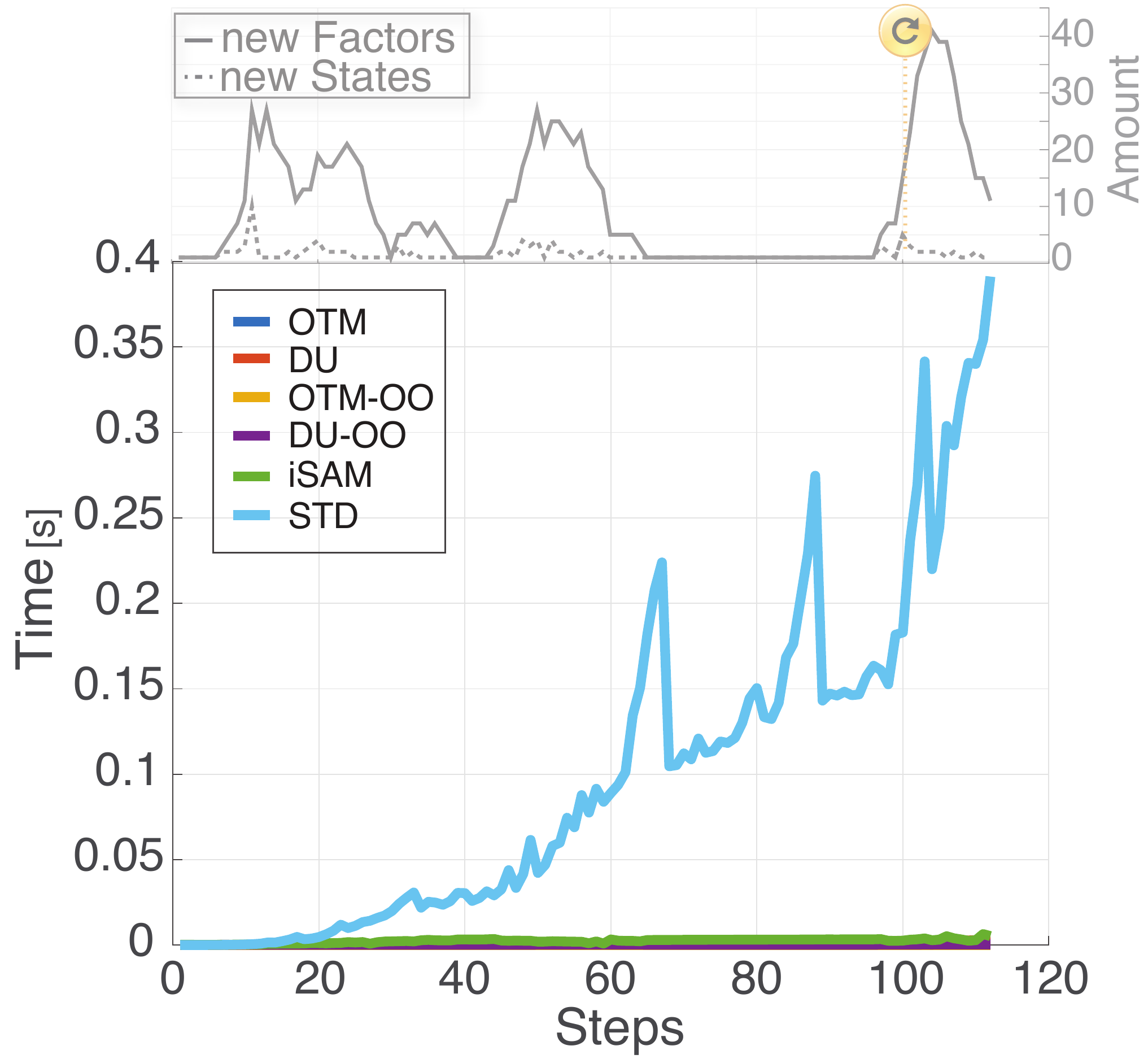}\label{fig:bspfg_Comp_all}}
        \subfloat[]{\includegraphics[bb={0 0 0 0},trim={5 295 5 0},clip, width=0.33\columnwidth]{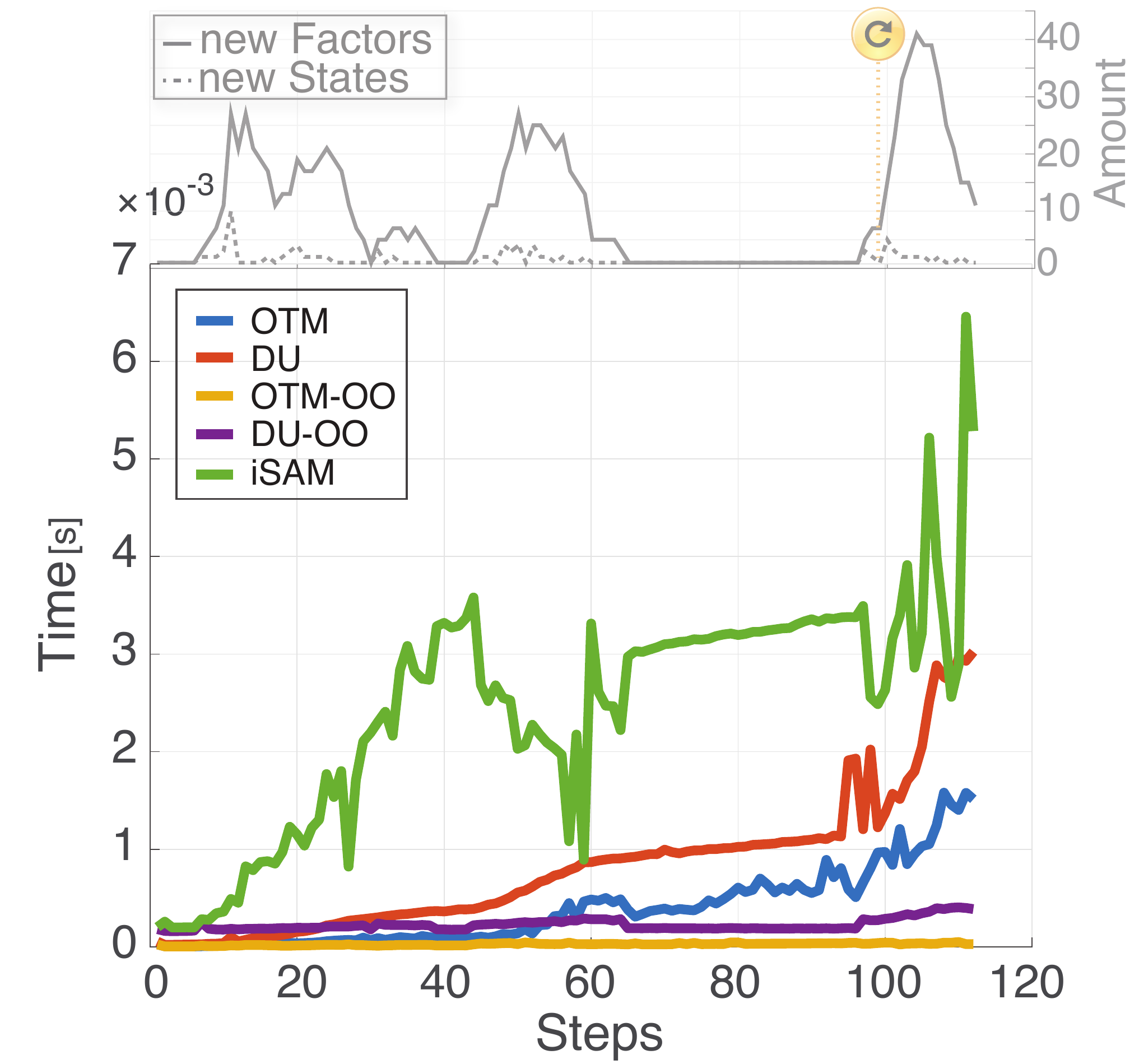}\label{fig:bspfg_Comp_1t46}}
        \subfloat[]{\includegraphics[bb={0 0 0 0},trim={5 290 5 0},clip, width=0.33\columnwidth]{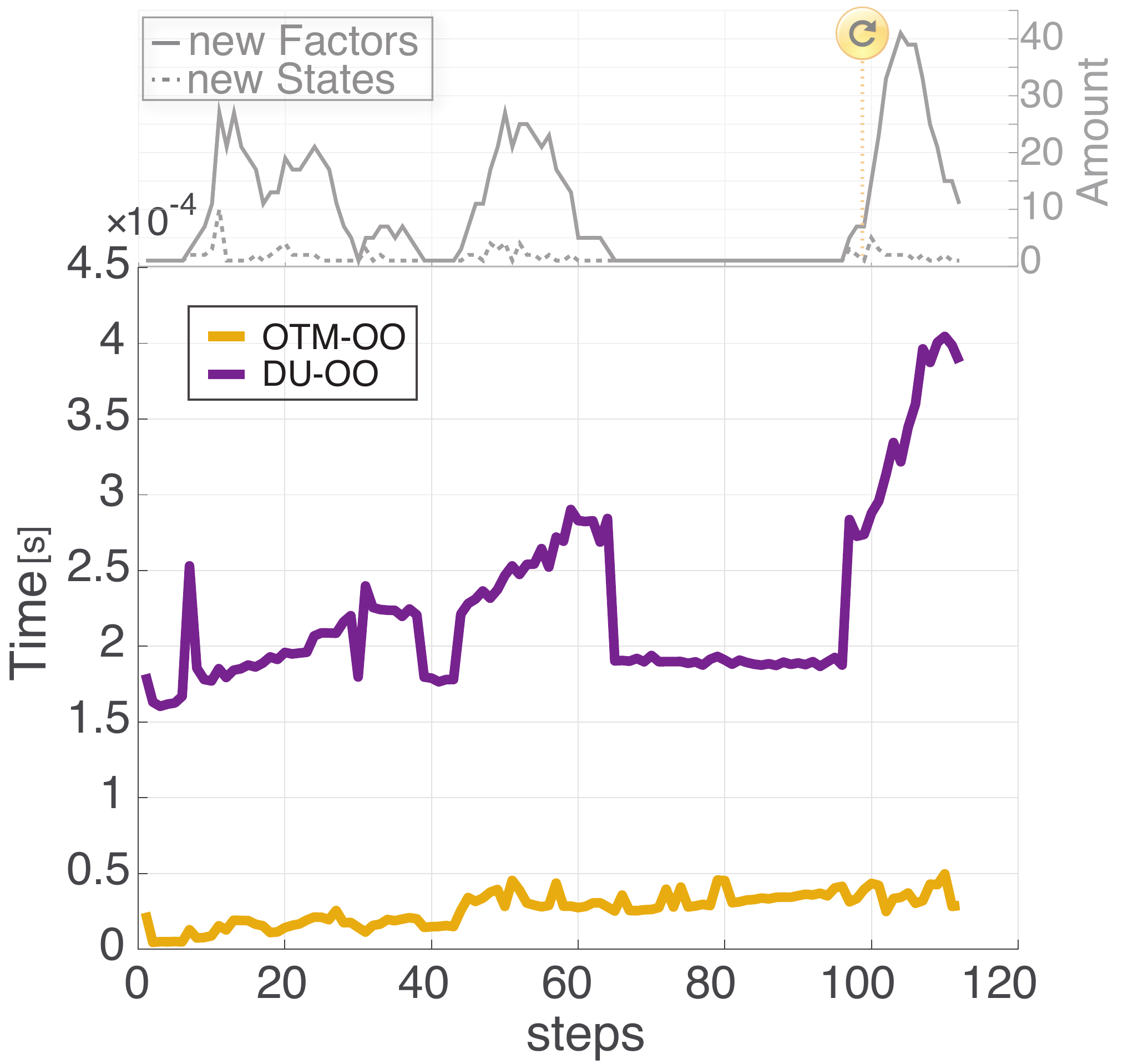}\label{fig:bspfg_Comp_34}}
        \caption{Second simulation timing results for the scenario presented in Figure~\ref{fig:map_nolc}. Upper part of each graph provides indication on new factors and new states per computation step while the lower presents the methods timing results: (a) All six methods (b) \texttt{OTM}, \texttt{DU}, \texttt{OTM-OO}, \texttt{DU-OO}and \texttt{iSAM} methods (c) \texttt{OTM-OO} and \texttt{DU-OO} methods. } 
        \label{fig:methodComp_sim2}
\end{figure}

After demonstrating the use of our novel paradigm drastically reduce cumulative running time, we continue on to showing that in a few aspects it is also less sensitive.
Figure~\ref{fig:methodComp_sim2} presents the performance results of each of the methods per simulation step. The upper graphs presents the number of new factors and new states per each step, while the lower graph presents the average running time of each method as a function of the simulation step. The $\circlearrowright$ sign, represents the first largest loop closure in a series of large loop closures.
While some of the behavior presented in Figure~\ref{fig:methodComp_sim2} can be related to machine noise, from carefully inspecting Figure~\ref{fig:methodComp_sim2}, alongside the trajectory of the robot in Figure~\ref{fig:map_nolc}, a few interesting observations can still be made.
The first observation relates to the "flat line" area noticeable in the upper graph of Figure~\ref{fig:bspfg_Comp_1t46} between time steps $60-90$. This time steps range is equivalent to the path between the third and fourth targets, were the only factor added to the belief is motion based. As a result, a single new state (the new pose) is presented to the belief, along with a single motion factor. In this range, the timing results of \texttt{iSAM} \texttt{DU} and \texttt{OTM} present a linear behavior with a relatively small gradient. This gradient is attributed to the computational effort of introducing a single factor, containing a new state, to the belief. While the vertical difference between the aforementioned can be attributed to the sensitivity of each method to the number of states and factors in the belief. 

From this observation, we can try to better understand the reason for the substantial time difference between the methods.
Basing a method on \texttt{RUB inference}, rather than on standard Bayesian inference, will not magically change the computational impact of introducing factors or new states to the belief. However, because \texttt{RUB inference} is re-using calculations from precursory planning, the computational burden is being "paid" once, rather than twice as in the standard Bayesian inference. For the simple example of strictly motion propagation, since this motion based factor has already been introduced during precursory planning, under \texttt{RUB inference} it offers no additional computational burden. In the same manner, the reason \texttt{RUB inference} is less sensitive to the state dimensionality originates in calculations re-use. Under incremental update performed by \texttt{iSAM}, the state dimension is mostly noticeable when in need of re-ordering and/or re-eliminating states. Although same mechanisms also affect \texttt{RUB inference}, our method avoids them whenever they were adequately performed during the precursory planning, thus reducing inference computation time. 

Another interesting observation refers to "pure" loop closures, were there are measurements with no addition of new variables to the state vector, i.e. measurements to previously observed landmarks. For the case of "pure" loop closures, \texttt{STD}, \texttt{iSAM} and the \texttt{DU} based methods (i.e. \texttt{DU} and \texttt{DU-OO}) experienced the largest timing spikes throughout the trajectory of each method while both \texttt{OTM} based methods experienced minor spikes if any. 

By introducing the \texttt{OO} methodology to both \texttt{DU} and \texttt{OTM}, we drastically reduce the methods sensitivity to the motion propagation e.g.~the once-positive gradient line in \texttt{DU} during time steps $60-90$, turned into a flat line in \texttt{DU-OO} as can easily be seen in Figure~\ref{fig:bspfg_Comp_34}. 
Moreover, while both \texttt{DU} and \texttt{OTM} present some sensitivity to different occurrences, i.e. the size of the state vector, new measurements and loop closures, this sensitivity is drastically reduced by introducing the \texttt{OO} methodology, e.g. \texttt{OTM-OO} is basically a flat line throughout the simulation as can easily be seen in Figure~\ref{fig:methodComp_sim2}.

In conclusion, our methods, based on \texttt{RUB inference}, particularly \texttt{OTM-OO}, seem to be more resilient to large loop closures that were already detected during planning, state vector size, belief size, number of newly added measurements or even the combination of the aforementioned. 

\subsubsection{BSP in unknown Environment - Relaxing Consistent DA Assumption}\label{ssubsec:inConstDA}
\begin{figure}
        \centering
        \subfloat[]{\includegraphics[bb={0 0 0 0},trim={0 0 0 0},clip, width=0.5\columnwidth]{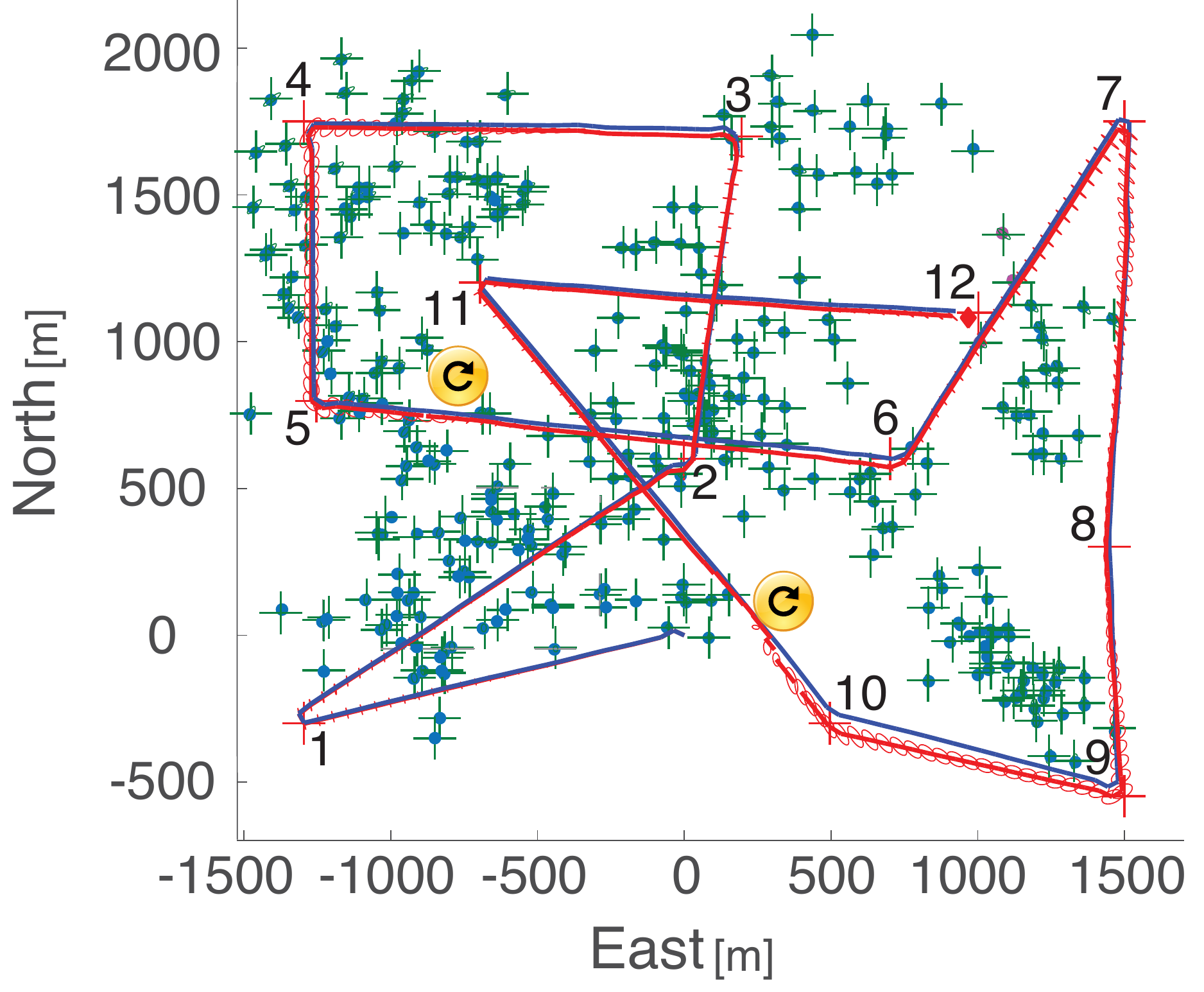}\label{fig:map}}
        \subfloat[]{\includegraphics[bb={0 0 0 0},trim={0 -33 0 0},clip, width=0.5\columnwidth]{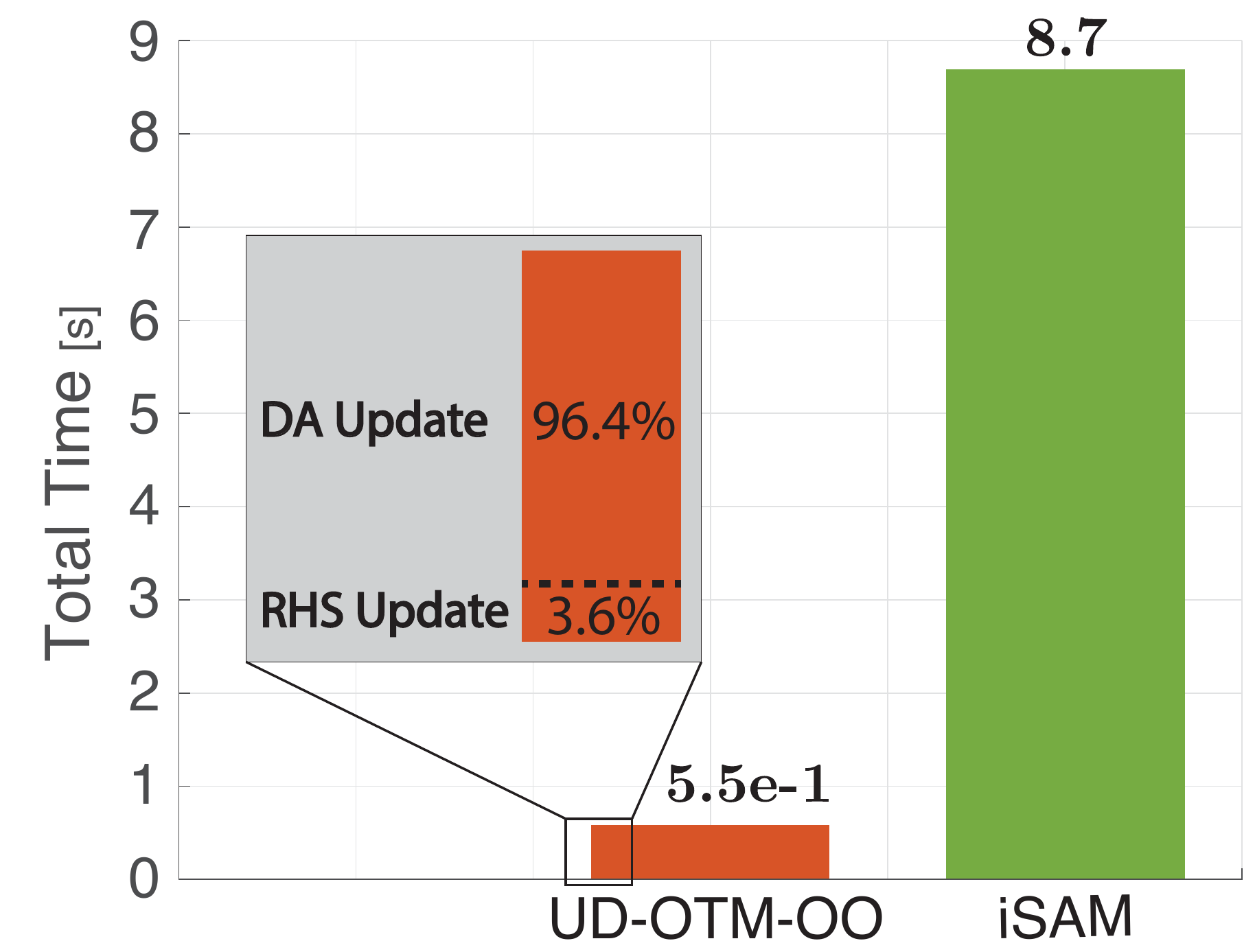}\label{fig:tot_time}}
        \caption{ Simulation layout and results: (a) The Synthetic Environment, where landmarks are marked in green, targets are numbered and marked with red crosses, the ground truth is denoted by a blue line, the estimated trajectory is denoted by a red line while the covariance is visualized by red ellipse. (b) Total average running time of inference update for each method, when 50\% of the steps were with inconsistent DA.}
        \label{fig:map_bar}
\end{figure}

The purpose of this experiment is to further examine the suggested paradigm of \texttt{RUB inference}, in a real world scenario, while relaxing the simplifying assumption of consistent DA.
The third simulation performs BSP over continuous action space, in an unknown synthetic environment. A robot was given twelve targets (see~Figure~\ref{fig:map}) while all landmarks were a-priori unknown, and was required to visit all targets whilst not crossing a covariance value threshold. 
The experiments presented in Sections~\ref{ssubsec:BasicAnalysis} and \ref{ssubsec:ConstDA} were based on the simplifying assumption of consistent DA between inference and precursory planning, which can often be violated in real world scenarios. 
In this simulation we relax this restricting assumption and test our novel paradigm under the more general case were DA might be inconsistent.

The main reason for inconsistent data association lies in the perturbations caused by imperfect system and environment models. These perturbations increase the likelihood of inconsistent DA between inference and precursory planning. While the planning paradigm uses state estimation to decide on future associations, the further it is from the ground truth the more likely for inconsistent DA to be received. This imperfection is modeled by formulating uncertainty in all models (see Section~\ref{sec:problem-formulation}).

For a more conservative comparison, in addition to the aforementioned, we force inconsistent DA between inference and precursory planning for all new variables. 
In contrast to planning paradigms that can provide DA to new variables, in addition to an unknown map, the robot's planning paradigm considers only previously-mapped landmarks. As a result of this limitation, the DA received from the planning stage can not offer new landmarks to the state vector. 
Consequently, each new landmark would essentially mean facing inconsistent DA, while the single scenario in which a consistent DA is obtained (see~Section~\ref{ssubsec:DAtypes}), occurs when both planning and inference are considering the same known landmark.
Both perturbations caused by uncertainty and considering only previously mapped landmarks, resulted in just 50\% DA consistency between planning and succeeding inference in this experiment.

\begin{figure}
\centering
\subfloat[]{\includegraphics[bb={0 0 0 0},trim={0 0 0 -45},clip, width=0.5\textwidth]{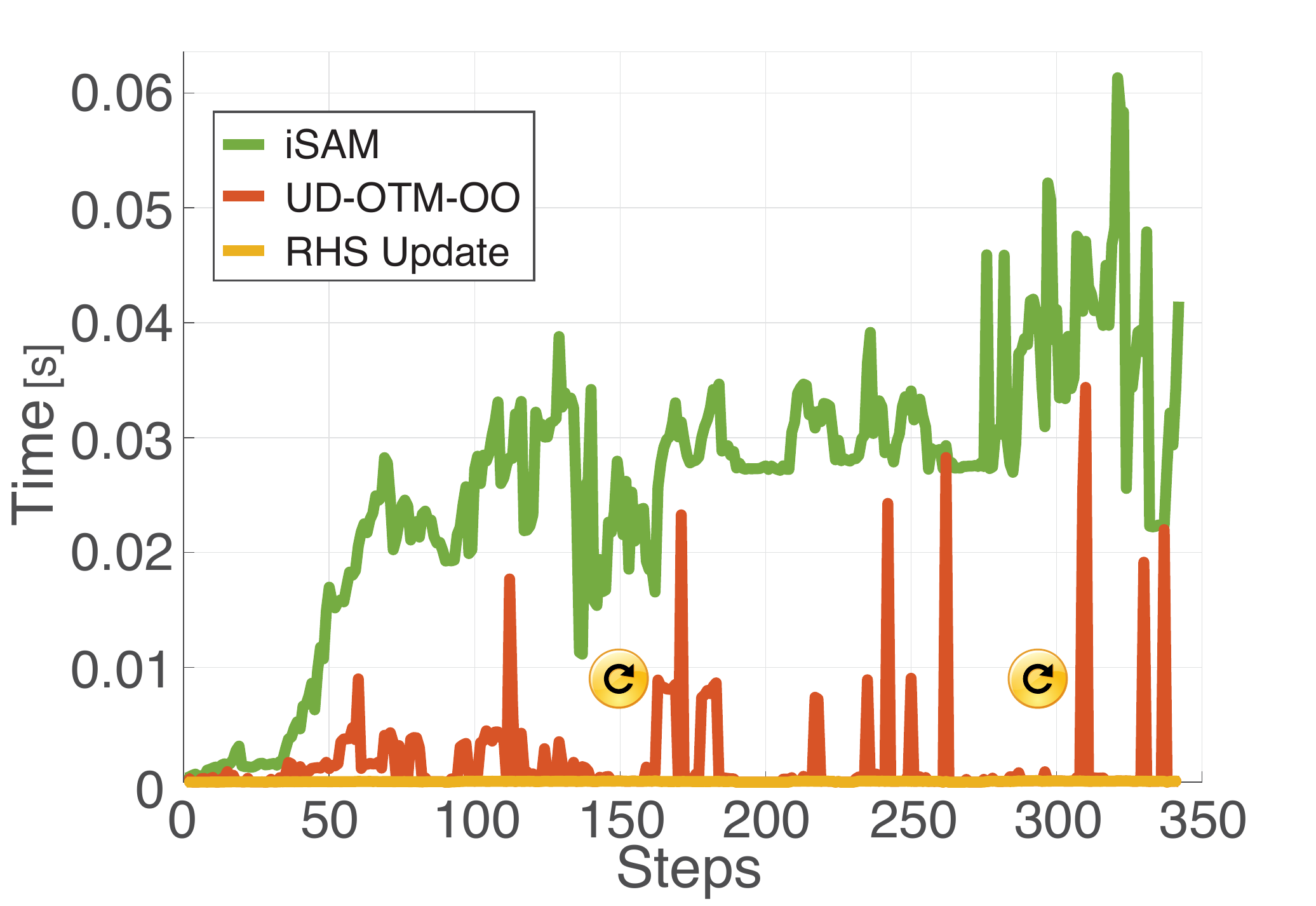}\label{fig:comp_time}}
\subfloat[]{\includegraphics[bb={0 0 0 0},trim={0 215 0 0},clip, width=0.5\textwidth]{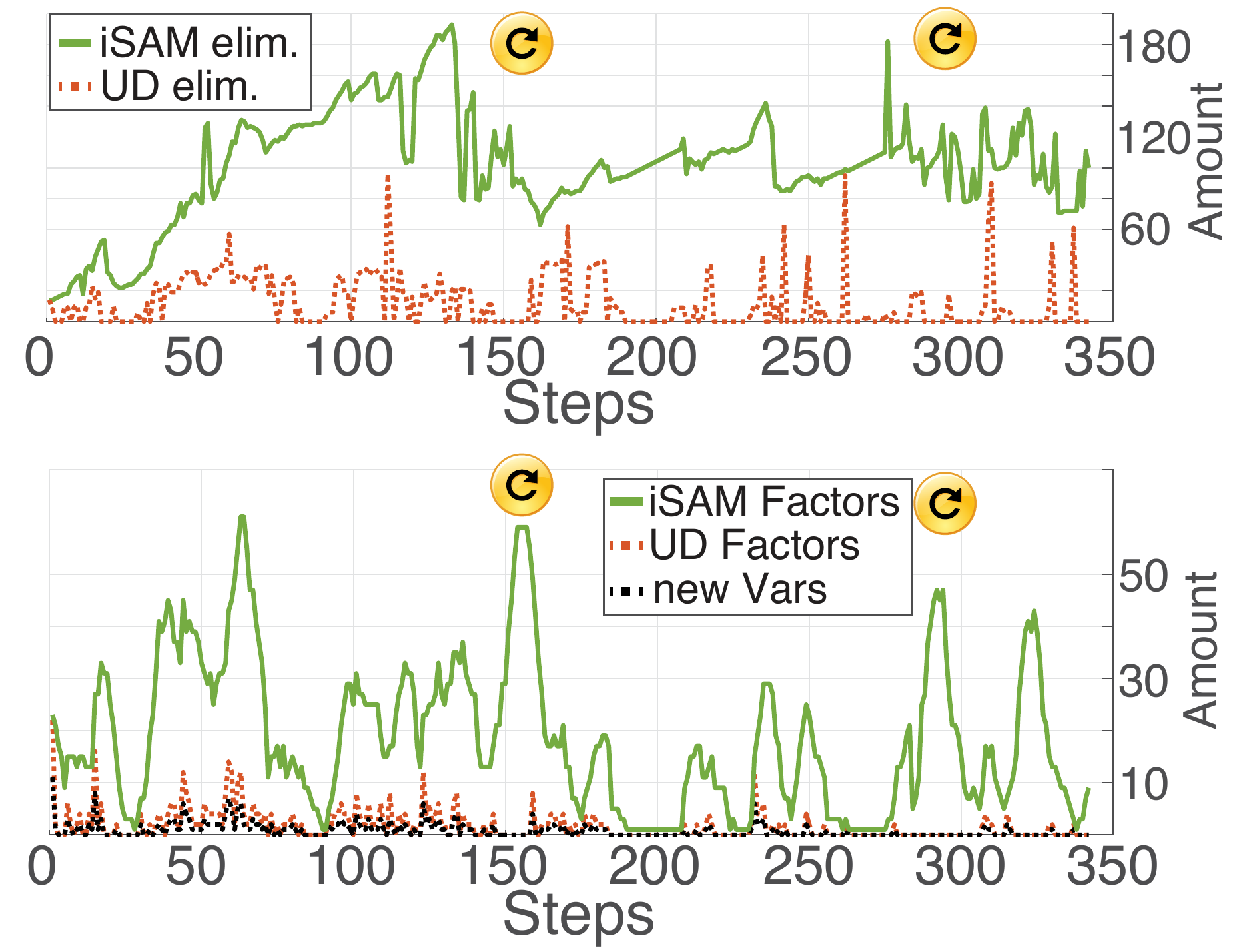}\label{fig:comp_reElim}}\\
\subfloat[]{\includegraphics[bb={0 0 0 0},trim={0 0 0 190},clip, width=0.5\textwidth]{Figures/PerStepFac.pdf}\label{fig:comp_fac}}
\caption{Per-step analysis of the simulation presented in Figure~\ref{fig:map_bar}. In 50\% of the steps, planning and succeeding inference are with consistent DA: (a) Per-step timing results of \texttt{iSAM} performing standard Beysian inference in green, \texttt{UD-OTM-OO} performing \texttt{RUB inference} in orange and the RHS update portion out of \texttt{UD-OTM-OO} in yellow. (b) Number of eliminations per-step, in the inference update stage for both \texttt{iSAM} and \texttt{UD-OTM-OO}. (c) Number of newly added factors in \texttt{iSAM} per step, newly added factors in \texttt{UD-OTM-OO} per step, and the number of new variables introduced to the belief per step.}
\label{fig:analysis}
\end{figure}

Following the findings of Section~\ref{ssubsec:ConstDA}, out of the four suggested methods we choose to continue the comparison just with the \texttt{OTM-OO} method. 
While \texttt{OTM-OO} assumed consistent DA, the more general approach deals with inconsistent DA before updating the RHS vector. We denote the complete approach, updating DA followed by \texttt{OTM-OO}, as \texttt{UD-OTM-OO}, where \texttt{UD} stands for Update Data association. 
It is important to clarify that \texttt{UD-OTM-OO} and for consistent DA also \texttt{OTM-OO}, yield the same estimation accuracy as \texttt{iSAM}, since the inference update using \texttt{RUB inference} results in the same topological graph with the same values. Such comparison will be presented later on using a real-world data in Section~\ref{subsec:kitti}. For that reason, the accuracy aspect will not be discussed further in this section.
While the scenario presented in Figure~\ref{fig:map} contains at least ten large loop closures, for the readers convenience we marked two of them using yellow $\circlearrowright$ signs. Same loop closures are also marked in Figure~\ref{fig:analysis} for comparison.

 Figure~\ref{fig:tot_time} presents the cumulative computation time of the inference update phase throughout the simulation. We can see that the majority of \texttt{UD-OTM-OO} computation time, i.e. $96.4\%$, is dedicated to DA update while only $3.6\%$ for updating the RHS vector. Although the need for DA update increased running time (as to be expected), \texttt{UD-OTM-OO} still outperforms \texttt{iSAM} by an order of magnitude. 
  
In addition to the improvement in total computation time of the inference update stage, we continue on analyzing the "per step" behavior of \texttt{UD-OTM-OO}, and demonstrate that in a few aspects it is less sensitive than \texttt{iSAM}.
 Figure~\ref{fig:comp_time} presents per step computation time of both \texttt{UD-OTM-OO} and \texttt{iSAM}, as well as the RHS update running time of \texttt{UD-OTM-OO}. Our suggested paradigm not only outperforms \texttt{iSAM} in the cumulative computation time, but also outperforms it for each individual step. 
While Figure~\ref{fig:comp_time} presents the difference in average computation time per-step, Figures~\ref{fig:comp_reElim} and \ref{fig:comp_fac} capture the reason for this difference as suggested in Section~\ref{ssubsec:ConstDA}.
Figure~\ref{fig:comp_fac} presents the number of added factors in \texttt{iSAM} denoted by a green line, as opposed to in \texttt{UD-OTM-OO} denoted by an orange line, and the number of new variables per step denoted by a black line. Figure~\ref{fig:comp_reElim} presents the number of eliminations made during inference update in both methods. Number of eliminations reflects the number of involved variables in the process of converting FG into a BT (see Appendix-B and Algorithm~\ref{alg:DaUpdate} line~\ref{alg:DaUpdate:invBT_upd} for the equivalent processes in \texttt{iSAM} and \texttt{UD-OTM-OO} accordingly).

After carefully inspecting both figures, alongside the robot's trajectory in Figure~\ref{fig:map}, the following observations can be made. 
Even with the limitation over the planning paradigm, both the number of new factors added and the number of re-eliminations during the inference update stage, are substantially smaller than their \texttt{iSAM} counterparts. These large differences are some of the reasons for \texttt{UD-OTM-OO}'s better performance.  
Due to the limitation over the planning paradigm, new observation factors (i.e. new landmarks added each step) in both \texttt{iSAM} and \texttt{UD-OTM-OO} are identical. While in \texttt{iSAM} new observation factors constitute a small fraction of total factors, for \texttt{UD-OTM-OO}, they constitute more than half of total factors. 
After comparing the re-elimination graph with the timing results for each of the methods, it appears both trends and peaks align, so we assume \texttt{UD-OTM-OO} as well as \texttt{iSAM} to be mostly sensitive to the amount of re-eliminations (further analysis is required).

Both re-elimination and added factors amounts, can be further reduced by smart reordering and relaxing the limitation over the planning paradigm accordingly.

As observed in Section~\ref{ssubsec:ConstDA}, our method seems to be more resilient to loop closures. By inspecting the yellow $\circlearrowright$ signs in Figure~\ref{fig:comp_fac}, we can see that in both cases, \texttt{iSAM} introduce around 50 factors of previously known variables (i.e. the black line representing new variables is zeroed), while \texttt{UD-OTM-OO} introduces no factors at all. These two loop closure examples beautifully demonstrate the advantage of using \texttt{RUB Inference}. For cases of consistent or partially consistent DA, when encountering a loop closure (i.e. observing a previously mapped landmark) our method saves valuable computation time since loop closures are only calculated once, in the planning stage (e.g. see timing response for loop closure at the appropriate yellow $\circlearrowright$ signs in Figure~\ref{fig:comp_time}).

Our method also seems to be less sensitive to state dimensionality. Inspecting steps $192-208$ and $263-275$ in Figure~\ref{fig:comp_fac}, we observe there are no new factors, i.e. the computation time is a result of motion factors; inspecting Figure~\ref{fig:comp_time} we observe that in spite of the aforementioned, \texttt{iSAM} computation time is much larger than our method. From this comparison we can infer our suggested method is less sensitive to state dimensionality. As explained earlier, this originates in the reduced number of re-eliminations and state re-ordering in \texttt{RUB inference} when compared to \texttt{iSAM}, e.g. when the amount of re-eliminations in Figure~\ref{fig:comp_reElim} is almost the same between the two (like in steps $171$, $262$, $310$), the equivalent computation time in Figure~\ref{fig:comp_time} is also almost identical.

\subsection{Real-World Experiment Using KITTI Dataset} \label{subsec:kitti}
After the promising performance in a simulated environment, we tested our paradigm for inference update via BSP in a real-world environment using KITTI dataset \citep{Geiger13ijrr}.
The KITTI dataset, recorded in the city of Karlsruhe, contains stereo images, Laser scans and GPS data. For this work, we used the raw images of the left stereo camera, from the Residential category file: 2011\_10\_03\_drive\_0027, as measurements, as well as the supplied ground truth for comparison. 

In this experiment we consider a robot, equipped with a single monocular camera, performing Active Full-SLAM in the previously unknown streets of Karlsruhe Germany. The robot starts with a prior over its initial pose and with no prior over the environment. At time $k$ the robot executes BSP on the single step action sequence taken in the KITTI dataset at time $k+1$. 
At the end of each BSP session, the robot executes the chosen action, and receives measurements from the KITTI dataset.  
Inference update is then being performed in two separate approaches, the first following the standard Bayesian inference approach and the second following our proposed \texttt{RUB inference} approach. The inference update following each is compared for computation time and accuracy.

The following sections explain in-detail how planning and perception are being executed in this experiment. 

\subsubsection{Experiment Parameters}
For the readers convenience, this section covers all the parameters used for this experiment and were not provided by KITTI. 
\begin{center}
  \begin{tabular}{ l | c }
    \hline
    Prior belief  standard deviation & $\begin{bmatrix}
    	1^o \cdot I_{3 x 3} & 0 \\
    	0 & 1_{[m]}\cdot I_{3 x 3}
    \end{bmatrix}$ \\ \hline
    Motion Model standard deviation & $\begin{bmatrix}
    	0.5^o \cdot I_{3 x 3} & 0 \\
    	0 & 0.5_{[m]}\cdot I_{3 x 3}
    \end{bmatrix}$  \\ \hline
    Observation Model standard deviation & $\begin{bmatrix}
    	1_{[px]}  & 0 \\
    	0 & 1_{[px]}
    \end{bmatrix}$  \\ \hline
    Camera Aperture & $90^o$  \\ \hline
    Camera acceptable Sensing Range & between $2_{[m]}$ and $40_{[m]}$ \\ 
    \hline
  \end{tabular}
\end{center}

The motion and observation models that were used are (\ref{eq:motion_mod}) and (\ref{eq:obs_mod}) appropriately, where $h(.)$ is given by the pinhole camera model, and the zero mean Gaussian noise is stated above.

\subsubsection{Planning using KITTI dataset}\label{ssubsec:kitti:planning}

Our proposed approach for \texttt{RUB inference}, leverages calculations made in the precursory planning phase to update inference more efficiently. KITTI is a pre-recorded dataset with a single action sequence, i.e.~the "future" actions of the robot are pre-determined. Nevertheless, we can still evaluate our approach by appropriately simulating the calculations that would be performed within BSP for that specific (and chosen) single action sequence. In other words,  BSP involves belief propagation and objective function evaluations for different candidate actions, followed by identifying the best action via Eq.~(\ref{eq:OptimalControl}) and its execution. 

In our case, the performed actions over time are readily available; hence, we only focus on the corresponding future beliefs for such actions given the partial information available to the robot at planning time. Specifically, at each time instant $k$, we construct the future belief $b[X_{k+1|k}]$ via Eq.~(\ref{eq:bsp_recursive}) using the supplied visual odometry as motion model and future landmark observations. Future landmark observations are generated by considering only landmarks projected within the camera field of view using MAP estimates for landmark positions and camera pose from the propagated belief $b[X_{k+1|k}]$. 
As in this work the planning phase considers only the already-mapped landmarks, without reasoning about expected new landmarks, each new landmark observation in inference would essentially mean facing inconsistent DA. 

To conclude, planning using the KITTI dataset is simulated over a single action in the following manner: current belief is propagated with future action, future measurements are generated by considering already-mapped landmarks, and future belief is solved. Since the "optimal" action is pre-determined by the KITTI dataset there is no need for an objective function evaluation.

\subsubsection{Perception using KITTI dataset}\label{ssubsec:kitti:preception}
After executing the next action, the robot receives a corresponding raw image from the KITTI dataset. The image is being processed through a standard vision pipeline, which produces features with corresponding descriptors \citep{Lowe04ijcv}.
Landmark triangulation is being made after the same feature has been observed at least twice, while following different standard conditions designed to filter outliers. Once a feature is triangulated, it is considered as a landmark, and is added as a new state to the belief.
Note that the robot has access only to its current joint belief, consisting  of the estimated landmark locations, and the robot past and present pose estimations.
Once the observation factors (\ref{eq:obs_mod}) are added to the belief, the inference update is being made in two different and separate ways. The first, used for comparison, follows the standard Bayesian inference, by using the efficient methodologies of \texttt{iSAM2} in order to update inference. The belief of the preceding inference $b[X_{k|k}]$ is being updated with the new motion $\prob{x_{k+1}|x_{k},u_k}$ and observation factors $\underset{j \in \mathcal{M}_{k+1|k+1}}{\prod} \!\!\!\!\!\!\!\! \prob{z^j_{k+1}|x_{k+1},l_j}$, thus obtaining $b[X_{k+1|k+1}]$.

The second method follows our proposed paradigm for \texttt{RUB inference}. The belief from the preceding planning phase, $b[X_{k+1|k}]$, which corresponds to $u_{k|k+1}$ (see~(\ref{eq:performed_action})), is updated with the new measurements. This update is done using \texttt{UD-OTM-OO} which consists of two stages, first using our DA update method (Section~\ref{ssubsec:updateDA}) which updates the predicted DA to the actual DA, followed by the \texttt{OTM-OO} method (Section~\ref{subsubsec:otmoo}) which updates measurement values.
 
\subsubsection{Results - KITTI dataset}\label{ssubsec:kitti:results}
\begin{figure}
        \centering
        \subfloat[]{\includegraphics[bb={0 0 0 0},trim={0 0 0 0},clip, width=0.45\columnwidth]{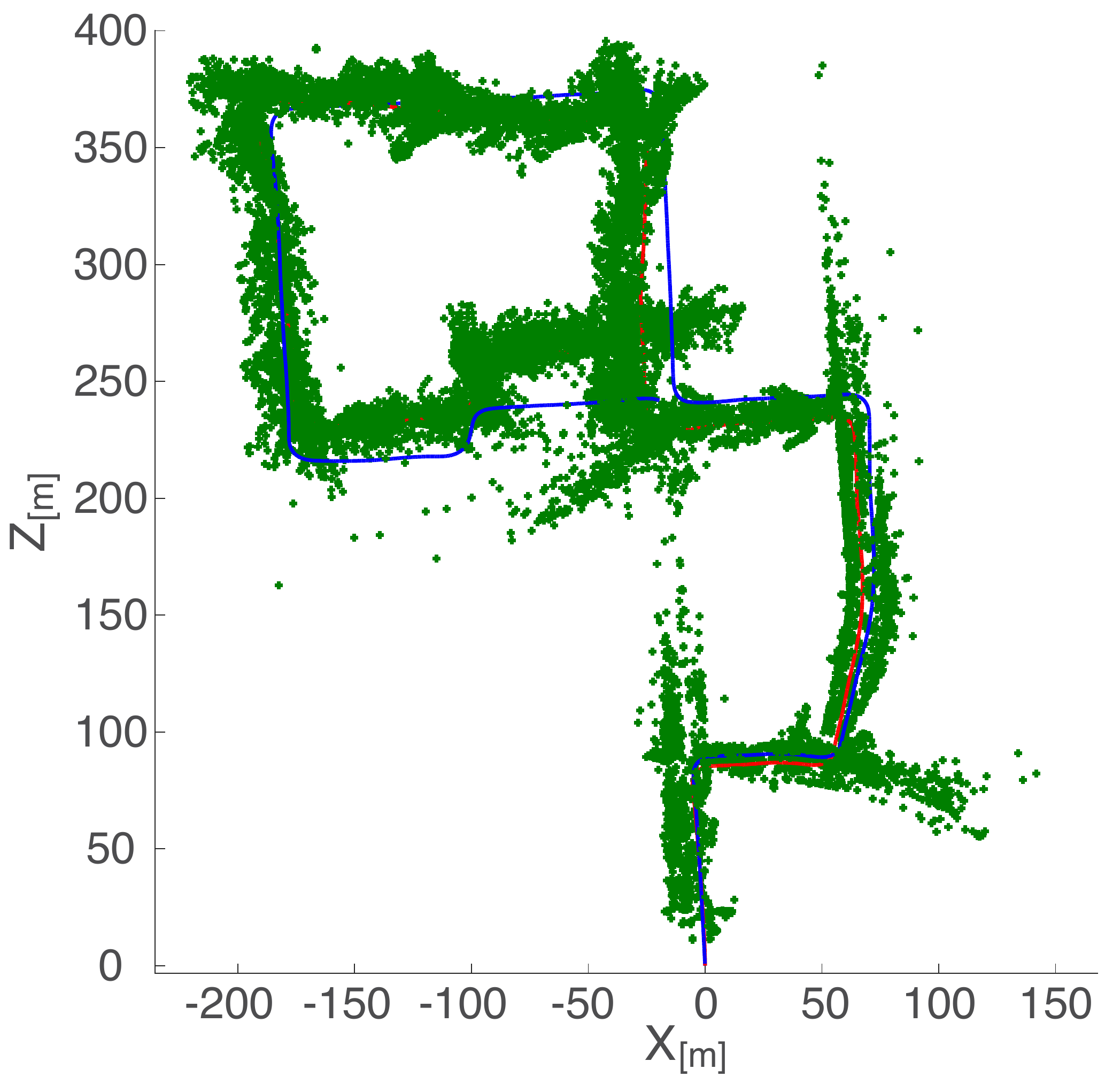}\label{fig:kitti:map}}
        \subfloat[]{\includegraphics[bb={0 0 0 0},trim={0 -18 0 0},clip, width=0.5\columnwidth]{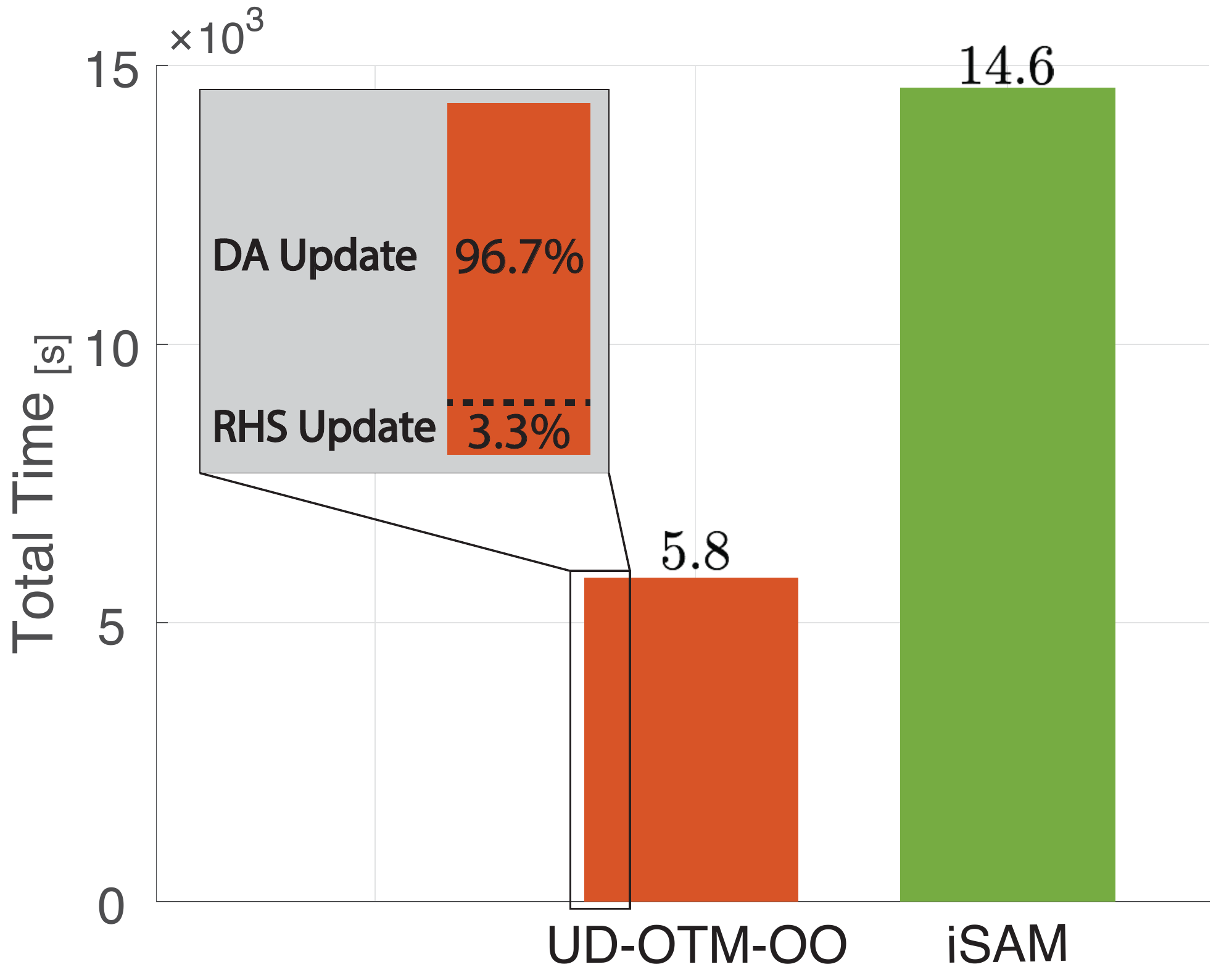}\label{fig:kitti:totTime}}
        \caption{Experiment layout and results: (a) The city of Karlsruhe, Germany, provided by the KITTI dataset. The robot ground truth is denoted in blue, the estimated trajectory denoted in dotted red line and the estimated landmark locations are denoted in green. (b) Total average running time of inference update for each method, when 100\% of the steps were with inconsistent DA.} 
        \label{fig:kitti:mapAndTTime}
\end{figure}
The robot travels $1400$ steps in the unknown streets of Karlsruhe Germany, while relying only on a monocular camera for localization and mapping and without encountering any substantial  loop closures. Differently than Section~\ref{ssubsec:ConstDA} and Section~\ref{ssubsec:inConstDA}, where the landmarks were omnidirectional and therefore can be spotted from every angle as long as they were in sensing range, when using real world data the angle from which we see a landmark would have crucial affect on data association.Figure~\ref{fig:kitti:map} presents the ground truth of the robot's trajectory in blue, the estimated robot's trajectory in dotted red and the estimated location of observed landmarks in green. 
Both \texttt{iSAM} and \texttt{UD-OTM-OO} produce the same estimation; therefore, the dotted red-line as well as the green marks represents both \texttt{iSAM} and \texttt{UD-OTM-OO} estimations.

\begin{figure}
        \centering
        \subfloat[]{\includegraphics[bb={0 0 0 0},trim={0 0 0 0},clip, width=0.5\columnwidth]{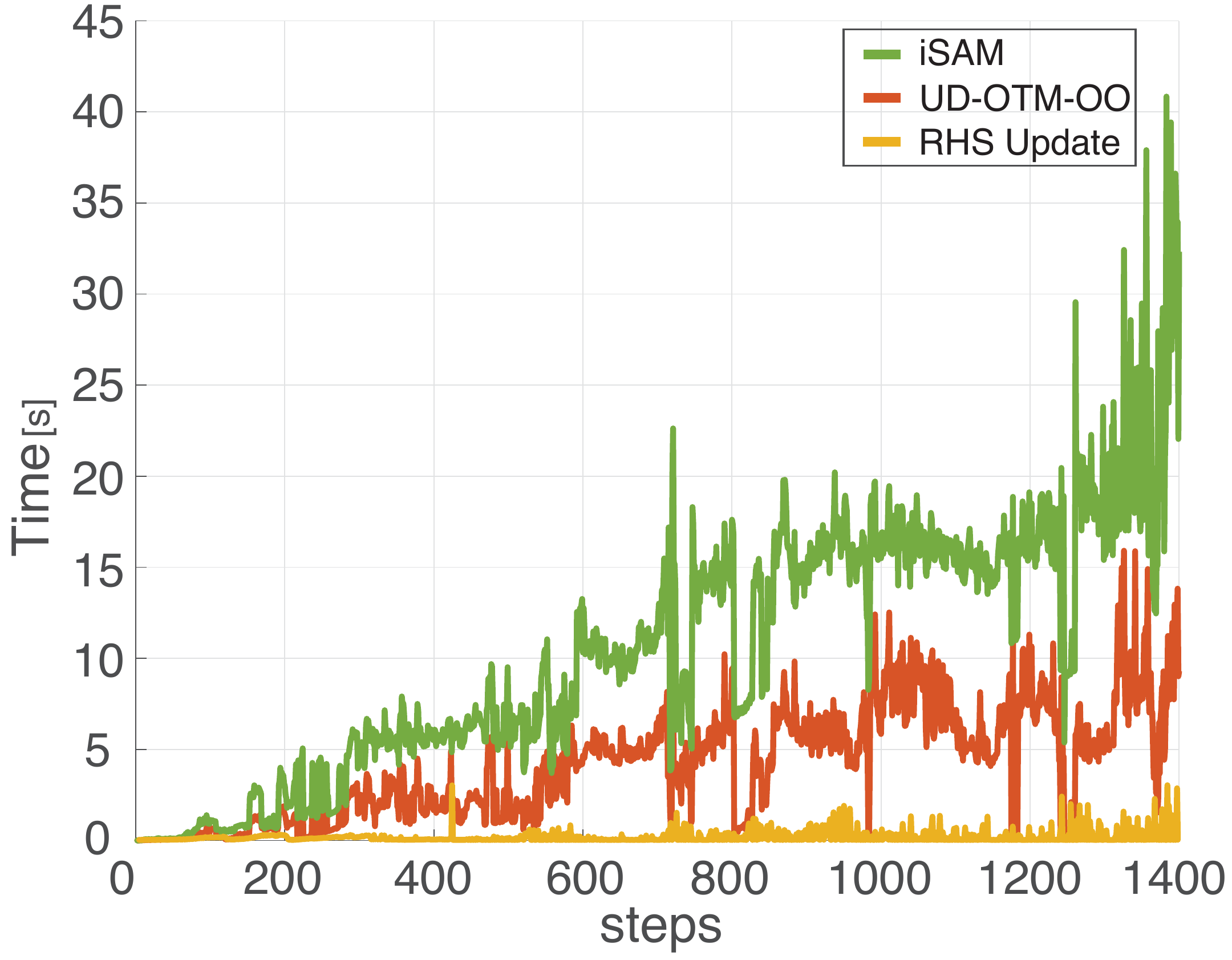}\label{fig:kitti:perstep}}
        \subfloat[]{\includegraphics[bb={0 0 0 0},trim={0 0 0 0},clip, width=0.5\columnwidth]{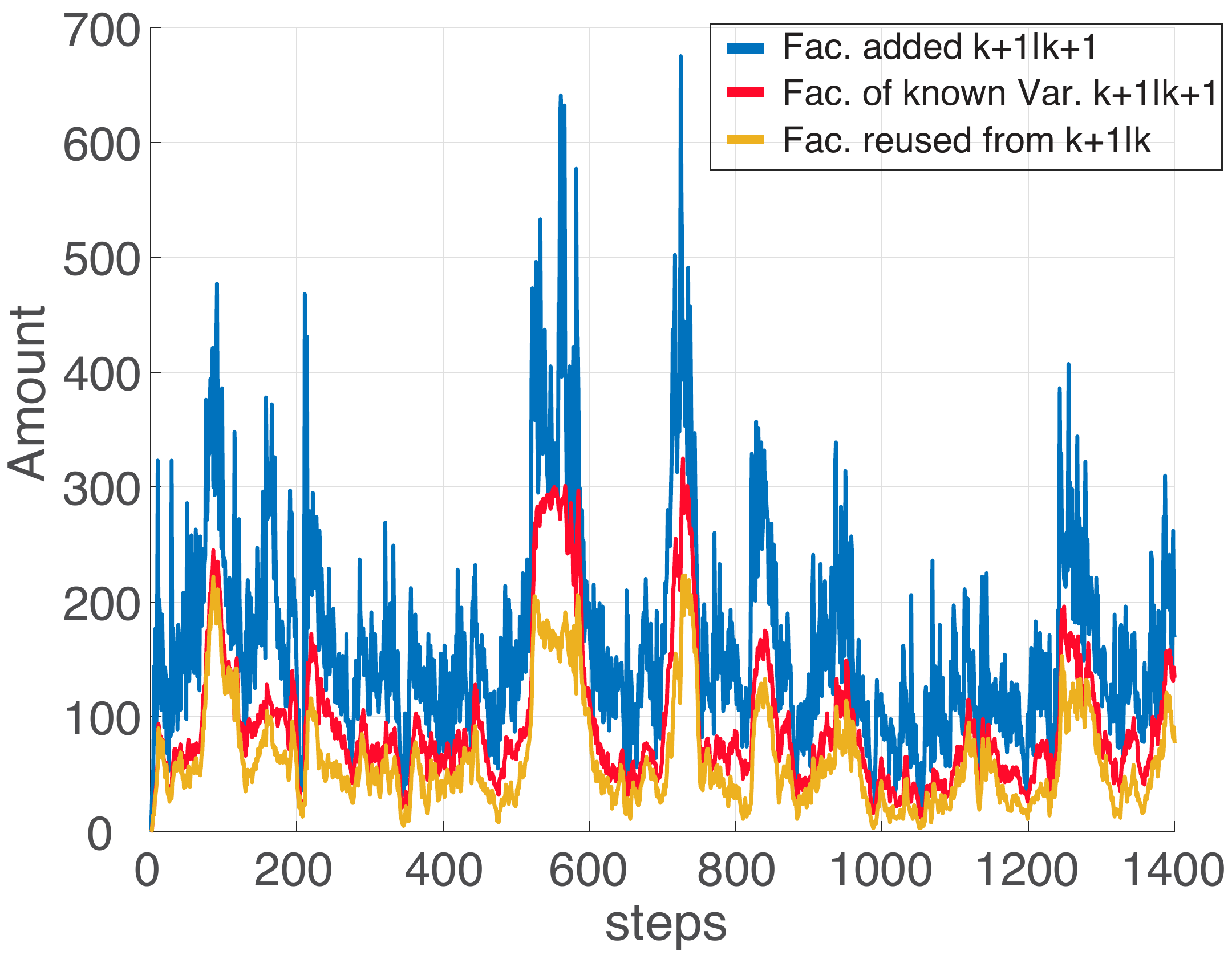}\label{fig:kitti:added_fac}}      
        \caption{Per-step analysis of computation time and added factors amount. (a) Inference update computation time per-step comparison between: \texttt{iSAM} - traditional Bayesian inference and \texttt{UD-OTM-OO} - inference update using belief from precursory planning. For reference the RHS update portion out of \texttt{UD-OTM-OO} is denoted in yellow. (b) Number of added factors per step. Number of all factors added in \texttt{iSAM} during inference at time $k+1$, denoted in blue. Number of factors added in \texttt{iSAM} during inference at time $k+1$ and relate to known variables, denoted in red. Number of factors that were originaly calculated during planning at time $k+1|k$ and were added by \texttt{UD-OTM-OO} in inference at time $k+1$, denoted in yellow.} 
        \label{fig:kitti:resaults}
\end{figure}

Figure~\ref{fig:kitti:totTime} presents the total computation time of inference update throughout the experiment, for both \texttt{iSAM}, and \texttt{UD-OTM-OO}. The importance of real-world data can be easily noticed by comparing Figures~\ref{fig:kitti:totTime} and \ref{fig:tot_time}. While the RHS update portion of \texttt{UD-OTM-OO} secured its advantage of two orders of magnitude over \texttt{iSAM}, it is not the case with \texttt{UD-OTM-OO} as a whole. Although for real-world data, \texttt{UD-OTM-OO} is still faster than \texttt{iSAM}, the difference has decreased from order of magnitude in Figure~\ref{fig:tot_time}, to less than half the computation time in Figure~\ref{fig:kitti:totTime}. Since the same machine has been used in both cases, the difference must originate from the data itself. As will be seen later in Figure~\ref{fig:kitti:added_fac}, the number of measurements per step is substantially higher when using the real-world data, as well as the occurrences of inconsistent DA. 
It is worth stressing that \texttt{iSAM} implementation for inference update is C++ based, while \texttt{UD-OTM-OO} implementation consists of a mixture of MATLAB based and C++ based implementation, so under the use of the same platform the computation time difference is expected to be higher.   

We continue by discussing the estimation difference, between \texttt{iSAM} and our method \texttt{UD-OTM-OO}. 
Although our method is algebraically equivalent to estimation via \texttt{iSAM}, for the reader's assurance we also provide estimation error comparison for both mean and covariance. Despite the algebraic equivalence, we expect to obtain small error values, related to numerical noise, which are different from absolute zero.
\begin{figure}
        \centering
        \subfloat[]{\includegraphics[bb={0 0 0 0},trim={0 0 0 0},clip, width=0.5\columnwidth]{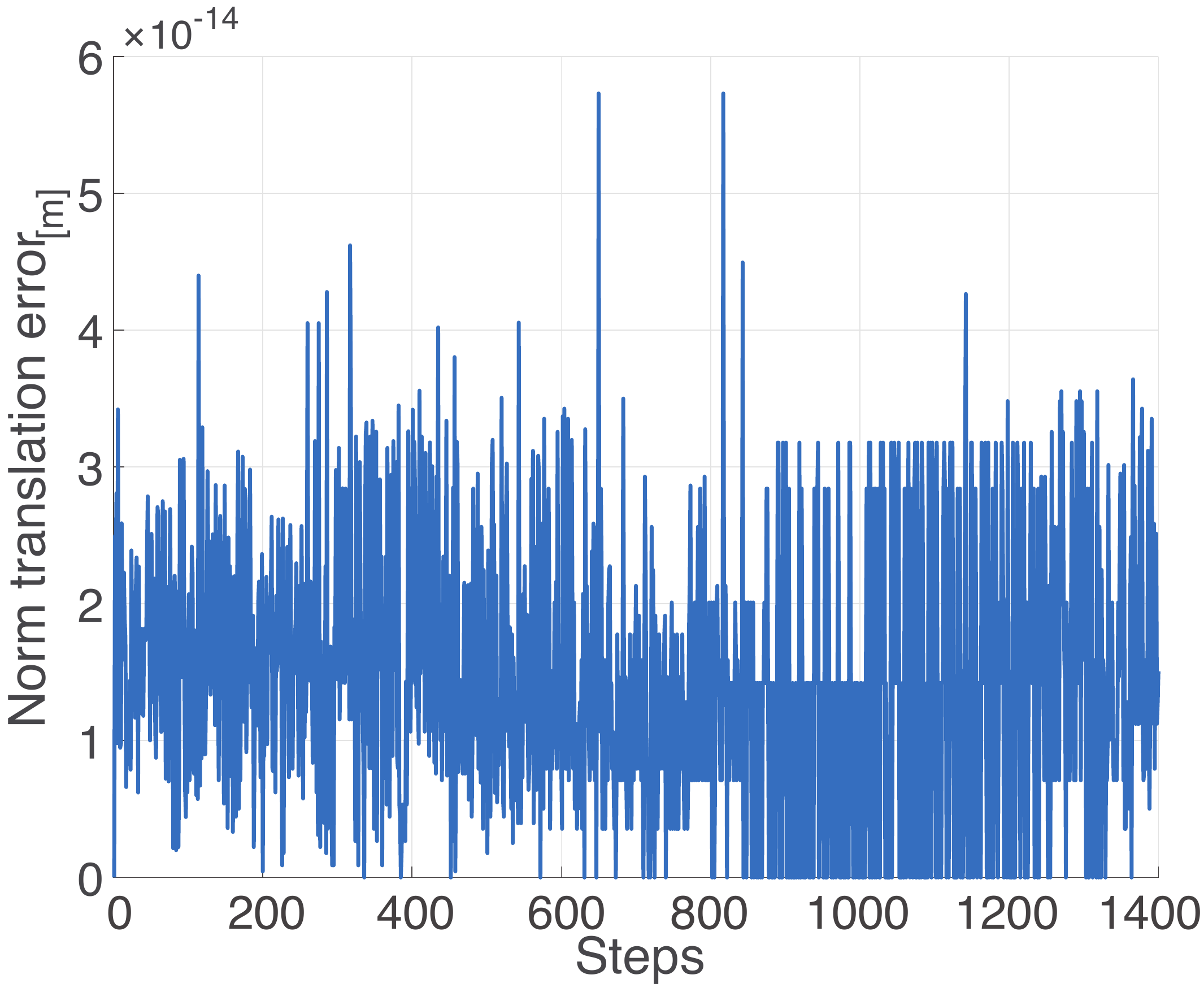}\label{fig:kitti:err:pos}}
        \subfloat[]{\includegraphics[bb={0 0 0 0},trim={0 0 0 0},clip, width=0.5\columnwidth]{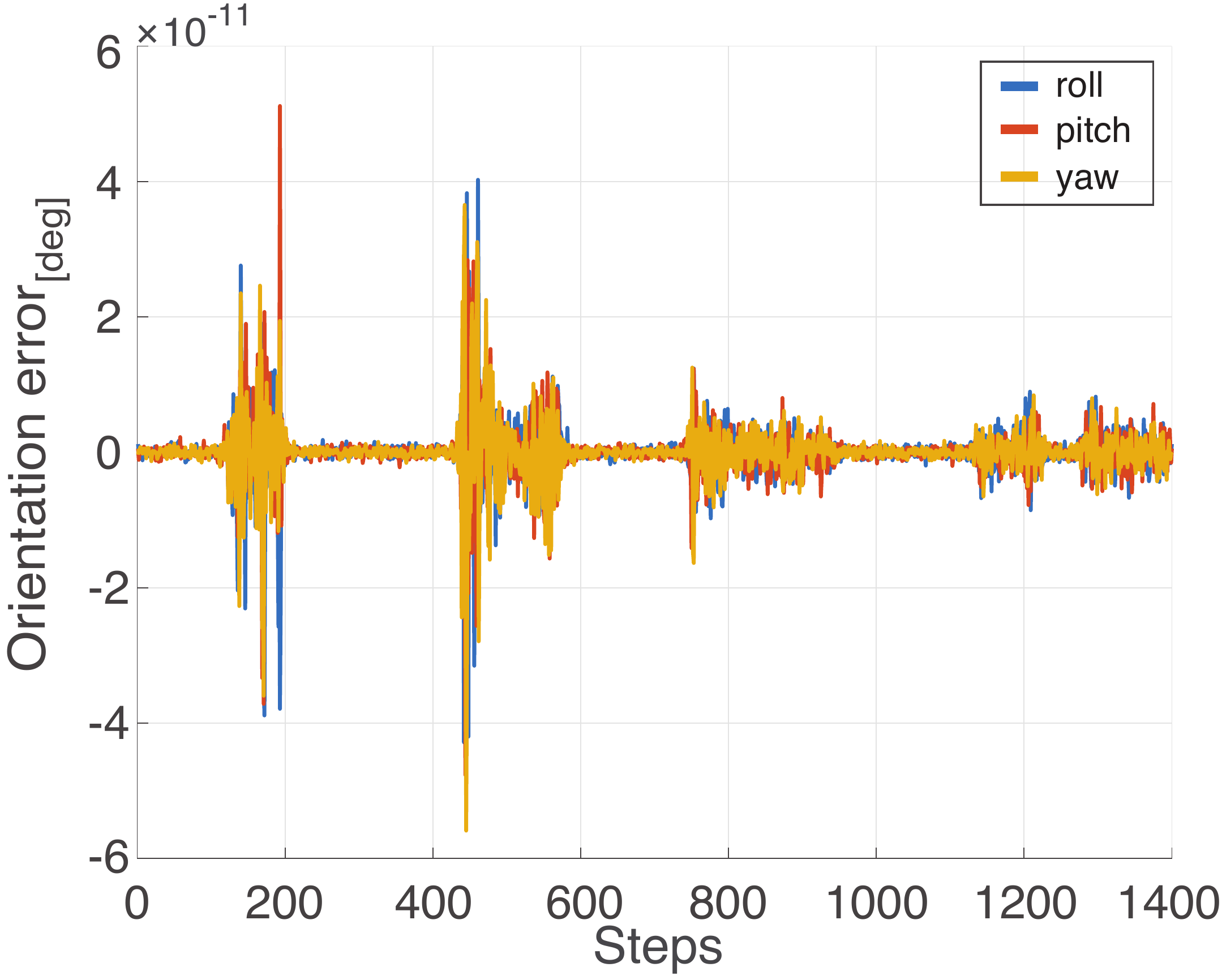}\label{fig:kitti:err:rot}}\\
        \subfloat[]{\includegraphics[bb={0 0 0 0},trim={0 0 0 0},clip, width=0.5\columnwidth]{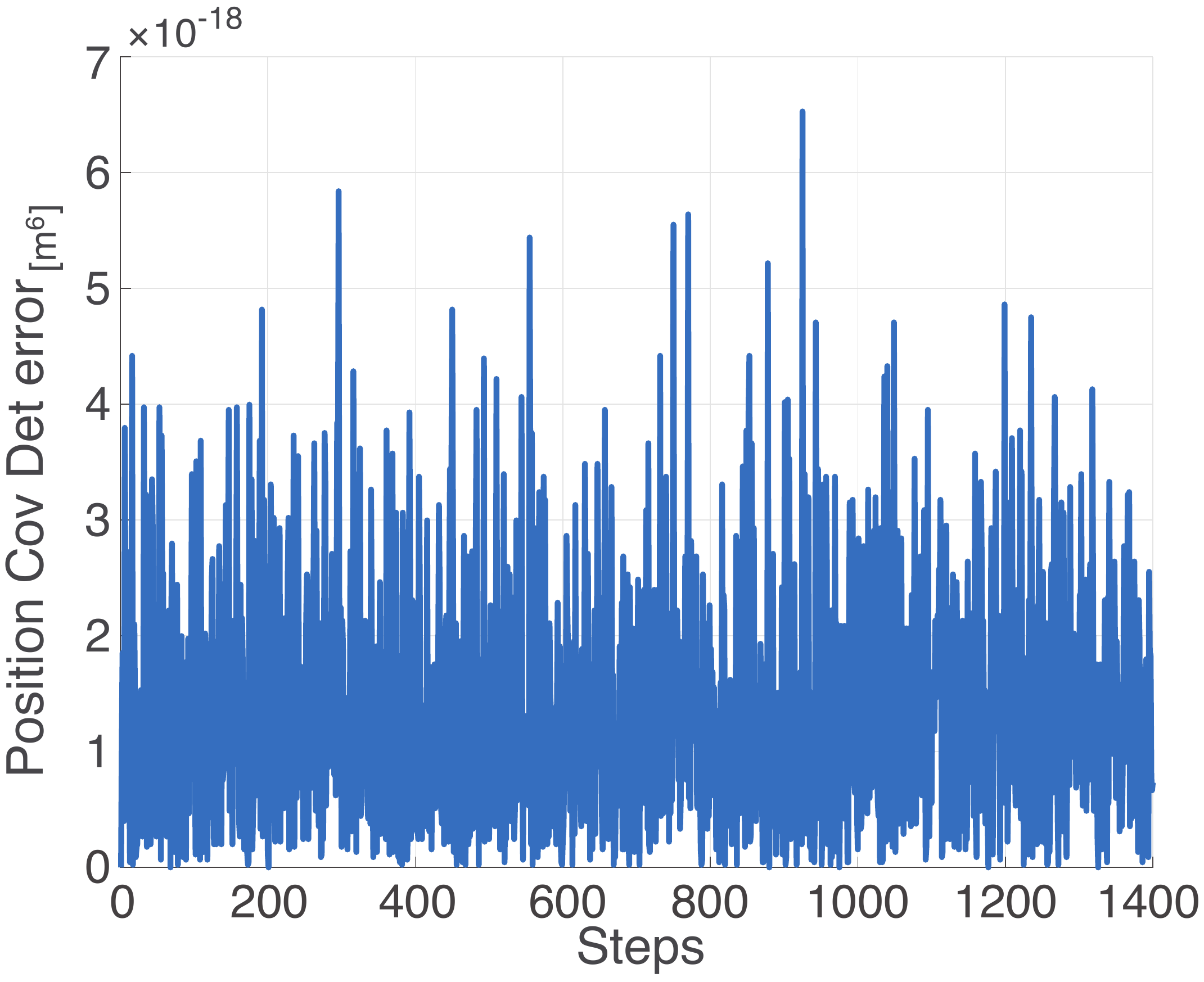}\label{fig:kitti:err:cov_pos}}
        \subfloat[]{\includegraphics[bb={0 0 0 0},trim={0 0 0 0},clip, width=0.5\columnwidth]{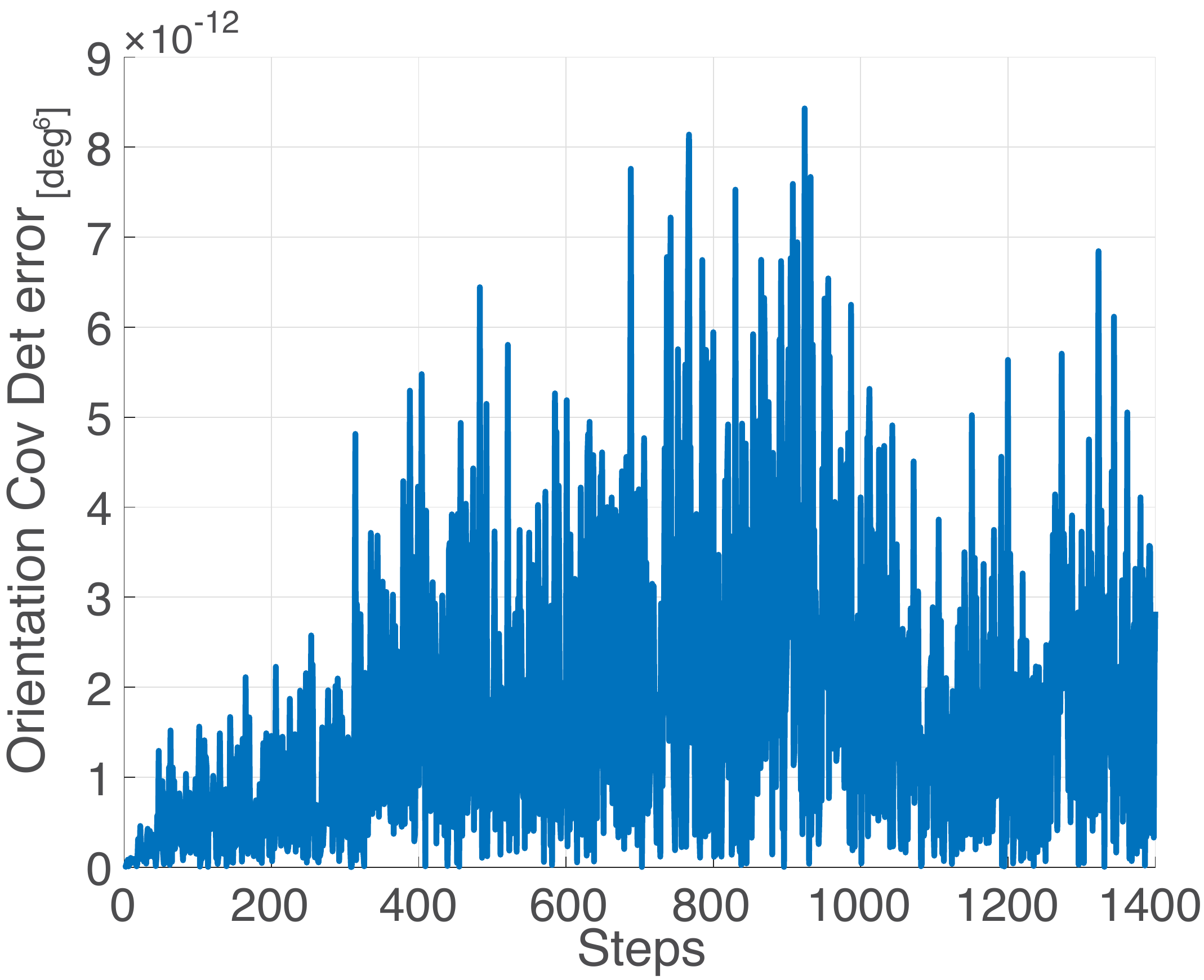}\label{fig:kitti:err:cov_rot}}\\
        \caption{Relative estimation error between \texttt{iSAM} and \texttt{UD-OTM-OO}, for KITTI dataset experiment (a) Relative translation error, calculated by taking the norm of the difference between the two translation vectors (b) Relative rotation error, calculated by taking the norm of the difference between the two orientation vectors, i.e. Euler angles (c) Relative position covariance error, calculated by taking the determinant of the difference between the two covariance matrices (d) Relative orientation covariance error, calculated by taking the determinant of the difference between the two covariance matrices.} 
        \label{fig:kitti:err}
\end{figure}
The estimation comparison results are presented in Figure~\ref{fig:kitti:err}: the translation mean in Figure~\ref{fig:kitti:err:pos}, the mean rotation of the robot in Figure~\ref{fig:kitti:err:rot} and the corresponding covariances in Figures~\ref{fig:kitti:err:cov_pos} and \ref{fig:kitti:err:cov_rot} accordingly.
The mean translation error is calculated by taking the norm of the difference between the two mean translation vectors. The mean rotation error is calculated by taking the norm of the difference between each of the mean body angles. The covariance error is calculated by taking the norm of the difference between the covariance determinants.
 As can be seen in Figure~\ref{fig:kitti:err}, the error has a noise like behavior, with values of $10^{-14}_{[m]}$ for translation mean, $10^{-11}_{[deg]}$ for mean rotation angles, $10^{-3}_{[m]}$ for translation covariance and $10^{-2}_{[deg]}$ for rotation angles covariance. For all practical purposes, these values points to a negligible accuracy difference between the two methods. 

Figure~\ref{fig:kitti:perstep} presents the per-step computation time for inference update of  \texttt{UD-OTM-OO} and \texttt{iSAM}, as well as the RHS update portion out of \texttt{UD-OTM-OO} for reference. 
The RHS Update timing of \texttt{UD-OTM-OO}, denoted by a yellow line, represents the per-step computation time of inference update through \texttt{RUB inference} for consistent DA, i.e. computation time for updating the RHS with the correct measurement values after the DA has been updated. \texttt{UD-OTM-OO} represents the per-step computation time of inference update through \texttt{RUB inference} for the entire process - DA update followed by RHS update. The difference in computational effort between the two, as seen in Figure~\ref{fig:kitti:perstep}, is equivalent to the computation time of the DA update, which represents the need to deal with inconsistent DA between belief from planning $b[X_{k+1|k}]$ and succeeding inference $b[X_{k+1|k+1}]$. The difference in computational effort between \texttt{UD-OTM-OO} and \texttt{iSAM} is attributed to the re-use of calculations made during the precursory planning. This calculation re-use manifests in salvaging factors that have already been considered during the precursory planning.

The reason for the considerable computational time differences between \texttt{UD-OTM-OO} and \texttt{iSAM} is better understood when comparing the factors involved in the computations of each method.

Figure~\ref{fig:kitti:added_fac} presents the sum of added factors per-step. In blue, the sum of all factors added at time $k+1|k+1$, as part of standard Bayesian inference update. In red, the portion of the aforementioned factors that relate to states which are already part of the belief $b[X_{k|k}]$.
In yellow, the amount of factors added in time $k+1$ as part of \texttt{UD-OTM-OO} and are shared by both beliefs, $b[X_{k+1|k}]$ and $b[X_{k+1|k+1}]$, i.e.~the amount of factors that were originally calculated in the precursory planning time, and were reused by \texttt{UD-OTM-OO}. It is worth stressing the noticeable difference between the number of measurements per step in Figure~\ref{fig:kitti:added_fac} when compared to Figure~\ref{fig:comp_fac}. The former is exceeding the latter by an order of magnitude.

The difference between the yellow and blue lines in Figure~\ref{fig:kitti:added_fac} represents the amount of factors "missing" from the belief $b[X_{k+1|k}]$ in order to match $b[X_{k+1|k+1}]$ (see Section~\ref{ssubsec:updateDA}), e.g. for step $725$ only $142$ have been reused (yellow line) while $675$ were eventually added (blue line), leaving $533$ new factors to be added during the DA update phase of \texttt{UD-OTM-OO}. This difference can be divided into factors containing only existing states and factors containing new states. Since the red line represents all factors of existing states, the difference between the red and blue lines represents all factors containing new states per time step, e.g. for step $725$, out of the $675$ factors added during inference (blue line), only $236$ are related to previously known states (red line). As mentioned earlier in Section~\ref{ssubsec:kitti:planning}, in this experiment the prediction of future factors does not involve new states, apart from the next future pose(s). For that reason, the amount of factors added during planning has an upper bound represented by the red line, e.g. for step $725$, the maximum number of factors that could have been utilized from precursory planning is $236$ (red line). Future work can consider a prediction mechanism for new states, such work would set the upper bound somewhere between the red and blue lines.   

The difference between the yellow and red lines, both related to factors of existing states, is attributed to the prediction accuracy of the planning stage. Since the factors represented by the red line are already part of the belief in planning time $k$, a perfect prediction mechanism would have added them all to the belief $b[X_{k+1|k}]$, e.g. for step $725$, while there are $236$ factors related to previously known states (red line), planning predicted only $142$ of them (yellow line). Since the prediction is inherently imperfect (see Section~\ref{subsec:bsp}), there would always be some difference between the two. Reducing the gap between the red and yellow lines is a function of the prediction mechanism, while closing the gap further up to the blue line is a function of predicting new variables during planning.

\begin{figure}
        \centering
        \includegraphics[bb={0 0 0 0},trim={0 0 0 0},clip, width=0.55\columnwidth]{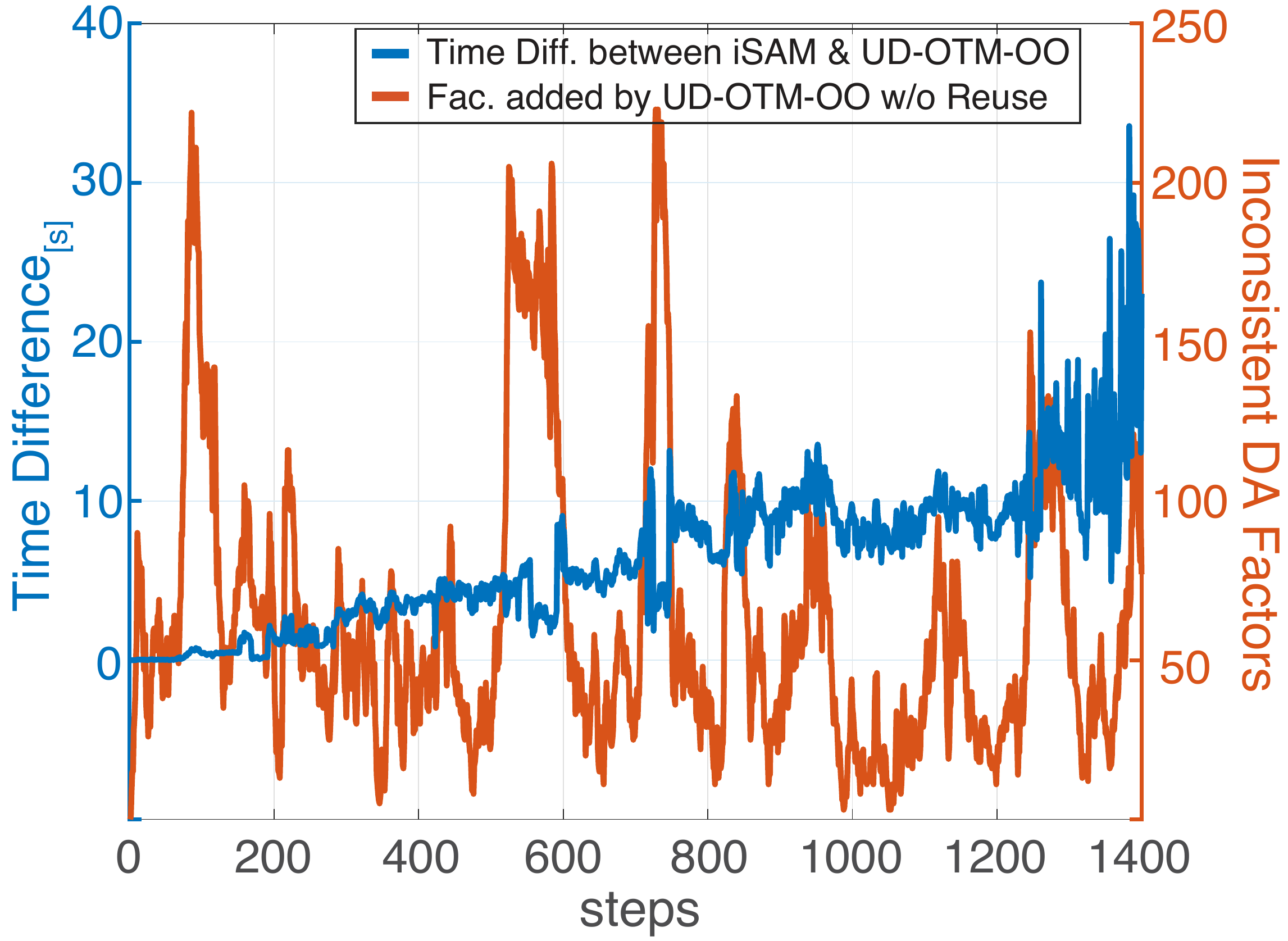}
        \caption{Inference update computation time analysis between \texttt{iSAM} and \texttt{UD-OTM-OO}. The left vertical blue axis, represents the inference update computation time difference between \texttt{iSAM} and \texttt{UD-OTM-OO}, where positive values suggest $t_{\texttt{iSAM}} > t_{\texttt{UD-OTM-OO}}$. The right vertical orange axis, represents the number of factors per step added by \texttt{UD-OTM-OO} that were \emph{not} reused from planning.} 
        \label{fig:kitti:timediff}
\end{figure}

After better understanding the meaning of Figure~\ref{fig:kitti:added_fac}, comparing the two graphs in Figure~\ref{fig:kitti:resaults}, reveals the connection between the added factors and the computation time, demonstrated by comparing steps $725$ and $803$ across the aforementioned. For time step $725$, we have $675$ new factors added in \texttt{iSAM} at inference, while only $142$ factors that have been reused by \texttt{UD-OTM-OO}, this difference resulted in inference update running time of $4.2_{[s]}$ to \texttt{UD-OTM-OO} and $6.1_{[s]}$ to \texttt{iSAM}. For time step $803$, we have $145$ new factors added in \texttt{iSAM} at inference, while only $33$ factors that have been reused by \texttt{UD-OTM-OO}, this difference resulted in inference update running time of $0.33_{[s]}$ to \texttt{UD-OTM-OO} and $6.9_{[s]}$ to \texttt{iSAM}. Although $6825$ new landmarks were added to the state vector between steps $725$ and $803$ (calculated by the cumulative difference between the blue and red lines between steps $725$ and $803$), the time difference between \texttt{UD-OTM-OO} and \texttt{iSAM} increased, while \texttt{UD-OTM-OO} running time dropped. This increase in relative running time, in-spite of the substantial growth of the state vector, can be attributed to the drop in the number of factors needed to be added by the DA update phase of \texttt{UD-OTM-OO}. 

As anticipated, the larger the gap between $b[X_{k+1|k}]$ and $b[X_{k+1|k+1}]$, i.e. more DA corrections to $b[X_{k+1|k}]$ are required in order to match $b[X_{k+1|k+1}]$, the smaller the computation time difference between \texttt{RUB inference} and standard Bayesian inference, as demonstrated in Figure~\ref{fig:kitti:timediff}. The left vertical axis (denoted in blue) presents the computation time difference between \texttt{iSAM} and \texttt{UD-OTM-OO} such that positive values suggest $t_{\texttt{iSAM}} > t_{\texttt{UD-OTM-OO}}$. From this blue graph we notice that the time difference between \texttt{iSAM} and \texttt{UD-OTM-OO} is strictly positive and ascending up to fluctuations. While some of these fluctuations can be attributed to machine noise of the measurement process, we provide some explanation for the large time difference drops, i.e. the steps in which the computation difference between \texttt{iSAM} and \texttt{UD-OTM-OO} diminished. The number of factors added by \texttt{UD-OTM-OO} and were not reused from precursory planning are denoted by the orange line in Figure~\ref{fig:kitti:timediff}. We can see correlation between large spikes in the number of factors added during the DA update phase of \texttt{UD-OTM-OO} (orange line), and the drops in the time difference between \texttt{iSAM} and \texttt{UD-OTM-OO} computation time difference (blue line), e.g. steps $554-591$ and $720-746$.

In contrast to previous experiments over synthetic data, we can better see here some dependency over the size of the belief in the \texttt{UD-OTM-OO} method. This dependency seems to be in correlation with that of \texttt{iSAM2} although less intense, as can be seen by comparing the two methods in Figure~\ref{fig:kitti:perstep}. As in Figure~\ref{fig:analysis}, we can attribute this correlation to the number of re-eliminations performed per step, which are a function of the newly added factors for \texttt{iSAM} and the DA update in \texttt{UD-OTM-OO} (see Section~\ref{subsec:UpdateDA_chapter}). As mentioned earlier, in each step \texttt{UD-OTM-OO} encounters inconsistent DA, judging by the difference between the blue and yellow lines in Figure~\ref{fig:kitti:added_fac}, each step \texttt{UD-OTM-OO} deals with at least 100 factors that were not reused from planning.
Since \texttt{UD-OTM-OO} makes use of \texttt{iSAM2} methodologies in order to update inconsistent DA, as does \texttt{iSAM2} to update inference, they share similar computational sensitivities, which manifest in similar computation time trends. This similarity sensitivity is attributed in our opinion to the elimination process required in order to introduce new factors into the belief. Future work for reducing eliminations by anticipating required ordering, would break this dependency and provide additional improvement in computation time as well as in reducing the sensitivity to state dimensionality.

\section{Broader Perspective}
\label{sec:motivation}
In this section we briefly discuss the motivation for \RUB and provide some broader perspective to possible future usage. 
As mentioned earlier, the \RUB paradigm deals with inference update within a plan-act-infer framework. 
By re-using calculations from the precursory planning session, it offers reduced computation time without affecting estimation accuracy.  

Decision making under uncertainty in high dimensional state spaces is computationally intractable, and as such the majority of the plan-act-infer computation time can be ascribed to it. 
For example, let us consider BSP under the simplified Maximum Likelihood (ML) assumption, with a planning horizon of three lookahead steps and three candidate actions per step. The first level of the belief tree would consist of three future beliefs, one for each candidate action, each of which is propagated with each of the three candidate actions, resulting in nine future beliefs in the second level of the belief tree, and again for the last lookahead step with $27$ future beliefs in the third and last level of the belief tree. This would result in total of $39$ future beliefs that constitute the belief tree, i.e. $39$ belief updates, whereas only a single belief update is required during inference update.
In this toy example the computational load of inference update constitutes therefore only $2.5\%$ of the plan-act-infer framework (assuming all belief updates have the same computational load). 
So why should we bother with the efficiency of the inference update process in the first place? 

The answer to this question is twofold, the first part deals with \RUB paradigm as a stand-alone approach for inference update, and the second with its possible contribution to for future research. 

Although we present \RUB as part of a plan-act-infer system, it can also be used in the passive case, i.e. not as part of a plan-act-infer ssytem. Imagine a set of candidate beliefs, calculated offline and stored away for future usage. When in need to perform belief update, we can search this set of candidate beliefs for the belief closest to last posterior as well as to the newly received information. Once we locate this closest belief, we use \RUB to update it to match current information thus saving computational load without affecting accuracy. 
The reason in this work we consider \RUB as part of a plan-act-infer system, lies within the problem of locating the closest candidate belief. By using beliefs from precursory planning as candidates we have a small set of candidates to look through, moreover we can ensure that in the worst case scenario (i.e. all predictions from precursory planning are wrong) we would match Bayesian inference performance, thus averting from the complicated problem of searching within belief space. 
For the general case of having a set of previously calculated beliefs, used as candidates for re-use under \RUB, one would need to deal with few issues some of which are: how to store the beliefs to facilitate an efficient search, how to efficiently search the set of candidate beliefs, what high-dimensional belief-distance to use, how to interpret belief-distance into computational load i.e. what will be considered as close enough.
It goes without saying that the computational load of locating the closest belief should be small enough for \RUB to still have a computational advantage.   
 
Secondly, the paradigm shift suggested by \RUB provides a pathway to new and exciting research directions. For example, \RUB is a key building block in the new concept of Joint Inference and belief space Planning (\JIP), first presented in \cite{Farhi17icra} and later in \cite{Farhi19icra_ws}, which 
strives to create a unified framework that deals with both inference and BSP under the same governing system, thus allowing to maximize the calculation re-use potential available in both inference and planning.

\begin{figure}
	\centering
        \includegraphics[bb={0 0 0 0},trim={0 0 0 0},clip, width=0.6\textwidth]{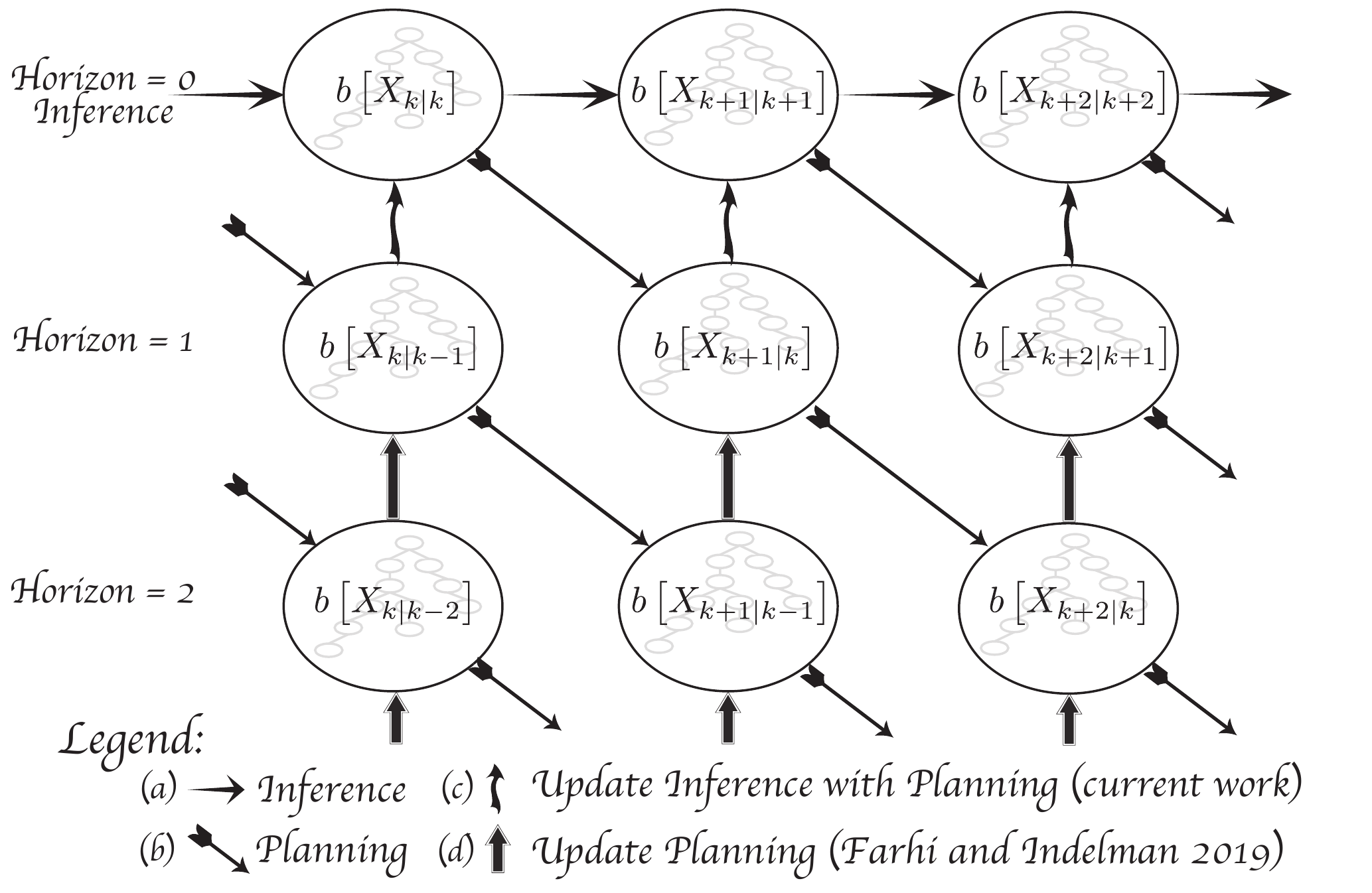}
        \caption{Visualization of \JIP, a novel approach to address both inference and belief space planning under a single process. Here $b[X_{k+1|k}]$ stands for the belief of the joint state in time instance $k+1$ while current time is $k$ and each row stands for a different planning horizon. The relations between different beliefs in the graph are denoted by different arrows. (a) Inference; (b) Planning step;  (c) Updating Inference with precursory planning (\emph{this paper}); (d) Update planning with precursory planning.}
        \label{fig:jip}
\end{figure}

Figure~\ref{fig:jip} provides a graphical illustration of \JIP. The joint inference and belief space planning approach incorporates both inference and decision making stages into a single process. Each node in the graph represents a belief, i.e. $b[X_{k+1|k}]$ denotes the joint belief of state $X$ at a future time instant $k+1$ given that the current time is $k$. The right facing arrows (a)  denote inference at sequential time instances i.e.~standard Bayesian inference. The diagonal arrows  (b)  represent optimal controls and future measurements that lead up to the appropriate beliefs. 
The upward facing arrow (c) represents our \emph{current work} on \RUB, as it denotes inference update using precursory planning session. The upward facing arrow (d) represents our continued work named Incremental eXpectation BSP (\texttt{iX-BSP}) \citep{Farhi19icra}, as it denotes updating a future belief using some previously calculated planning session.  
For the reader's convenience this \JIP illustration is presented in 2D, but in-fact this 2D pattern is repeating itself in the 3D space to represents all possible future beliefs along with all possible candidate actions and future measurements, simply visualize rotating Figure~\ref{fig:jip} around the top row representing inference.

While \RUB provides efficient inference update, denoted by (c) in Figure~\ref{fig:jip}, it is also relied upon to facilitate calculation re-use between different planning sessions, denoted by (d) in Figure~\ref{fig:jip}.  
As part of BSP, it is required to create a belief tree, as deep as the planning horizon, where each node in the said tree represents a future belief with specific candidate actions and future measurements. In \cite{Farhi19icra} we make use of \RUB paradigm in order to incrementally update the belief tree as part of planning, thus saving valuable computation time from the planning stage. 

Consequently, apart from being a more computationally efficient approach for inference update, the paradigm shift suggested by \RUB provides a basis for exciting new research, i.e.~\RUB is the building block that facilitates  the \JIP concept as a whole.

\vspace{-5pt}
\section{Conclusions}
\label{sec:conclusions}
Conventional Bayesian inference updates the belief from a previous time step with new incoming information. In this work we introduced an alternative paradigm, utilizing the similarities between inference and planning to efficiently update inference using information from precursory planning phase. 
Given a future belief from precursory planning and newly acquired data, we appropriately update the former with the latter while taking into consideration data association inconsistencies which might occur.
The resulting approach, \texttt{RUB inference}, saves valuable computation time in inference without affecting the estimation accuracy.

We evaluated our approach in simulation and using real-world data from the KITTI dataset, considering active SLAM as application, and compared it against \texttt{iSAM2}, a state-of-the-art incremental Bayesian inference approach. Results from real-world evaluation indicate that our approach is more efficient computationally by at least a factor of two compared to \texttt{iSAM2}, without affecting the solution accuracy. The improvement magnitude is in direct correlation with the quality of the prediction mechanism being used in planning, meaning a better prediction mechanism would increase the approach's efficiency. 
A particular appealing aspect of our method, that we demonstrated using synthetic data, is that loop closures computational burden during inference is elevated, thanks to the utilization of similar calculations already made during precursory planning. When loop closures were correctly predicted during the planning phase, our method utilized these calculations instead of re-calculating them in inference, resulting in reducing computation time by a factor of two orders of magnitude in the shown results.

This paper suggests a novel general concept for leveraging calculations from the decision making stage for efficient inference update, thus enabling to reduce inference computation time without affecting accuracy. 
Based on this concept, under the assumption of high-dimensional Gaussian beliefs, we developed approaches based on the square root information matrix, to efficiently update inference. We strongly believe this novel concept is applicable for more general distributions in any autonomous system involving both inference and decision making under uncertainty. 
Based on our findings, we strongly believe this paradigm shift opens new research directions and can be further extended in various ways, e.g. our ongoing work on \texttt{ix-BSP} - incremental expectation BSP \citep{Farhi19icra} leverages \texttt{RUB Inference} to reuse calculations across different planning sessions.


\bibliography{paper.bib}

\section*{Appendix A: Derivation of Equation (\ref{eq:Ax_b})}
In this appendix we complete the derivation of Eq.~(\ref{eq:Ax_b}) from Eq.~(\ref{eq:mahal_factors}). Let us consider the NLS presented in Eq.~(\ref{eq:mahal_factors})
\begin{equation*}
X_{k|k}^{\star} = \argmin_{X_{k}} \Vert x_{0}-x_{0}^{\star}\Vert_{\Sigma_{0}}^{2} +
 \sum_{i=1}^{k}\!\! \left[ \Vert x_{i}\!-\!f(x_{i-1},u_{i-1|k})\Vert_{\Sigma_w}^{2}\!\!+\!\!\!\!\! \sum_{j \in \mathcal{M}_i|k}\!\!\!\!\! \Vert z_{i|k}^{j}\!\!-\!h(x_{i},l_{j}) \Vert_{\Sigma_v}^{2} \right]\! .
\end{equation*}
In general, the motion model $f(\cdot)$ and the measurement model $h(\cdot)$ are non-linear functions. A standard way to solve this problem is the Gauss-Newton method, where a single iteration involves linearizing about the last known estimate, calculating the delta around this linearization point, and updating the latter with the former. This process should be repeated until convergence. 
 
We start by linearizing the terms in (\ref{eq:mahal_factors}) using first order Taylor approximation around the best estimate we have for the joined state $\bar{X}_{k|k-1}$ which is the state estimate for time $k$ before including measurements, i.e. $X^*_{k|k-1}$. 

The prior term yields,  
\begin{equation}\label{eq:nls:prior}
x_{0}-x_{0}^{\star} = \bar{x}_{0} + \Delta x_0 - x_{0}^{\star} = \Delta x_0.
\end{equation} 
The motion model term yields,
\begin{equation}\label{eq:nls:motion}
x_{i}-f(x_{i-1},u_{i-1|k}) = \bar{x}_{i} - f(\bar{x}_{i-1},u_{i-1|k}) - \Sigma_w^{-\frac{1}{2}}\mathcal{F}_{i} \begin{bmatrix} \Delta x_{i-1} \\ \Delta x_{i}\end{bmatrix}
\end{equation} 
where $\Sigma_w^{-\frac{1}{2}}\mathcal{F}_{i}$ represents the Jacobian matrix of the motion model at time $i$, around  the linearization point $\bar{x}_{i-1:i}$.
The measurement model term yields,
\begin{equation}\label{eq:nls:measurement}
z_{i|k}^{j} - h(x_{i},l_{j})  = z_{i|k}^{j} - h(\bar{x}_{i},\bar{l}_{j}) - \Sigma_v^{-\frac{1}{2}}\mathcal{H}_{i,j}
\begin{bmatrix}
	\Delta x_i \\
	\Delta l_j
\end{bmatrix}
\end{equation} 
where $\Sigma_v^{-\frac{1}{2}}\mathcal{H}_{i,j}$ represents the measurement model jacobian matrix at time $i$ around the linearization point $\left[ \bar{x}_i, \bar{l}_j \right]^T$.

In order to re-write (\ref{eq:mahal_factors}) into the common form of Least Squares $Ax=b$, we introduce Eqs.~(\ref{eq:nls:prior}-\ref{eq:nls:measurement}) back to (\ref{eq:mahal_factors}), 
\begin{equation*}
\Delta X_{k|k}^{\star} = \argmin_{\Delta X_{k}} \Vert \Sigma_0^{-\frac{1}{2}} \Delta x_{0} \Vert^{2} +
 \sum_{i=1}^{k}\!\! \left[ \Vert \Sigma_w^{-\frac{1}{2}}\Delta x_{i} - \mathcal{F}_{i} \Delta x_{i-1} - \breve{b}_{i}^{\mathcal{F}}\Vert^{2}\!\!+\!\!\!\!\! \sum_{j \in \mathcal{M}_i|k}\!\!\!\!\! \Vert \mathcal{H}_{i,j}
\begin{bmatrix}
	\Delta x_i \\
	\Delta l_j
\end{bmatrix}
 - \breve{b}_{i}^{\mathcal{H}} \Vert^{2} \right]\! ,
\end{equation*}
where the RHS terms $\breve{b}_{i}^{\mathcal{F}}$ and $\breve{b}_{i}^{\mathcal{H}}$ are given by
\begin{equation*}
	\breve{b}_{i}^{\mathcal{F}} = \Sigma_w^{-\frac{1}{2}} \left( f(\bar{x}_{i-1},u_{i-1|k}) - \bar{x}_{i} \right) \qquad , \qquad
	\breve{b}_{i}^{\mathcal{H}} = \Sigma_v^{-\frac{1}{2}}  \left( z_{i|k}^{j} - h(\bar{x}_{i},\bar{l}_j) \right).
\end{equation*}

We now make use of the fact that the minimum sum of quadratic expressions is the minimum of each quadratic expression individually and is equal to zero. Thus enabling us to stack up all equations to form, 
\begin{equation*}
\Delta X_{k|k}^{\star} =\argmin_{\Delta X_{k}}\Vert A_{k|k} \Delta X_{k}-b_{k|k}\Vert^{2} ,
\end{equation*}
where the Jacobian matrix and the RHS are given by,

\begin{equation*}
A_{k|k} =
\begin{bmatrix}
	\Sigma_{0}^{-\frac{1}{2}}\\
	\mathcal{F}_{1:k|k}\\
	\mathcal{H}_{1:k|k}
\end{bmatrix} 
\quad , \quad b_{k|k}= 
 \begin{bmatrix}
 	0\\
 	\breve{b}_{1:k|k}^{\mathcal{F}}\\
 	\breve{b}_{1:k|k}^{\mathcal{H}}
 \end{bmatrix}.
\end{equation*}

\section*{Appendix B: Inference as a Graphical Model}

The inference problem can be naturally represented and efficiently solved using graphical models such as factor graph (FG) \citep{Kschischang01it} and Bayes tree (BT) \citep{Kaess10tr}. Since FG and BT graphical models pose key components in the suggested paradigm, the theoretical foundation is supplied next.
We use Figure~\ref{fig:FG-BT} as illustration to belief representation in graphical models. Figures \ref{fig:FG_k1_k} and \ref{fig:FG_k1_k1} are FG representations for the beliefs $b(X_{k+1|k})$ and $b(X_{k+1|k+1})$, respectively. BT representation of the belief is obtained through an elimination process, Figure~\ref{fig:BT_k1_k} presents the BT of $b[X_{k+1|k}]$ for the elimination order $x_0 \! \cdots l_i \!\! \rightarrow \!\! x_{k-1} \!\!\! \rightarrow \!\! x_k \!\! \rightarrow \!\! l_j \!\!\! \rightarrow \!\! x_{k+1}$, while Figure~\ref{fig:BT_k1_k1} presents the BT of $b[X_{k+1|k+1}]$ for the elimination order $x_0 \! \cdots l_i \!\! \rightarrow \!\! x_{k-1} \!\!\! \rightarrow \!\! x_k \!\! \rightarrow \!\! l_j \!\!\! \rightarrow \!\! l_r \!\!\! \rightarrow \!\! x_{k+1}$.

A FG is a bipartite graph with two node types, factor nodes $\{f_{i}\}$ and variable nodes $\{\theta_{j}\} \in \Theta$. All nodes are connected through edges $\{e_{ij}\}$, which are always between factor nodes to variable nodes.	A factor graph defines the factorization of a certain function $g ( \Theta )$ as
\begin{equation} \label{eq:FG_def}
	g(\Theta) = \prod_{i} f_{i} ( \Theta_{i}),
\end{equation}
where $ \Theta_{i}$ is the set of variables $\{ \theta_j \}$ connected to the factor $f_{i}$ through the set of edges $\{ e_{ij} \}$. After substituting $\Theta$ with our joint state $X$ and the factors $\{ f_{i} \}$ with the conditional probabilities from Eq.~(\ref{eq:factors}) we receive the definition of the belief $b(X_{t|k})$ in a FG representation.

\begin{figure}
        \centering
        \includegraphics[bb={0 0 0 0},trim={40 510 100 10},clip, width=0.5\columnwidth]{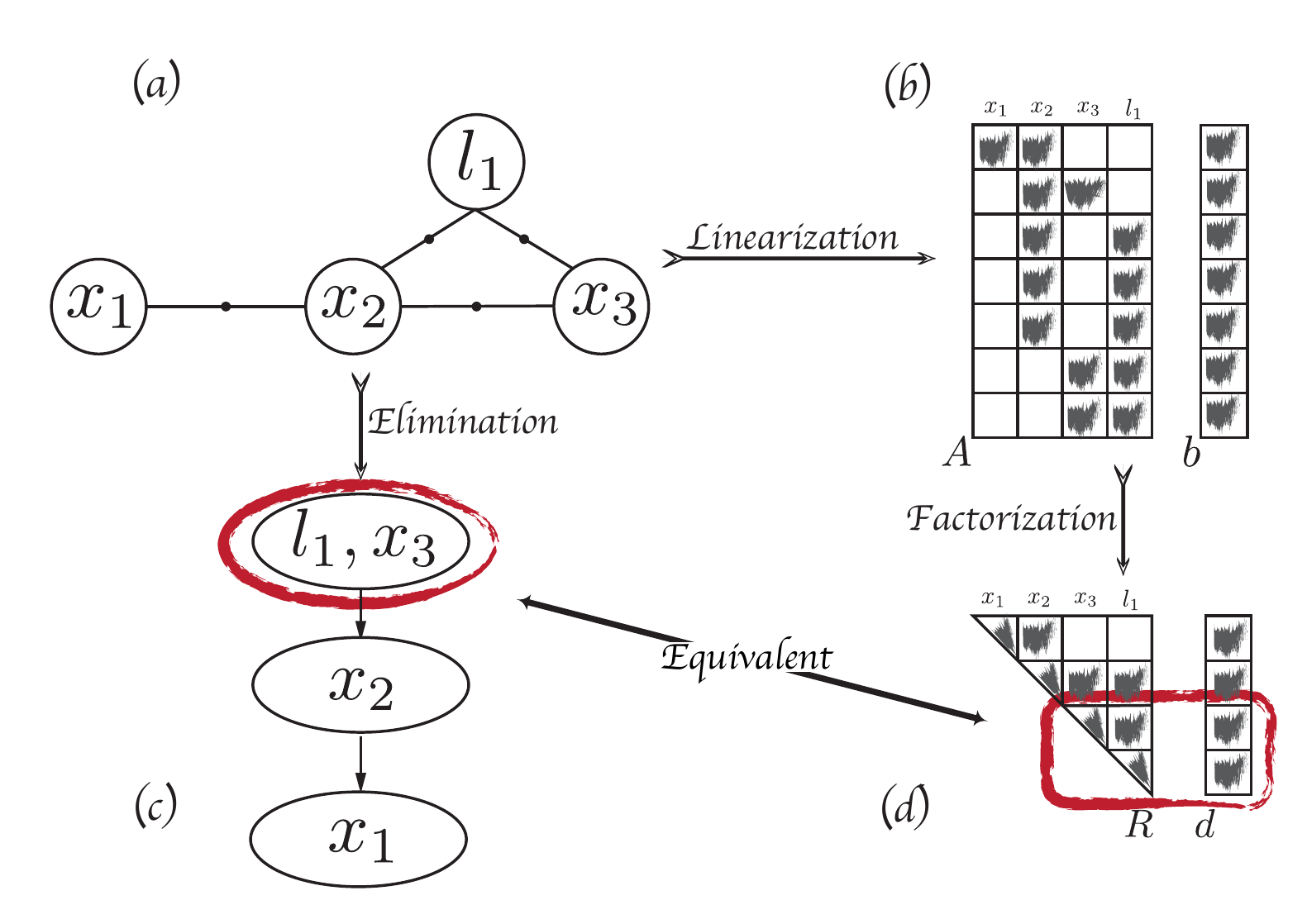}
        \caption{The relations between different problem representations. (a) Factor graph (b) Jacobian matrix $A$ with RHS vector $b$ (c) Bayes Tree (d) Factorized Jacobian matrix $R$ with equivalent RHS vector $d$.}
        \label{fig:FG_BT_Ab_Rd}
\end{figure}

Through bipartite elimination game, a FG can be converted into a BN, this elimination is required for solving the Inference problem (as shown in \cite{Kaess12ijrr}). After eliminating all variables the BN pdf can be defined by a product of conditional probabilities,
\begin{equation} \label{eq:BN_def}
	P(\Theta) = \prod_{j} P(\Theta_{j} | S_{j}),
\end{equation}
where $S_{j}$ is addressed as the \emph{separator} of $\Theta_{j}$, i.e. the set of variables that are directly connected to $\Theta_{j}$.
In order to ease optimization and marginalization, a BT can be used \citep{Kaess12ijrr}. By converting the BN to a directed tree, where the nodes represent \emph{cliques} $\{ C_{r} \}$, we receive a directed graphical model that encodes a factored pdf. Bayes Tree is defined using a conditional density per each node.
\begin{equation}\label{eq:BT_def}
		P(\Theta) = \prod_{r} P(F_{r} | S_{r}),
\end{equation}
where $S_{r}$ is the separator, defined by the intersection $C_{r} \bigcap \Pi_{r}$ of the clique $C_{r}$ and the parent clique $\Pi_{r}$. The complement to the variables in the clique $C_{r}$ is denoted as $F_{r}$, the \emph{frontal variables}. Each clique is therefor written in the form $C_{r}=F_{r}:S_{r}$. \newline
The correspondence between matrix and graphical representation is conveniently demonstrated in Figure~\ref{fig:FG_BT_Ab_Rd}. The first rows of $R$ are equivalent to the deepest cliques in the BT, when the last rows of $R$ equivalent to the root of the tree. The elimination order that created the BT is identical to the ordering of $R$ state vector, and fill-ins in $R$ equivalent to the connectivity of the corresponding BT.

\section*{Appendix C: Non-Zeros in Q matrix}
In this appendix we discuss the number of non zeros in the rotation matrix $Q_{k+1|k+1}$, in order to do so we first cover the creation of $Q_{k+1|k+1}$, and later get to an expression for the number of non zeros and analyze it to gain better understanding over the governing parameters.
 
The rotation matrix $Q_{k+1|k+1}$ is created as part of the factorization of the Jacobian, designed to rotate the Jacobian into a square upper triangular form (e.g. Eqs.~(\ref{eq:QR_A}) and (\ref{eq:QR_A_bsp})). As such, we can deduce an expression for the number of non zeros in $Q_{k+1|k+1}$ as a function of the state size and the size of added factors, but first let us review how $Q_{k+1|k+1}$ is being created. 
\begin{figure}
        \centering
        \subfloat[]{\includegraphics[bb={0 0 0 0},trim={0 0 0 0},clip, width=0.5\columnwidth]{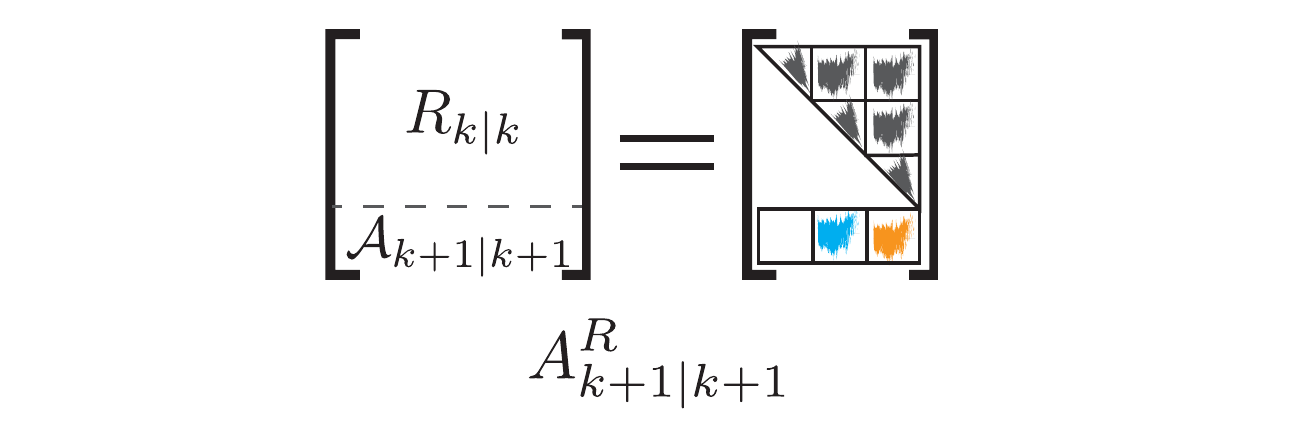}\label{fig:Q:Q:A}}
        \subfloat[]{\includegraphics[bb={0 0 0 0},trim={0 0 0 0},clip, width=0.5\columnwidth]{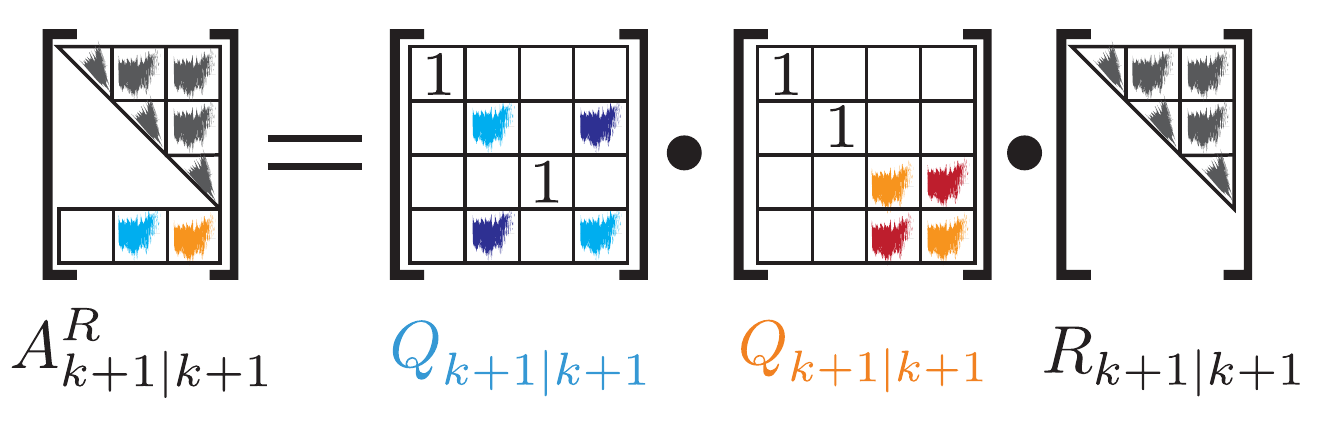}\label{fig:Q:Q:AR}}\\
        \subfloat[]{\includegraphics[bb={0 0 0 0},trim={0 0 0 0},clip, width=0.5\columnwidth]{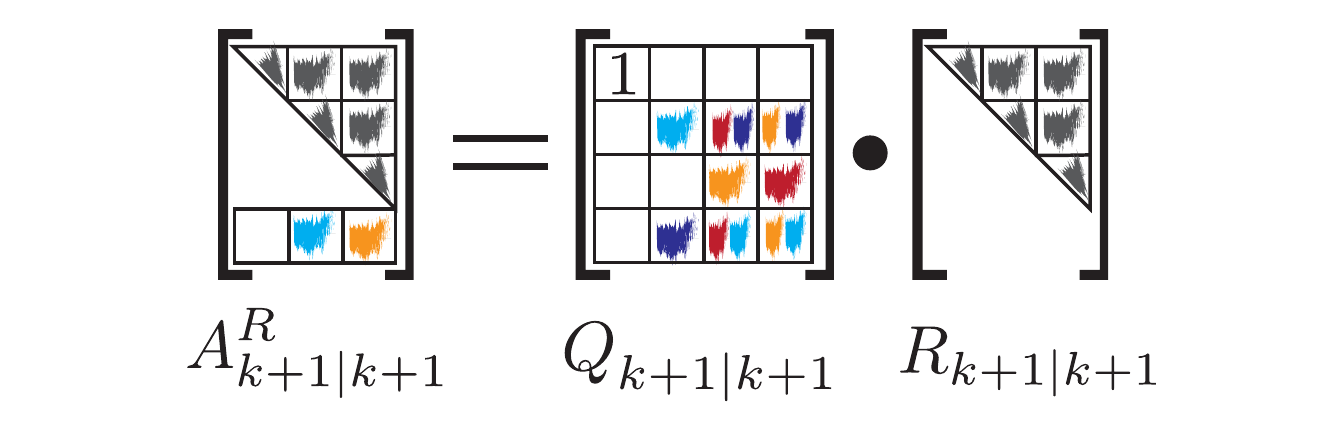}\label{fig:Q:Q:AQR}}
        \caption{(a) A Jacobian matrix at time $k+1$, consisting of the previously factorized Jacobian from time $k$ and the linearized newly added factor from time $k+1$. The RHS visualize the non zeros of the aforementioned Jacobian. (b) Visualizing the factorization procedure of the Jacobian in (a) using two Given's rotation matrices. The light and dark colors represent the cosine and sine values respectively, attributed to each of the original non zeros in (a). (c) Visualizing the non zeros in the rotation matrix required to factorize the Jacobian (a), this rotation matrix is the product of the two matrices in (b), as such the non zeros are affected by the cosine and sine values in (b).} 
        \label{fig:Q:Q}
\end{figure}
Figure~\ref{fig:Q:Q:A} illustrates a simple example for the Jacobian matrix $A^R_{k+1|k+1}$, where the precursory factorized Jacobian is denoted by $R_{k|k}$, the newly added factors by $\mathcal{A}_{k+1|k+1}$ and the columns denote the different states. The involved variables in $\mathcal{A}_{k+1|k+1}$ are marked with light blue and orange. As can be deduced from Figure~\ref{fig:Q:Q:A}, the number $A^R_{k+1|k+1}$ columns equals the joint state size at time $k+1$, and the number of $A^R_{k+1|k+1}$ rows equals the sum of the joint state size plus the number of $\mathcal{A}_{k+1|k+1}$ rows.
The purpose of factorization is to rotate $A^R_{k+1|k+1}$ to a square upper triangular form without loosing information, i.e. so that ${A^R_{k+1|k+1}}^T A^R_{k+1|k+1} = {R_{k+1|k+1}}^T R_{k+1|k+1}$. While there are many different factorization algorithms, we would consider for simplicity without affecting generality the Given's Rotation (see \cite{Golub96}). Given's rotation creates $Q_{k+1|k+1}$ by a series of simple one cell rotations. For the simple case presented in Figure~\ref{fig:Q:Q}, two rotations are required as presented in Figure~\ref{fig:Q:Q:AR}. First the left-most non zero entry in $\mathcal{A}_{k+1|k+1}$, denoted by light blue, is addressed. The appropriate rotation matrix, consists of two off-diagonal non zeros denoted by dark blue, is denoted in Figure~\ref{fig:Q:Q:AR} as the light blue $Q_{k+1|k+1}$. Next we are left to address the orange non zero entry in $\mathcal{A}_{k+1|k+1}$, while its appropriate rotation matrix, also consists of two off-diagonal non zeros denoted by dark red, is denoted in Figure~\ref{fig:Q:Q:AR} as the orange $Q_{k+1|k+1}$. From Figure~\ref{fig:Q:Q:AR} we can see that each sequential rotation matrix has the same number of non zeros, $diag\left(Q_{k+1|k+1} \right) + 2$, but due to the multiplication between them we get more non zeros as seen in Figure~\ref{fig:Q:Q:AQR}. For some intuition we marked the entries of the equivalent $Q_{k+1|k+1}$ presented in Figure~\ref{fig:Q:Q:AQR}, in accordance to the color coding in Figure~\ref{fig:Q:Q:AR}. 

Now that we understand that the number of non zeros in $Q_{k+1|k+1}$ is affected by the size of the joint state, the size of the newly added factors and the location of the left-most involved state, we are in position to formulate the expression for the number of non zeros in $Q_{k+1|k+1}$. We invite the reader to refresh his memory regarding the notations used in this analysis using Figure~\ref{fig:Q:param}, nevertheless all notations are also defined here.

Let $j$ be the column index of the left-most involved state in the newly added factors $\mathcal{A}_{k+1|k+1}$, $n^s$ be the size of the state vector (i.e. number of states multiplied by the state dimension), and $n^f$ be the number of rows of $\mathcal{A}_{k+1|k+1}$ (i.e. number of factors multiplied by the factors' dimension). The number of non zeros can be defined as the sum of three values: the number of diagonal entries equal to $1$, the contribution of the Jacobian line with the left-most state to the non zeros, and the contribution of the rest of the Jacobian lines. We will now calculate each of them.

As can be seen from Figure~\ref{fig:Q:Q:AR}, the incremental rotation matrix (i.e. colored $Q_{k+1|k+1}$) created to rotate an entry in the $i^{th}$ column, would have $i-1$ diagonal entries equal to $1$. Since the left-most state is located in the $j^{th}$ column the number of diagonal entries equal to $1$ in $Q_{k+1|k+1}$ would be
\begin{equation}
	j-1, 
\end{equation}
where $j$ is bounded by the size of the state such that
\begin{equation}\label{eq:Q:jDef}
	j \in [1,n^s]. 
\end{equation}
The rotation matrix $Q$ for rotating an entire Jacobian row located in the $i^{th}$, with a left-most non zero located in the $j^{th}$ column, would have non zeros in the $i^{th}$ row from column $j$ up to the last column and an fully dense upper triangle of non zeros over the same columns. This means that rotating the row with the left-most index in column $j$ would contribute the following number of non zeros to $Q_{k+1|k+1}$   
\begin{equation}
	\frac{n^2 - n}{2} + n + n - 1 = \frac{n^2}{2} + \frac{3}{2}n -1,
\end{equation}
where n is defined by
\begin{equation}\label{eq:Q:nDef}
	n = n^s + n^f - j + 1. 
\end{equation} 
Assuming the left-most state in the Jacobian is located in the $j^{th}$ column, rotating the rest of the rows of the Jacobian will only add non zeros at the appropriate rows in Q, without adding new non zeros to the appropriate upper triangle. The remaining $n^f-1$ rows will contribute to $Q_{k+1|k+1}$ the following number of non zeros  
\begin{equation}
	\sum_{i=1}^{n^f-1} \left( n^s + n^f - i + 1 \right) = \left( n^f-1 \right) \left( n^s + n^f + 1 \right) - \frac{n^f \left( n^f-1\right)}{2} = (n^f-1)(n^s +\frac{n^f}{2} + 1).
\end{equation}
Evidently, the number of non zeros in $Q_{k+1|k+1}$ is given by 
\begin{equation}\label{eq:Q:Qnonzeros}
	\underbrace{j-1}_i + \underbrace{\frac{n^2}{2} + \frac{3}{2}n -1}_{ii} + \underbrace{(n^f-1)(n^s +\frac{n^f}{2} + 1)}_{iii} , 
\end{equation}
where term (i) in Eq.~(\ref{eq:Q:Qnonzeros}) denotes the number of diagonal entries equal $1$,  term (ii) in Eq.~(\ref{eq:Q:Qnonzeros}) denotes the non zeros added after factorizing the factor with the left-most state j, term (iii) in Eq.~(\ref{eq:Q:Qnonzeros}) denotes the non zeros added after factorizing the rest of the factors. 
It is worth stressing that the value of $j$ is acutely affected by the ordering of the joint state vector. For better ordering, $j$ would receive larger values. 

Now that we have an expression to the number of non zeros in $Q_{k+1|k+1}$, we would like to investigate which part of it is dominant. In the sequel we reformulate Eq.~(\ref{eq:Q:Qnonzeros}) into a sum of quadratic terms, and then find conditions to determine which term is dominant.

We start by introducing (\ref{eq:Q:nDef}) into Eq.~(\ref{eq:Q:Qnonzeros}) and using simple arithmetics in order to get quadratic forms,
\begin{equation}
	\frac{1}{2}{n^s}^2 + {n^f}^2 + 2n^s n^f + \frac{3}{2}n^s - n^s j + 3n^f - n^f j + \frac{1}{2}j^2 - \frac{3}{2}j - 1	
\end{equation}
\begin{equation}
	\frac{1}{2}\left( n^s + n^f - j + \frac{3}{2} \right)^2 + \frac{1}{2}{n^f}^2 + \frac{3}{2}n^f + n^s n^f - \frac{17}{8}
\end{equation}
\begin{equation}\label{eq:Q:nzero}
	\underbrace{\frac{1}{2}\left( n^s + n^f - j + \frac{3}{2} \right)^2}_a + \underbrace{\frac{1}{2}\left( n^f + \frac{3}{2} \right)^2}_b + \underbrace{n^s n^f}_c - \frac{26}{8}.
\end{equation}
We have three candidates to be the dominant part of Eq.~(\ref{eq:Q:nzero}), denoted by terms (a) (b) and (c). Let us examine them to decide which is the dominant one and under what conditions. First we can see that term (b) in (\ref{eq:Q:nzero}) is a special case of term (a) in (\ref{eq:Q:nzero}) where $j=n^s$. Subsequently we are left with comparing terms (a) and (c) in (\ref{eq:Q:nzero}), i.e. we would like to check when
\begin{equation}
		\left(n^s+n^f-j+\frac{3}{2}\right)^2 > n^s n^f,
\end{equation} 
we define $a \triangleq n^s -j+\frac{3}{2}$ and get
\begin{equation}
		a^2 +2a n^f +{n^f}^2 - n^s n^f > 0.
\end{equation} 
So we can say term (a) in (\ref{eq:Q:nDef}) is bigger than term (c) in (\ref{eq:Q:nDef}) when
\begin{equation}
		\left(  n^s - j + \frac{3}{2} > \sqrt{n^s n^f} - n^f \right)   \quad \cup \quad   \left( n^s - j + \frac{3}{2} < -\sqrt{n^s n^f} - n^f\right) .
\end{equation} 
Considering Eq.~(\ref{eq:Q:jDef}), we can dismiss $n^s - j + \frac{3}{2} < -\sqrt{n^s n^f} - n^f$ because the smallest the LHS can be is $\frac{3}{2}$, which will always be greater than the non positive number $-\sqrt{n^s n^f} - n^f$, so the condition on $j$ so that term (a) is the dominant part of (\ref{eq:Q:nDef}) is 
\begin{equation}\label{eq:Q:jCond}
		 n^s - \sqrt{n^s n^f} + n^f  + \frac{3}{2} > j,  
\end{equation} 
which after considering Eq.~(\ref{eq:Q:jDef}) is true if and only if 
\begin{equation}
		 - \sqrt{n^s n^f} + n^f  + \frac{3}{2} > 0. 
\end{equation} 
We can now solve the aforementioned to get a condition to assure (\ref{eq:Q:jCond}) holds,
\begin{equation}
		 n^f  + \frac{3}{2} > \sqrt{n^s n^f} 
\end{equation} 
\begin{equation}
		 {n^f}^2  -(n^s-3)f + \frac{9}{4} > 0 
\end{equation} 
\begin{equation}
	 \left(n^f > \frac{n^s-3}{2} + \frac{\sqrt{{n^s}^2-6n^s}}{2}\right)  \quad \cup \quad  \left( 0 < n^f < \frac{n^s-3}{2} - \frac{\sqrt{{n^s}^2-6n^s}}{2}\right) ,
\end{equation} 
\begin{figure}
        \centering
        \includegraphics[bb={0 0 0 0},trim={0 0 0 0},clip, width=0.45\columnwidth]{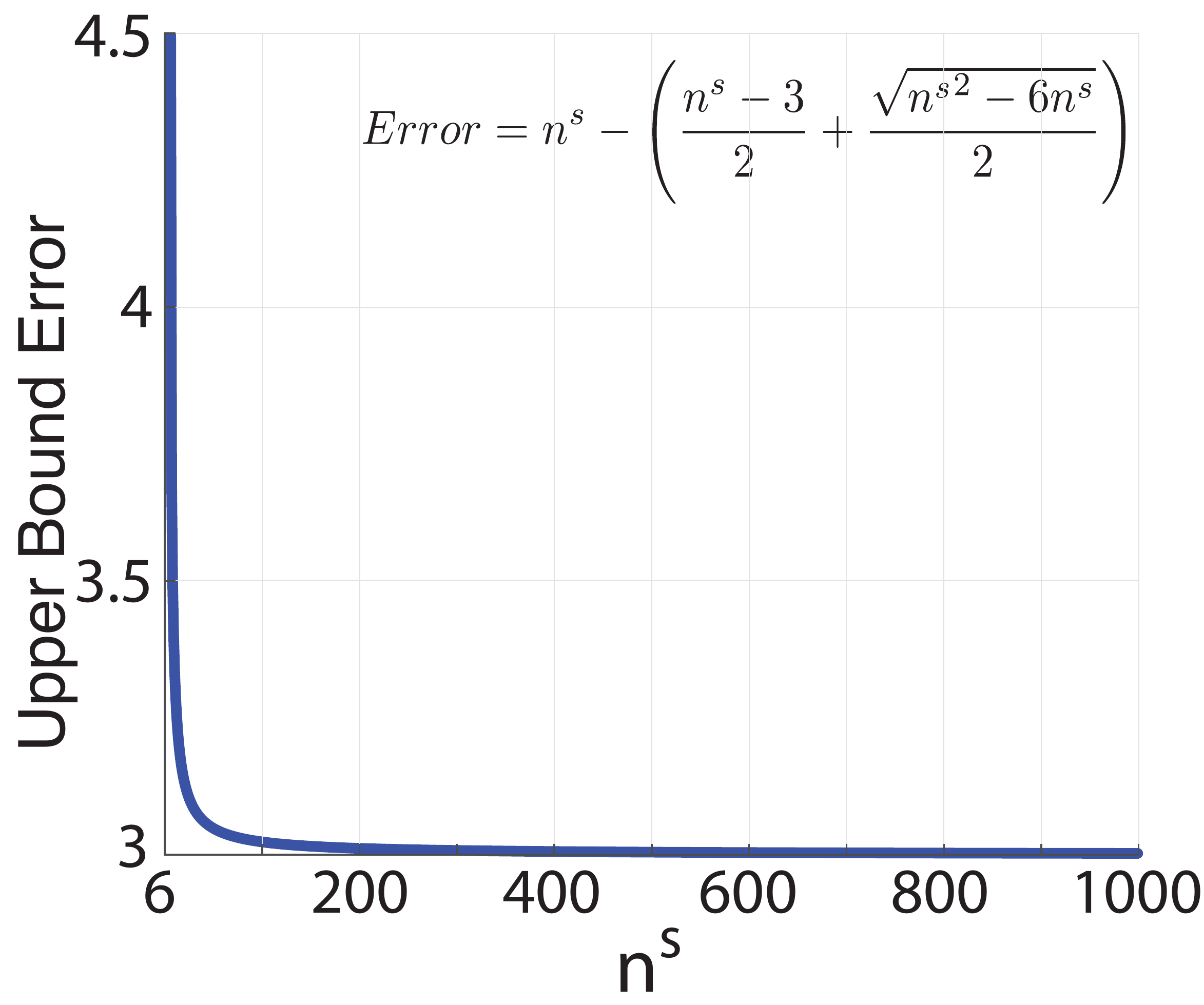}
        \caption{Illustrating the effectiveness of the bound for $\frac{n^s-3}{2} + \frac{\sqrt{{n^s}^2-6n^s}}{2}$ in the form of the error between the two as a function of different state sizes $n^s$.} 
        \label{fig:Q:bound}
\end{figure}
where $\frac{n^s-3}{2} - \frac{\sqrt{{n^s}^2-6n^s}}{2}$ is non negative $\forall n^s$, and both conditions are defined for $n^s \geq 6$ which for a 6DOF problem means a single state. For a value of $n^s=6$, $\frac{n^s-3}{2} - \frac{\sqrt{{n^s}^2-6n^s}}{2} = 1.5$ and for $n^s=7$, $\frac{n^s-3}{2} - \frac{\sqrt{{n^s}^2-6n^s}}{2} < 1$ so affectively this condition is irrelevant $\forall n^s \neq 6$, so we are left with 
\begin{equation}
	\left(n^f > \frac{n^s-3}{2} + \frac{\sqrt{{n^s}^2-6n^s}}{2}\right) \quad \cup \quad  \left( n^s \geq 6\right).
\end{equation}
Although this is the exact condition to insure term (a) is the dominant part of (\ref{eq:Q:nDef}), in order to provide a more convenient condition we suggest an upper bound in the simple form of $n^f > n^s$. Figure~\ref{fig:Q:bound} illustrates the effectiveness of the suggested bound in the form of the distance 
\begin{equation}\label{eq:Q:err}
	Error = n^s -  \frac{n^s-3}{2} + \frac{\sqrt{{n^s}^2-6n^s}}{2}.	
\end{equation}
For $n^s = 6$ the distance is $4.5$, and for $n^s=8$ it is already $3.5$, which makes this bound very affective for simplicity reasons.

To conclude, term (a) is the dominant part of (\ref{eq:Q:nDef}) if and only if the following holds
\begin{equation}\label{eq:Q:realCond}
	\left( n^f > \frac{n^s-3}{2} + \frac{\sqrt{{n^s}^2-6n^s}}{2}\right) \quad \cap \quad  \left( n^s \geq 6 \right),
\end{equation} 
or for simpler upper bound 
\begin{equation}
	n^f > n^s \geq 6.
\end{equation} 
Otherwise, term (c) is the dominant part of (\ref{eq:Q:nDef}), i.e.~given simply the size of the state and the number of rows of the newly added factors we can determine what will be the governing expression for determining the number of non zeros in $Q_{k+1|k+1}$.

\end{document}